%% file: main.tex
\documentclass{article}

\PassOptionsToPackage{numbers, compress, sort}{natbib}
\usepackage[final]{neurips_2020}

\input{preamble.tex}
\usepackage[utf8]{inputenc} % allow utf-8 input
\usepackage[T1]{fontenc}    % use 8-bit T1 fonts
\usepackage{hyperref}       % hyperlinks
\usepackage{url}            % simple URL typesetting
\usepackage{booktabs}       % professional-quality tables
\usepackage{amsfonts}       % blackboard math symbols
\usepackage{nicefrac}       % compact symbols for 1/2, etc.
\usepackage{microtype}      % microtypography
\usepackage{soul}
\usepackage{arydshln}

\def\mytitle{CoSE: Compositional Stroke Embeddings}
\title{\mytitle}

\author{%
    Emre Aksan\\ETH Zurich\\\texttt{eaksan@inf.ethz.ch} 
    \And 
    Thomas Deselaers\thanks{Work done while at Google. Unrelated to affiliation with Apple}\\Apple Switzerland\\\texttt{deselaers@gmail.com}
    \AND
    Andrea Tagliasacchi\\Google Research\\\texttt{atagliasacchi@google.com}
    \And
    Otmar Hilliges\\ETH Zurich\\\texttt{otmar.hilliges@inf.ethz.ch}
}

\begin{document}
\maketitle
\input{0_abstract.tex}

% spacing around floats - chnaged after the first page to not affect the notice string on page 1
\setlength{\topsep}{0pt}
\setlength{\parskip}{.5ex}
\renewcommand{\floatsep}{1ex}
\renewcommand{\textfloatsep}{1ex}
\renewcommand{\dblfloatsep}{1ex}
\renewcommand{\dbltextfloatsep}{1ex}

\input{1_introduction.tex}
\input{2_related.tex}
\input{3_method.tex}

\input{4_experiments.tex}

\input{8_conclusion.tex}

\clearpage
\begin{ack}
\begin{minipage}{0.70\textwidth}
The authors would like to thank Henry Rowley, Philippe Gervais, David Ha, and Andrii Maksai for their contributions to this work. 
This project has received funding from the European Research Council (ERC) under the European Union's Horizon 2020 research and innovation programme grant agreement No 717054 and from a Google Research Agreement. 
\end{minipage}%
\hfill
\begin{minipage}{0.3\textwidth}
%\begin{wrapfigure}{r}{0.30\textwidth}
\centering
\includegraphics[width=\textwidth, trim=20 150 20 150, clip]{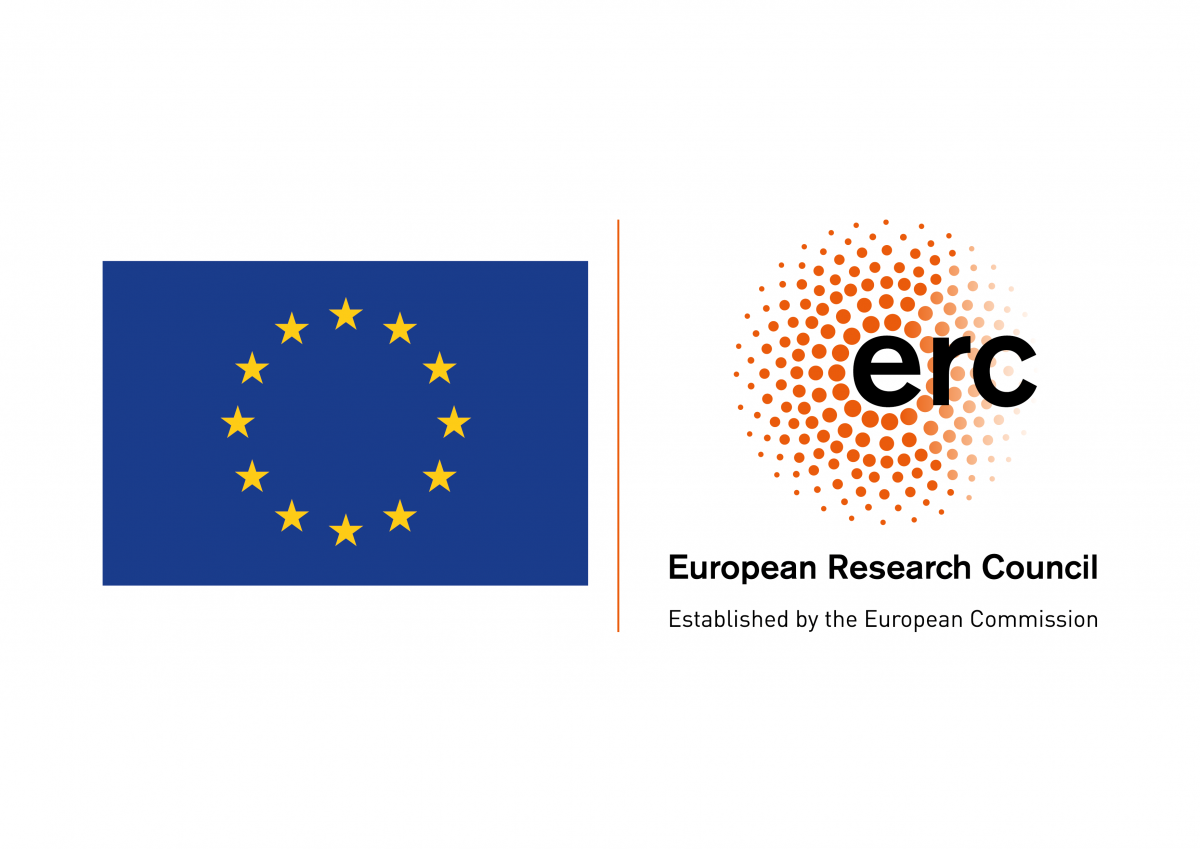}
\label{fig:erc}
%\end{wrapfigure}
\end{minipage}

\end{ack}

\input{impact_statement.tex}

{\small
\bibliographystyle{unsrtnat}
\bibliography{main}
}

\clearpage
\input{9_appendix.tex}
\end{document}

%% file: preamble.tex
\usepackage[usenames,dvipsnames]{xcolor}
\usepackage{overpic}
\usepackage{wrapfig}

\usepackage{algorithmic}
\usepackage{algorithm}
\usepackage{rotating}
\usepackage{diagbox}
\usepackage{balance}

\usepackage[utf8]{inputenc} % allow utf-8 input
\usepackage[T1]{fontenc}    % use 8-bit T1 fonts
\usepackage{url}            % simple URL typesetting
\usepackage{booktabs}       % professional-quality tables
\usepackage{amsfonts}       % blackboard math symbols
\usepackage{amsmath}
\usepackage{nicefrac}       % compact symbols for 1/2, etc.
\usepackage{microtype}      % microtypography
\usepackage[export]{adjustbox}
\usepackage{graphicx}       % figures

\usepackage{paralist}
\usepackage{xspace}
\usepackage{comment}
\usepackage{subfig}

\usepackage{array}
\newcolumntype{P}[1]{>{\centering\arraybackslash}p{#1}}
\newcolumntype{H}{>{\setbox0=\hbox\bgroup}c<{\egroup}@{}}
\usepackage{gensymb}

\newcommand{\ie}{\textit{i}.\textit{e}.\xspace}
\newcommand{\eg}{\textit{e}.\textit{g}.\xspace}

\newcommand{\vectr}[1]{\textbf{#1}}

\newcommand{\expectation}{\mathbb{E}}
\usepackage{color}
\definecolor{turquoise}{cmyk}{0.65,0,0.1,0.3}
\definecolor{purple}{rgb}{0.65,0,0.65}
\definecolor{dark_green}{rgb}{0, 0.5, 0}
\definecolor{andrea}{rgb}{0, 0.5, 0}
\definecolor{orange}{rgb}{0.8, 0.6, 0.2}
\definecolor{red}{rgb}{0.8, 0.2, 0.2}
\definecolor{blueish}{rgb}{0.0, 0.7, 1}
\definecolor{light_gray}{rgb}{0.7, 0.7, .7}
\definecolor{pink}{rgb}{1, 0, 1}
\definecolor{dark_red}{rgb}{0.5, 0, 0}
\definecolor{firststroke}{HTML}{1F77B4}
\definecolor{secondstroke}{HTML}{AEC7E8}
\definecolor{gtemb}{RGB}{1, 56, 156}
\definecolor{predemb}{RGB}{247, 177, 1}

\renewcommand{\paragraph}[1]{{\vspace{1em}\noindent \textbf{#1.}}}

\usepackage{blindtext}

\newcommand{\Fig}[1]{Fig.~\ref{fig:#1}}
\newcommand{\Figure}[1]{Figure~\ref{fig:#1}}
\newcommand{\Tab}[1]{Tab.~\ref{table:#1}}
\newcommand{\Table}[1]{Table~\ref{table:#1}}

\newcommand{\Eq}[1]{Eq.~\ref{eq:#1}}

\newcommand{\Sec}[1]{Sec.~\ref{sec:#1}}
\newcommand{\Section}[1]{Section~\ref{sec:#1}}

\renewcommand{\paragraph}[1]{{\textbf{#1.}}}
\makeatletter
\newcommand{\settitle}{\@maketitle}
\makeatother

\def\weights{\theta}
\def\drawing{\mathbf{x}}
\def\stroke{\mathbf{s}}
\def\strokes{\{\stroke_k\}}
\def\numstrokes{K}
\def\strokeindexed{\strokes_{k=1}^\numstrokes}

\def\strokecode{\boldsymbol{\lambda}}
\def\encoder{\mathcal{E}_\weights}
\def\decoder{\mathcal{D}_\weights}
\def\relational{\mathcal{R}_\weights}
\def\transformer{\mathcal{T}_\theta}
\newcommand{\initial}[1]{\bar{#1}}
\newcommand{\didi}{\texttt{DiDi}\xspace}
\newcommand{\quickdraw}{\texttt{QuickDraw}\xspace}
\newcommand{\iamondb}{\texttt{IAM-OnDB}\xspace}
\newcommand{\inlinemath}[1]{{\small #1}}

\def\modelemb{CoSE\xspace}
\def\modelseq{seq2seq\xspace}
\def\sketchrnn{Sketch-RNN\xspace}

\DeclareMathSizes{10}{9}{6.3}{4.5}
\makeatletter
\g@addto@macro\normalsize{%
\addtolength{\abovedisplayskip}{-2pt}
\addtolength{\belowdisplayskip}{-2pt}
\addtolength{\abovedisplayshortskip}{-4pt}
\addtolength{\belowdisplayshortskip}{-2pt}
}
\makeatother

\usepackage[font=small]{caption}
\newcommand{\dashrule}[1][black]{%
  \color{#1}\rule[\dimexpr.5ex-.2pt]{4pt}{.4pt}\xleaders\hbox{\rule{4pt}{0pt}\rule[\dimexpr.5ex-.2pt]{4pt}{.4pt}}\hfill\kern0pt%
}

%% file: 0_abstract.tex
% !TEX root = ../main.tex
\begin{abstract}
We present a generative model for complex free-form structures such as stroke-based drawing tasks. While previous approaches rely on sequence-based models for drawings of basic objects or handwritten text, we propose a model that treats drawings as a collection of strokes that can be composed into complex structures such as diagrams (e.g., flow-charts). At the core of the approach lies a novel auto-encoder that projects variable-length strokes into a latent space of fixed dimension. This representation space allows a relational model, operating in latent space, to better capture the relationship between strokes and to predict subsequent strokes. We demonstrate qualitatively and quantitatively that our proposed approach is able to model the appearance of individual strokes, as well as the compositional structure of larger diagram drawings. Our approach is suitable for interactive use cases such as auto-completing diagrams. We make code and models publicly available at \small{\url{https://eth-ait.github.io/cose}}.
\end{abstract}

%% file: 1_introduction.tex
% !TEX root = ../main.tex
\section{Introduction}
\label{sec:intro}

\begin{wrapfigure}{R}{0.4\textwidth}
\vspace{-3ex}   
\centering
\includegraphics[width=.7\linewidth]{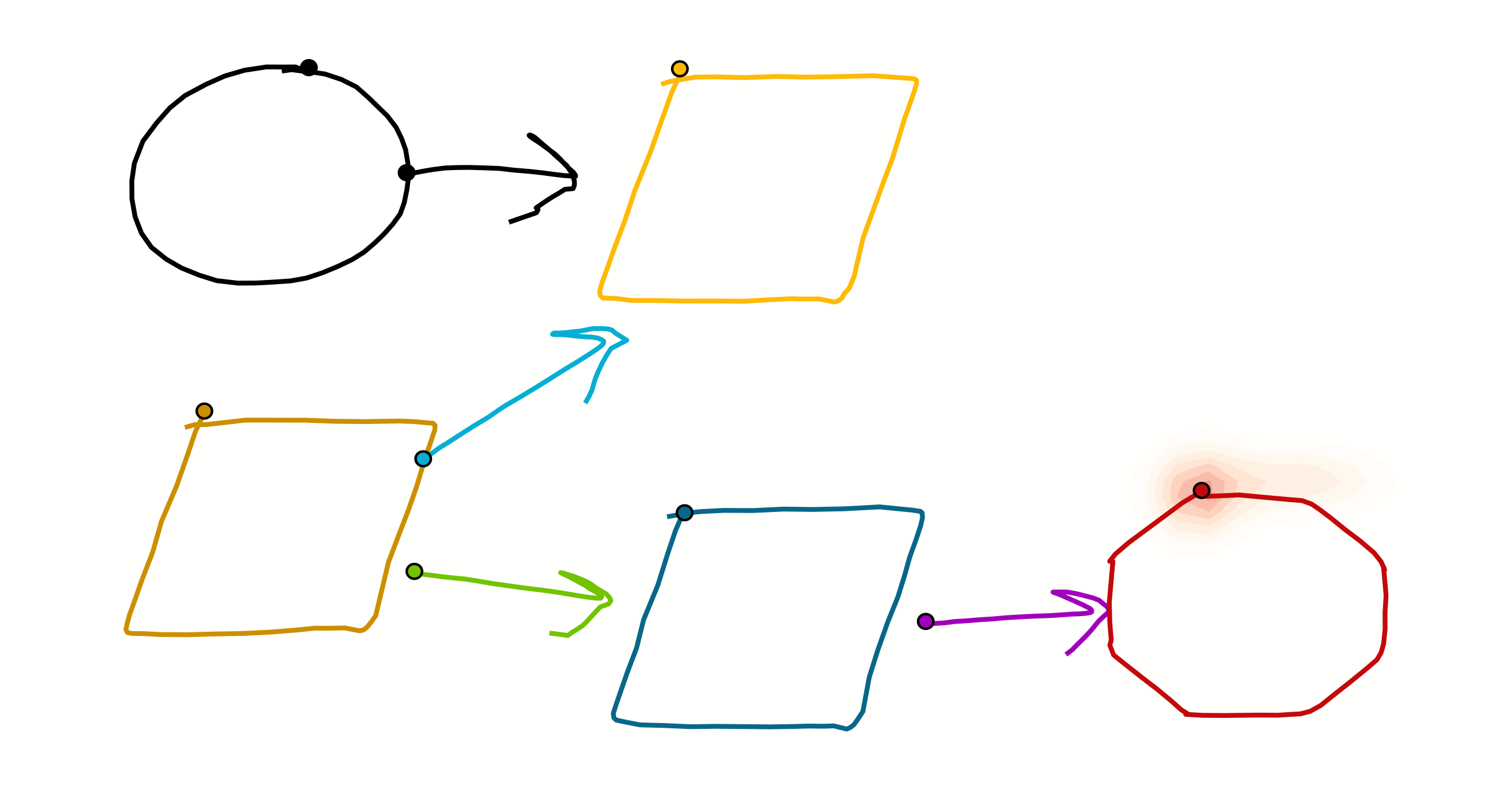} \\[-1ex]
\includegraphics[width=.4\linewidth]{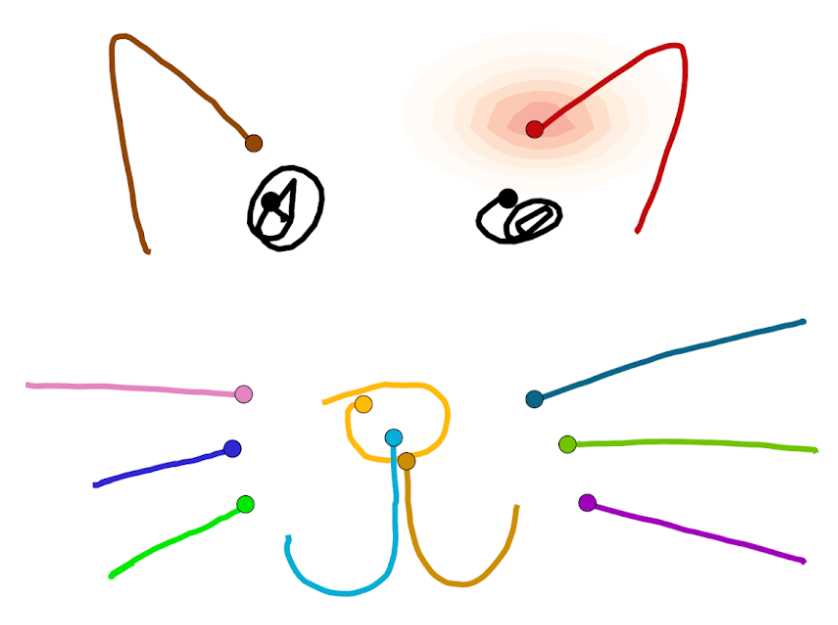}
\vspace{.1in}
\caption{\textbf{Teaser} -- We model complex drawings as a collections of strokes. Given only sparse strokes as input (black) the model predicts the most likely next strokes and their starting positions (heatmap), each color corresponds to one prediction step. This functional decomposition allows for generative modelling of varied and complex structures such as flow-charts (top), or freehand sketches (bottom). The drawings are model outputs.}
\label{fig:teaser}
\vspace{-2em}
\end{wrapfigure}

Sketches and drawings have been at the heart of human civilization for millennia.
%, diagrams are an essential element that drives innovation and communication. 
While free-form sketching is a powerful and flexible tool for humans, it is a surprisingly hard task for machines, especially if interpreted in the generative sense. Consider~\Figure{teaser}:~\textit{``when given only a sparse set of strokes (in black), what is the most likely continuation of a sketch or diagram (colored strokes are predicted)?''}
The answer to this question is highly context sensitive and requires reasoning at the local~(i.e., stroke) and global~(i.e., the diagram or sketch) level. 

Existing work has been focused on the recognition~\cite{yaeger:aaai98,pittman:2007,google-hwrlstm} and generation of handwritten text~\cite{graves:nips07-online-hwr,aksan2018deepwriting} or the modelling of entire drawings~\cite{ha2017neural,sketchformer,sketchcgn,Xu2020DeepLF} from the \emph{Quick, Draw!} dataset \cite{quickdraw-game}. However, the more recent \didi dataset introduced by Gervais et al.~\cite{didi_dataset20}, consisting of much more realistic and challenging \emph{complex structures} such as diagrams and flow-charts, has been shown to be challenging for existing methods \cite{ha-sketchrnn-on-didi}, due to the combinatorially many ways individual strokes can be combined into a complex drawing (see \Fig{sketchrnn}). %\td{because of the combinatorial explosion of possible combinations.}
% \oh{If we have a one sentence insight why this is hard we should add it here}     
% \AT{we make the argument later? (K-1)! complexity}

In this paper we propose a novel \emph{compositional} generative model, called \modelemb, for complex stroke based data such as drawings, diagrams and sketches. While existing work considers the entire drawing as a single temporal sequence~\cite{ha2017neural,sketchformer,google-hwrlstm}, our key insight is to factor \emph{local} appearance of a stroke from the \emph{global} structure of the drawing. To this end we treat each stroke as an \textit{ordered} sequence of 2D positions \inlinemath{$\stroke{=}\{(x_t, y_t)\}_{t=0}^T$}, where $(x,y)$ represents the 2D location on screen. Importantly we treat the entire drawing $\drawing$ as an \textit{unordered} collection of strokes \inlinemath{$\drawing{=}\strokeindexed$}. Since the stroke ordering does not impact the semantic meaning of the diagram, this modelling decision has profound implications. In our approach the model does not need to understand the difference between the $(K{-}1)!$ potential orderings of the previous strokes to predict the $k$-th stroke, leading to a much more efficient utilization of modelling capacity. To achieve this we propose a generative model that first projects variable-length strokes into a latent space of fixed dimension via an encoder. A relational model then predicts the embeddings of future strokes, which are then rendered by the decoder. 

The whole network is trained end-to-end and we experimentally show that the architecture can model complex diagrams and flow-charts from the \didi dataset, free-form sketches from \quickdraw and handwritten text from the \iamondb datasets \cite{iam}. We demonstrate the predictive capabilities via a proof-of-concept interactive demo (video in supplementary) in which the model suggests diagram completions based on initial user input.
% Figure \ref{fig:teaser} \oh{Or provide url to the online tool?}).
We show that our model outperforms existing models quantitatively and qualitatively and we analyze the learned latent space to provide insights into how predictions are formed.

%% file: 2_related.tex
\section{Related Work}
\label{sec:related_work}

% We are interested in generative models of stroke drawings for complex drawings of structured nature.
%
% A recent survey of related work on drawing recognition is available~\cite{Xu2020DeepLF}. 
% In the following we only review the most important papers:

% \paragraph{Recognition}
The interpretation of stroke data has been pursued before deep learning, often on small datasets of a few hundred samples targeting a particular application:
Costagliola et al.~\cite{Costagliola2006AMP} presented a parsing-based approach using a grammar of shapes and symbols where shapes and symbols are independently recognized and the results are combined using a non-deterministic grammar parser.
%
%Ouyang et al.~\cite{ouyang-iui2011,Ouyang2009NIPS} built models for chemical molecules and electrical circuits respectively using CRFs for combining multiple hand-engineered features and classifiers.
%
Bresler et al.~\cite{bresler:icfhr2014,bresler:ijdar2016} investigated flowchart and diagram recognition using a multi-stage approach including multiple independent segmentation and recognition steps.

% \paragraph{Discriminative ink models}
For handwriting recognition, neural networks have been successfully used since Yaeger at al.~\cite{yaeger:aaai98} and LSTMs have been shown to be quite successful~\cite{graves:nips07-online-hwr}.
Recently, \cite{online-diagrams2019} have applied graph attention networks to 1,300 diagrams from \cite{bresler:icfhr2014,bresler:ijdar2016,Costagliola2006AMP} for text/non-text classification using a hand-engineered stroke feature vector.
%Recently, \cite{online-diagrams2019} have applied graph attention networks to the~1,300 diagrams from \cite{bresler:icfhr2014,bresler:ijdar2016,Costagliola2006AMP} for text/non-text classification using a hand-engineered stroke feature vector.
%Interestingly, the method outperforms \cite{bresler:ijdar2016} only by a small margin. 
%Possibly this is due to the data representation not being optimized along the network parameters for the task.  
%
%One possible explanation for that could be that the input data representation is not optimized for the task. 
%
Yang et al.~\cite{sketchcgn} apply graph convolutional networks for semantic segmentation at the stroke level to extensions of the \quickdraw data~\cite{sketchseg-data,spg-data}.
For an in-depth treatment of drawing recognition, we refer the reader to the recent survey by Xu et.~al.~\cite{Xu2020DeepLF}.

% \paragraph{Generative ink models}
Particularly relevant to our work are approaches that apply generative models to stroke data. Ha et. al.~\cite{ha2017neural} and Ribeiro et. al.~\cite{sketchformer} build LSTM/VAE-based and Transformer-based models respectively to generate samples from the  \quickdraw dataset~\cite{quickdrawdata}. 
These approaches model the entire drawing as a single sequence of points. The different categories of drawings are modelled holistically without taking their internal structure into account. 
% An extension to \cite{ha2017neural} was proposed to model vector graphics which using using image en- and decoders as well as an SVG decoder~\cite{lopes2019learned}, \at{but applications are limited to fonts.} \emre{not sure if this paper is relevant.}
% It is applied on font data.
Graves proposed an auto-regressive handwriting generation model with LSTMs~\cite{graves_sequence_data}; it explicitly models the sequence structure, hence making a full-sequence representation of the ink data a reasonable choice. In \cite{aksan2018deepwriting}, an auto-regressive latent variable model is used to control the content and style aspects of handwriting, allowing for style transfer and synthesis applications.

Existing work either models the whole drawing (as an image) or as a complete sequence of points. In contrast, we  model stroke-based drawings as order invariant 2D compositional structures and in consequence our model scales to more complex settings. 
Albeit in diverse and different domains, the following works are also relevant to ours in terms of explicitly considering the compositional nature of the problems.

% \paragraph{Program synthesis and transformation}
Ellis et al.~\cite{ellis-nips17-learning-to-infer} use a program synthesis approach to analyze drawing images by recognizing primitives and combining them through program synthesis.
This approach models the structure between components, but unlike our approach the model is applied to dense image-data rather than sparse strokes directly. 
% \AT{putting together make sense?}
% \thomas{okishly}
One approach that applies neural networks to understand complex structures
based on embeddings of basic building blocks is Lee et al.~\cite{lee2019mathematical}.
%Lee et al.~\cite{lee2019mathematical} use neural networks to understand complex structures based on embeddings of basic building blocks.
They learn an embedding space for mathematical equations and use a higher-level model to predict valid transformations of equations.

% \paragraph{Layout and scene synthesis} 
Wang et al. \cite{wang2018deep} follow an iterative approach to synthesize indoor scenes where the model picks an object from a database and decides where to place it. LayoutGAN \cite{li2019layoutgan} learns to generate realistic layouts from 2D wireframes and semantic labels for documents and abstract scenes.

% \thomas{removed didi here - as it is in the intro}
%More recently, a new dataset of about 60,000 drawings of diagrams was made available \cite{didi_dataset20}, which aims to allow for deep learning methods to be applied to model more structured drawings. 
%We are using this dataset for our experiments. 

% \thomas{a few more potential citations are now commented out - most of which are cited in the method section already}

% \paragraph{Geometric deep learning}
% \AT{I'll write this later, if space allows?}
% AtlasNet, OccNet, DeepPoly (fonts), IM-NET, PointNet
% \lorem{1}

% \paragraph{Generative Models} 
% Relevant technical stuff
% \begin{itemize}
%     \item \cite{Park_2019_CVPR} \cite{atlasnet} \cite{cppn}
%     \item \cite{transformer} transformer
% \end{itemize}

% Modeling in a latent space:
% \begin{itemize}
%     \item  Mathematical equation prediction in latent space.
%     \item \cite{ha2018world} World models.
% \end{itemize}

% \AT{we'd not go beyond page 2 with related!}
% \thomas{yes - but shortening is easier than writing}

%% file: 3_method.tex
% !TEX root = ../main.tex

\begin{figure*}[t]
\centering
\vspace{-1em}
\includegraphics[width=\linewidth]{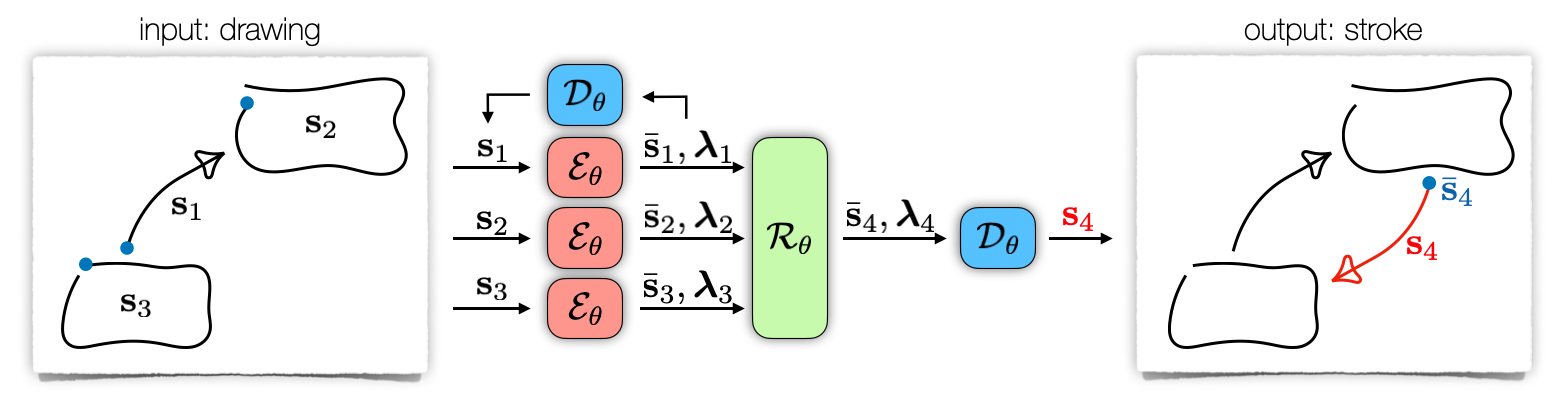}
\caption{
\textbf{Architecture overview} --
(left) the input drawing as a collection of strokes $\strokes$;
(middle) our embedding architecture, consisting of a shared encoder $\encoder$, a shared decoder $\decoder$, and a relational model~$\relational$;
(right) the input drawing with the next stroke $\stroke_4$ and its starting position $\initial\stroke_4$ predicted by $\relational$ and decoded by~$\decoder$. Note that the relational model $\relational$ is permutation-invariant.
}
\label{fig:outline}
\end{figure*}

\section{Method}
\label{sec:method}

We are interested in modelling a drawing $\drawing$ as a collection of strokes $\strokeindexed$, in the following abbreviated as $\{\stroke_k\}$, which requires capturing the semantics of a sketch and learning the relationships between its strokes.
We propose a generative model, dubbed \modelemb, that first projects variable-length strokes into a fixed-dimensional latent space, and then models their relationships in this latent space to predict future strokes. This approach is illustrated in~\Figure{outline}.
More formally, given an initial set of strokes (\eg~$\{\stroke_1,\stroke_2,\stroke_3\}$), we wish to predict the next stroke (\eg~$\stroke_4$). 
We decompose the joint distribution of the sequence of strokes $\vectr{x}$ as a product of conditional distributions over the set of existing strokes:
\begin{align}
    p(\vectr{x}; \weights) &= \prod_{k=1}^K p(\stroke_k, \initial\stroke_k | \stroke_{<k}, \initial\stroke_{<k}; \weights),% \text{where} \quad {<}k = \{1 \dots k-1\}
\label{eq:stroke_factorization}
\end{align}
with $\initial\stroke_k$ referring to the starting position of the $k$-th stroke, and ${<}k$ denotes $\{1 \dots k{-}1\}$. Note that we assume a fixed but not chronological ordering of $K$. An encoder $\encoder$ first encodes each stroke $\stroke$ to its corresponding latent code $\strokecode$.
A decoder $\decoder$ reconstructs the corresponding~$\stroke$, given a code $\strokecode$ and the starting position $\initial\stroke$.
A transformer-based relational model $\relational$ processes the latent codes $\{\strokecode_{<k}\}$ and their corresponding starting positions $\{\initial\stroke_{<k}\}$ to generate the next stroke starting position $\initial\stroke_{k}$ and embedding $\strokecode_{k}$, from which~$\decoder$ reconstructs the output stroke~$\stroke_{k}$.
Overall, our architecture factors into a \textit{stroke embedding} model~($\encoder$ and $\decoder$) and a \textit{relational model}~($\relational$).

\paragraph{Stroke embedding -- \Section{stroke_embedding}}
We force the embedding model to capture \emph{local} information such as the shape, size, or curvature by preventing it  from accessing \emph{any} global information such as the canvas position or existence of other strokes and their inter-dependencies. 
The auto-encoder generates an abstraction of the variable-length strokes $\stroke$ by encoding them into fixed-length embeddings $(\strokecode, \initial\stroke) {=} \encoder(\stroke)$ and decoding them into strokes $\stroke {=} \decoder(\strokecode, \initial\stroke)$.

\paragraph{Relational model -- \Section{composition}}
Our relational model learns how to \emph{compose} individual strokes to create a sketch by considering the relationship between latent codes.
Given an input drawing encoded as $\vectr{x} {=} \{(\strokecode_{<k}, \initial\stroke_{<k})\}$,
%_{k=1}^K$, 
we predict: 
i) a starting position for the next stroke $\initial\stroke_{k}$, and ii) its corresponding embedding $\strokecode_{k}$.
Introducing the embeddings into~\Eq{stroke_factorization}, we obtain our \emph{compositional stroke embedding} model that decouples \emph{local} drawing information from \emph{global} semantics:
\begin{align}
    p(\vectr{x}; \weights) &= \prod_{k=1}^K p(\strokecode_k, \initial\stroke_k | \strokecode_{<k}, \initial\stroke_{<k}; \weights)
\label{eq:cose}
\end{align}
We train by maximizing the log-likelihood of the network parameters $\weights$ on the training set.

\subsection{Stroke embeddings}
\label{sec:stroke_embedding}
We represent variable-length strokes $\vectr{s}$
with fixed-length embeddings $\strokecode {\in} \mathbb{R}^D$.
The goal is to learn a representation space of the strokes such that it is informative both for \emph{reconstruction} of the original strokes, and for \emph{prediction} of future strokes.
We now detail our auto-encoder architecture: 
the encoder $\encoder$ is based on transformers~\cite{transformer}, 
while the decoder $\decoder$ extends ideas from neural modeling of differential geometry~\cite{atlasnet}. The parameters of both networks are trained via:
\begin{align}
\arg\max_\theta \:\: \expectation_{t\sim [0,1]} \:
\sum_{m=1}^M \pi_{t,m} \: \mathcal{N}(\stroke_k(t) \:|\: \mu_{t,m}, \sigma_{t,m} ),
\quad\quad
\{\mu_{t,m}, \sigma_{t,m}, \pi_{t,m}\} = \decoder(t | \encoder(\stroke))
\label{eq:embedding_training}
\end{align}
where we use mixture densities \cite{Bishop94mixturedensity, graves_sequence_data} with $M$ Gaussians with mixture coefficients $\pi$, mean $\mu$ and variance $\sigma$; $t\in[0,1]$ is the curve parameter.
Note that we use log-likelihood rather than Chamfer Distance as in~\cite{atlasnet}. While we do interpret strokes as 2D curves, we observe that modelling of prediction uncertainity is commonly done in the ink modelling literature \cite{graves_sequence_data, aksan2018deepwriting, ha2017neural} and has been shown to result in better regression performance compared to minimizing an L2 metric~\cite{kumar2019uglli} (cf. Sec. \ref{sec:appendix_MSE} in supplementary).

\begin{figure}[t!]
\centering
\includegraphics[width=\columnwidth, trim={0pt 0pt 0pt 0pt}]{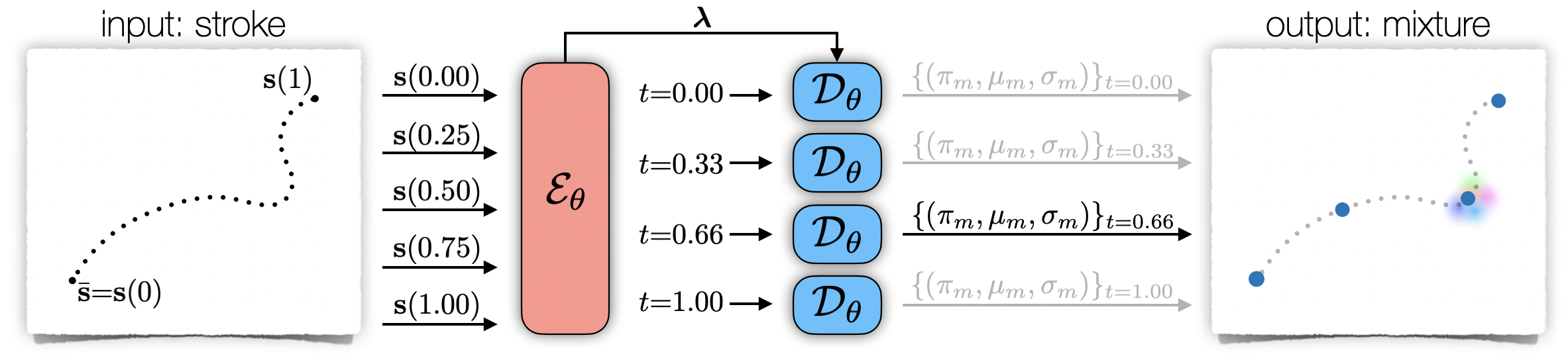}
\caption{\textbf{Stroke embedding} -- The input stroke $\stroke$ is passed to the encoder, which produces a latent code $\strokecode$. The decoder parameterizes a Gaussian mixture model for arbitrary positions $t{\in}[0,1]$ from which we sample points on the stroke.
We only visualize the mixture model associated with $t{=}.66$ (non-grayed out arrow). 
}
\label{fig:stroke_autoencoder}
\end{figure}

\paragraph{\modelemb Encoder -- $\encoder(\stroke)$}
We encode a stroke by viewing it as a sequence of 2D points, and generate the corresponding latent code with a transformer $\transformer^t$, where the superscript $t$ denotes use of positional encoding in the temporal dimension~\cite{transformer}. The encoder outputs $(\initial\stroke, \strokecode){=}\encoder(\stroke)$, where $\initial\stroke$ is the starting position of a stroke and $\strokecode{=}\transformer^t(\stroke-\initial\stroke)$.
The use of positional encoding induces a point ordering and emphasizes the geometry, where most sequence models focus strongly on capturing the drawing dynamics.
Furthermore, avoiding the modelling of explicit temporal dependencies between time-steps allows for inherent parallelism and is hence computationally advantageous over RNNs.

\paragraph{\modelemb Decoder -- $\decoder(t | \initial\stroke, \strokecode)$}
We consider a stroke as a curve in the differential geometry sense: a 1D manifold embedded in 2D space.
As such, there exists a differentiable map $s: \mathbb{R} \rightarrow \mathbb{R}^2$ between $t{\in}[0,1]$ and the 2D planar curve~$s(t){=}(x_t, y_t)$.
Groueix et~al.~\cite{atlasnet} proposed to represent 2D (curves) and 3D (surfaces) geometry via MLPs that approximate $s(t)$.
Recently it has been shown that representing curves via dense networks induces an implicit smoothness regularizer~\cite{williams2019gradient, gadelha2020dmp} akin to the one that CNNs induce on images~\cite{ulyanov2018deep}.
Since we do not want to reconstruct a single curve~\cite{gadelha2020dmp}, we employ the latent code provided by $\encoder$ to condition our decoder jointly with the curve parameter $t$: $\decoder(t | \strokecode, \initial\stroke) = \initial\stroke + \text{MLP}_\theta([t, \strokecode])$~\cite{atlasnet} which parameterizes the Gaussian mixture from \Eq{embedding_training}.

\paragraph{Inference}
At inference time, a stroke is reconstructed by using $t$ values sampled at (consecutive) regular intervals as determined by an arbitrary sampling rate; see~\Figure{stroke_autoencoder}.
Note how compared to RNNs, we do not need to predict an ``end of stroke'' token, as the decoder output for $t{=}1$ corresponds to the end of a stroke.
Therefore the length of the reconstructed sequence depends on how densely we sample the parameter $t$.

\subsection{\modelemb Relational model -- $\relational$}% -- \Figure{relational_model}}
\label{sec:composition}
We propose a generative model that auto-regressively estimates a joint distribution over stroke embeddings and positions given a latent representation of the current drawing in the form $\drawing {=} \{(\strokecode_{<k}, \initial\stroke_{<k})\}$.
We hypothesize that, in contrast to handwriting, local context and spatial layout are important factors that are not influenced by the drawing order of the user.
We exploit the self-attention mechanism of the transformer \cite{transformer} to learn the \textit{relational dependencies} between strokes in the latent space.
In contrast to the stroke embedding model (\Sec{stroke_embedding}), we \emph{do not} use positional encoding to prevent any temporal information to flow through the relational model.

\paragraph{Prediction factorization}
In drawings, the starting position is an important degree of freedom. Hence, we split the prediction of the next stroke into two tasks:
i) the prediction of the stroke's starting position $\initial\stroke_{k}$,
and 
ii) the prediction of the stroke's embedding $\strokecode_{k}$.
Given the (latent codes of) initial strokes, and their starting positions~$\{(\strokecode_{<k}, \initial\stroke_{<k})\}$, this results in the factorization of the joint distribution over the strokes $\stroke_k$ and positions~$\initial\stroke_k$ as a product of conditional distributions:
\begin{align}
    p(\strokecode_k, \initial\stroke_k | \strokecode_{<k}, \initial\stroke_{<k}; \weights) = 
    \:\:
    \underbrace{p(\initial\stroke_k | \strokecode_{<k}, \initial\stroke_{<k}; \weights)}_\text{starting position prediction}
    \:\:
    \underbrace{p(\strokecode_k | \initial\stroke_k, \strokecode_{<k}, \initial\stroke_{<k}; \weights)}_\text{latent code prediction}
\label{eq:stroke_pos_factorization}
\end{align}
By conditioning on the starting position, the attention mechanism in $\relational$ focuses on a \textit{local} context, allowing our model to perform more effectively (see also Sec. \ref{Sec:ablations}).
We use two separate transformers with the same network configuration yet slightly different inputs and outputs:
i) the position prediction model takes the set $\{(\stroke_{<k}, \initial\stroke_{<k})\}$  as input and produces $\initial\stroke_{k}$;
ii) the embedding prediction model takes 
the next starting position $\initial\stroke_{k}$ as additional input to predict~$\strokecode_{k}$.
Factorizing the prediction in this way has two advantages:
i) all strokes start at the origin, hence we can employ the translational-invariant embeddings from~\Sec{stroke_embedding};
ii) it enables interactive applications, where the user specifies a starting position and the model predicts an auto-completion; see video in the supplementary and~\Figure{relational_model}.

\begin{figure}[t!]
\centering
\strut\hfill
\includegraphics[width=.4\columnwidth]{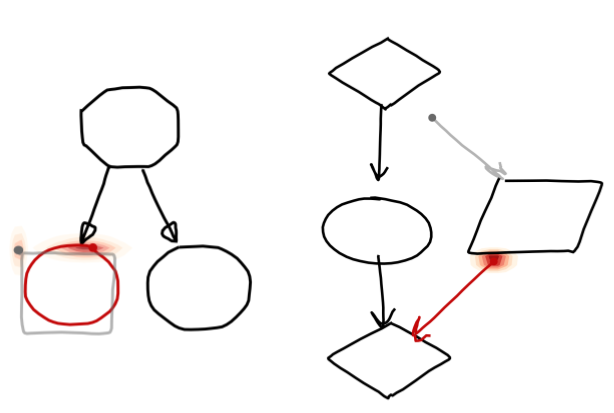}
\hfill
\includegraphics[width=.4\columnwidth]{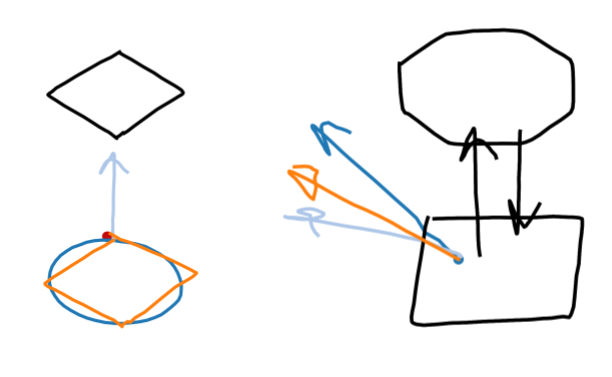}
\hfill\strut
\caption{\textbf{Relational model} --
A few snapshots from our live demo.
(Left) Given a drawing, our model proposes \textit{several} starting position for auto-completion (we draw the most likely strokes associated with the two most likely starting positions (red, gray)).
(Right) Given a stating position, our model can predict \textit{several} stroke alternatives; here we show the top $3$ most likely predictions (orange, light blue, dark blue). 
}
\label{fig:relational_model}
\end{figure}

\paragraph{Starting position prediction}
The prediction of the next starting positions is inherently multi-modal, since there may be multiple equally good predictions in terms of drawing continuation; see~\Figure{relational_model} (left).
We employ multi-modal predictions in the form of a $2$-dimensional Gaussian Mixture. In the fully generative scenario, we sample a position $\initial\stroke_{k}$ from the predicted GMM, rather than expecting a user input or ground-truth data as at training time.

\paragraph{Latent code prediction} 
Given a starting position, multiple strokes can be used to complete a given drawing; see~\Figure{relational_model} (right).
We again use a Gaussian Mixture to capture this multi-modality.
At inference time, we sample from $p(\strokecode_k | \initial\stroke_k, \strokecode_{<k}, \initial\stroke_{<k}; \weights)$ to generate~$\strokecode_k$. Thanks to order-invariant relational model, \modelemb can predict over long prediction horizons  (see \Fig{prediction}).

\subsection{Training}
\label{sec:training}

Given a random pair of target $(\strokecode_k, \initial\stroke_k)$ and a subset of inputs $\{(\strokecode_{\neq k}, \initial\stroke_{\neq k})\}$, we make a prediction for the position and the embedding of the target stroke.
This subset is obtained by selecting $H{\in}[1,K]$ strokes from the drawing. 
We either pick $H$ strokes i) in order, or ii) at random. 
This allows the model to be good in completing existing partial drawings but also be robust to arbitrary subsets of strokes.
During training, the model has access to the ground-truth positions $\initial\stroke$ (like teacher forcing~\cite{teacher-forcing}).
Note that while we train all three sub-modules (encoder, relational model, decoder) in parallel, we found that the performance is slightly better if gradients from the relational model (\Eq{cose}), are not back-propagated through the stroke embedding model. 
We apply augmentations in the form of random rotation and re-scaling of the entire drawing (see supplementary for details).

%% file: 4_experiments.tex
% !TEX root = ../main.tex
\section{Experiments}
\label{sec:experiments}

We evaluate our model on the recently released \didi dataset~\cite{didi_dataset20}. 
In contrast to existing drawing~\cite{quickdrawdata} or handwriting datasets~\cite{iam}, this task requires learning of the \emph{compositional structure} of flowchart diagrams, consisting of several shapes. In this paper we focus on the predictive setting in which an existing (partial) drawing is extended by adding more shapes or by connecting already drawn ones. 
State-of-the-art techniques in ink modelling treat the entire drawing as a \emph{single} sequence. Our experiments demonstrate that this approach does not scale to complex structures such as flowchart diagrams (cf. \Fig{sketchrnn}). 
We compare our method to the state-of-the-art~\cite{ha2017neural} via the Chamfer Distance~\cite{qi2017pointnet} between the ground-truth strokes and the model outputs (\ie reconstructed or predicted strokes).

The task is inherently stochastic as the next stroke highly depends on where it is drawn. To account for the high variability in the predictions across different generative models, the ground-truth starting positions are passed the models in our quantitative analysis (note that the qualitative
results rely only on the predicted starting positions). Moreover, similar to most predictive tasks, there is no single, correct prediction in the stroke prediction task (see \Fig{relational_model}). To account for this multi-modality of fully generative models, we employ a \textit{stochastic} variant of the Chamfer Distance~(CD):
\begin{align}
\min_{\strokecode_k \sim p(\strokecode_k | \initial\stroke_k, \strokecode_{<k}, \initial\stroke_{<k}; \weights)} \left\{ \text{CD}\left(\decoder(t | \hat{\strokecode}_k), \: \stroke_k\right)\right\}.
\end{align}
We evaluate our models by sampling one $\strokecode_k$ from each mixture component of the relational model's prediction which are decoded into $10$ strokes (see Fig. \ref{fig:gmm_k_ablation}). This results in a broader exploration of the predicted strokes than a strict Gaussian mixture sampling. Note that while our training objective is NLL (as is common in ink modelling), the Chamfer Distance allows for a fairer comparison since it allows to compare models trained on differently processed data (i.e., positions vs offsets). 

\begin{figure}[t]
\includegraphics[width=.48\columnwidth,trim=0pt 0pt 0pt 90pt]{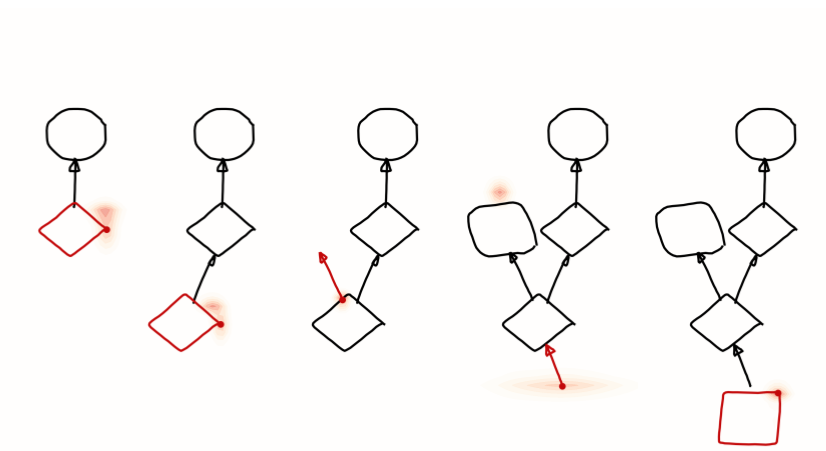}
\hfill{}
\includegraphics[width=.48\columnwidth, trim={0pt 0pt 0pt 90pt}]{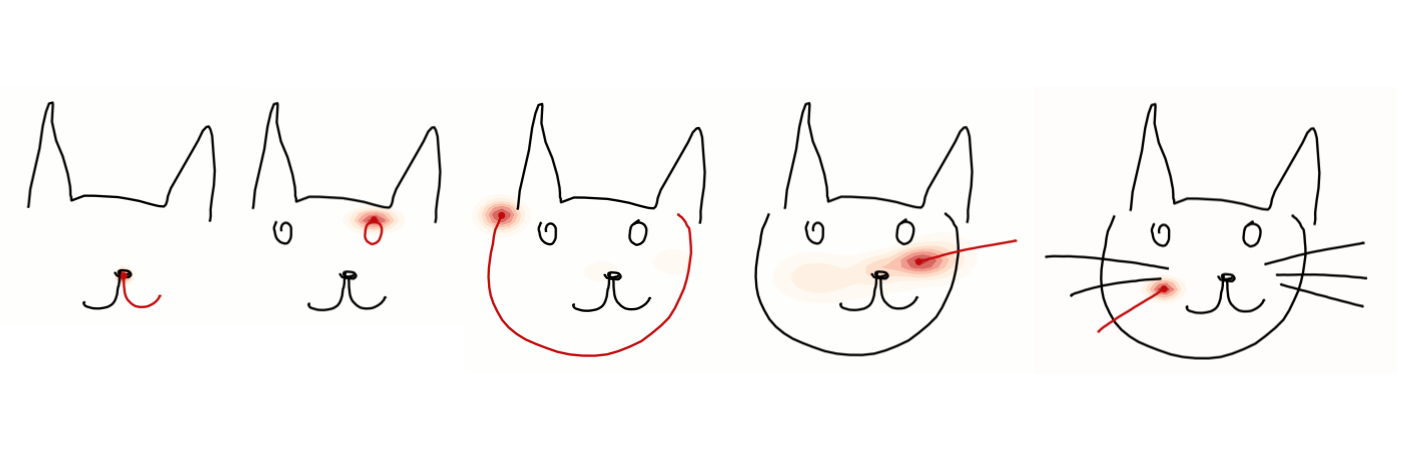}\\
\vspace{-1ex}
\caption{\textbf{Auto-regressive completion.} Performed by CoSE trained on \didi and \quickdraw datasets.}
\label{fig:prediction}
\end{figure}

\subsection{Stroke prediction}
\label{sec:exp_full_model}

\input{fig/res_table}

We first evaluate the performance in the \emph{stroke prediction} setting. Given a set of strokes and a target position, the task is to predict the next stroke. For each drawing, we start with a single stroke and incrementally add more strokes from the original drawing (in the order they were drawn) to the set of given strokes and predict the subsequent one. 
In this setting we evaluate our method in an ablation study, where we replace components of our model with standard RNN-based models: 
a sequence-to-sequence (\modelseq) architecture~\cite{sutskever2014sequence} for stroke-embeddings, and an auto-regressive RNN for the relational model (see supplementary for architecture details). 
Furthermore, following the setting in~\cite{ha2017neural}, we compare to the decoder-only setup from Sketch-RNN (itself conceptually similar to Graves~et~al.~\cite{graves_sequence_data}).
% \oh{We should keep setup and results separate} Now details on the models:
For the \modelseq-based embedding model we use bi-directional LSTMs \cite{hochreiter1997long} as the encoder, and a uni-directional LSTM as decoder.
Informally we determined that a deterministic encoder with a non-autoregressive decoder outperformed other \modelseq architectures; see \Sec{exp_emb_model}.
The RNN-based relational model is an auto-regressive sequence model~\cite{graves_sequence_data}.

\paragraph{Analysis}
The results are summarized in~\Table{res_table}.
While the stroke-wise reconstruction performance across all models differs only marginally, the predictive performance of our proposed model is substantially better. 
This indicates that a standard \modelseq model is able to learn an embedding space that is suitable for accurate reconstruction, this embedding space however \emph{does not} lend itself to predictive modelling.
The combination of our embedding model (\modelemb-$\encoder$/$\decoder$) with our relational model (\modelemb-$\relational$) outperforms all other models in terms of predicting consecutive strokes, giving an indication that the learned embedding space is better suited for the predictive downstream tasks. 
The results also indicate that the contributions of both are necessary to attain the best performance. This can be seen by the increase in prediction performance of the \modelseq when augmented with our relational model (\modelemb-$\relational$). However, a significant gap remains to the full model (cf. row 2 and 5). 

\begin{figure}[tb]
\newcommand{\sketchrnnfig}[1]{%
{\begin{minipage}[b]{.32\linewidth}
\includegraphics[width=.49\linewidth,height=\linewidth,keepaspectratio,valign=t]{fig/sketchrnn_vs_cose/cose_sample#1}
\includegraphics[width=.49\linewidth,height=\linewidth,keepaspectratio,valign=t]{fig/sketchrnn_vs_cose/sketchrnn_sample#1}\\[-2ex]
\scriptsize\bf\strut\hfill\modelemb\hfill\hfill\sketchrnn\hfill\strut
\end{minipage}}}
\sketchrnnfig{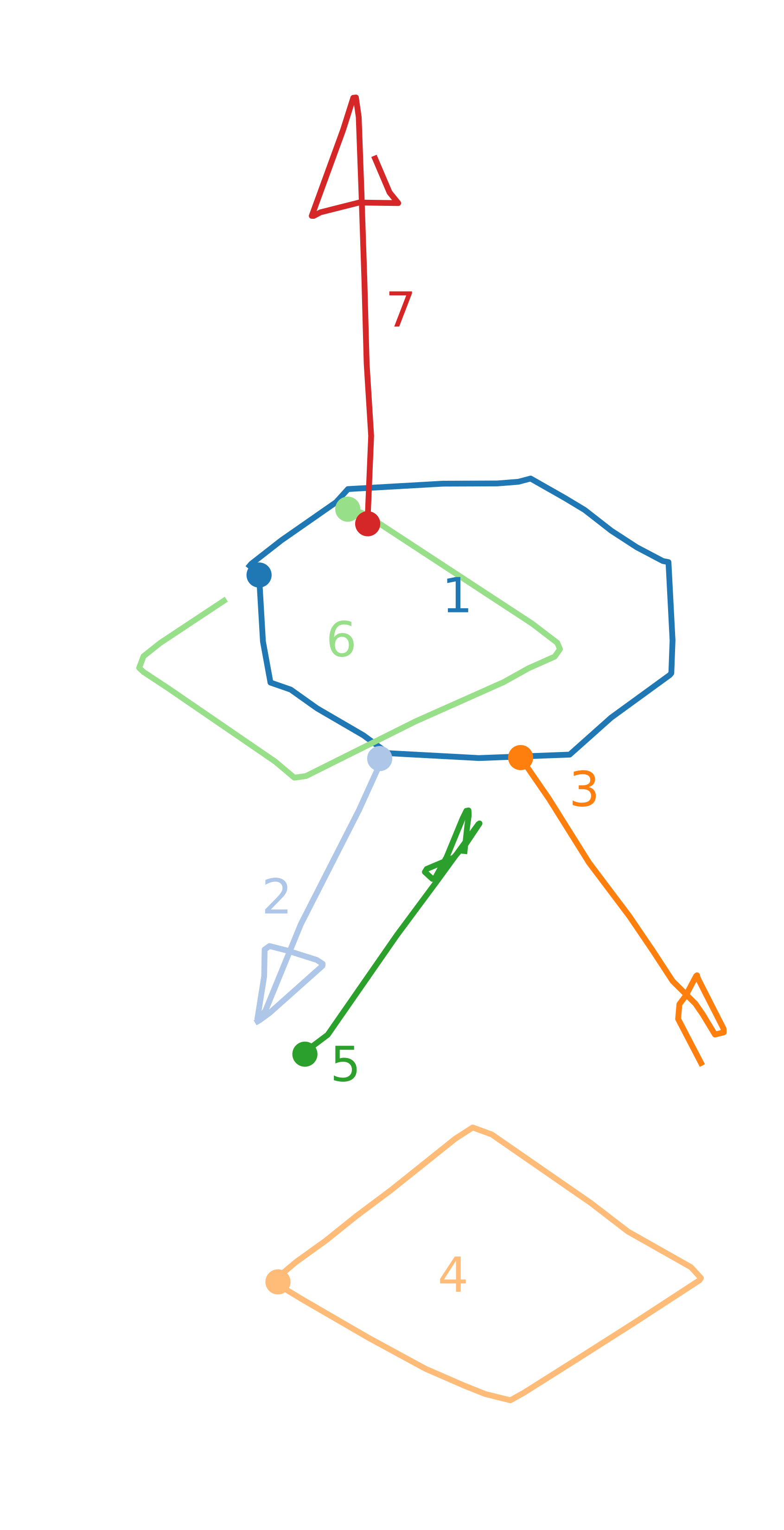}%
\sketchrnnfig{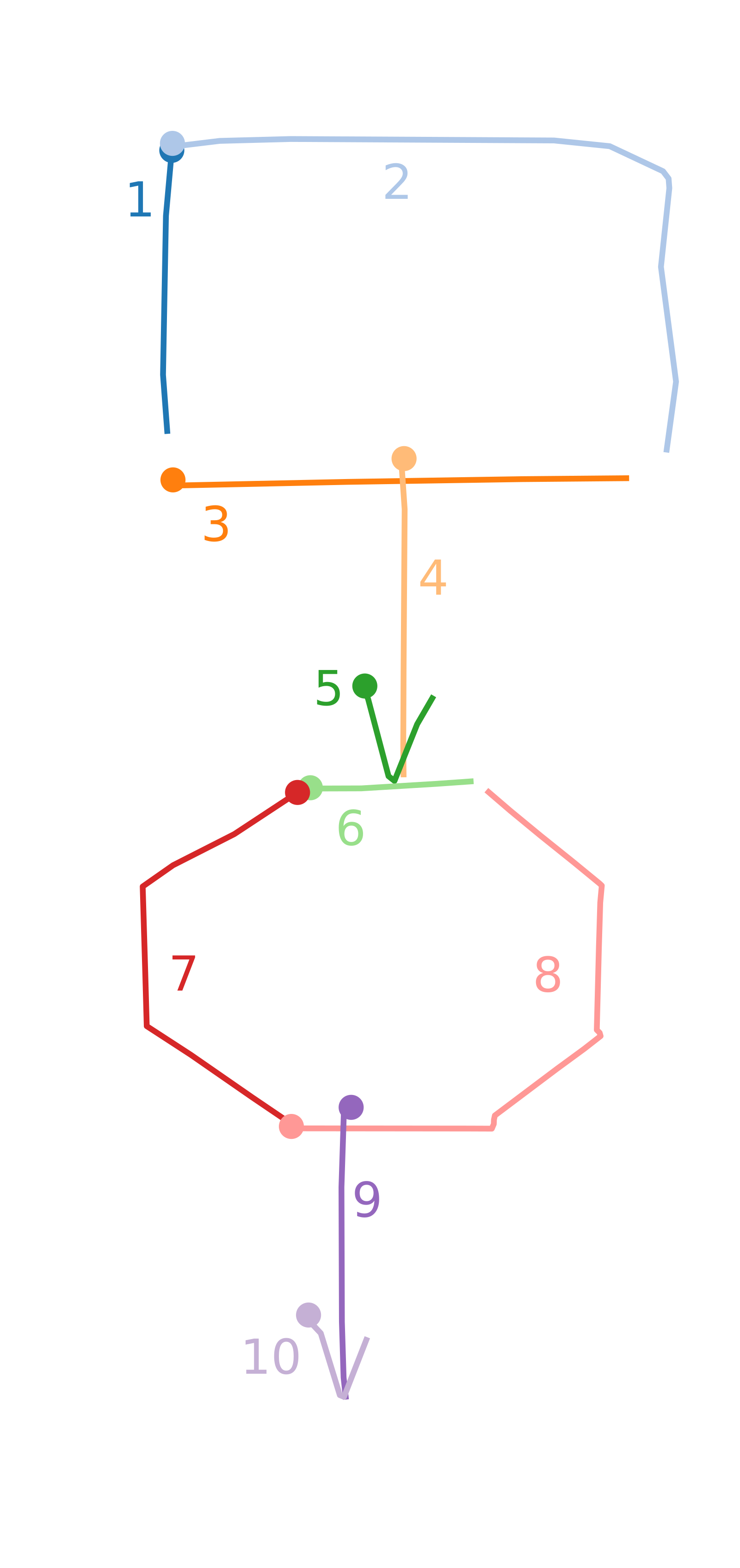}%
{\begin{minipage}[b]{.32\linewidth}
\centering
\scriptsize\bf
\includegraphics[width=\linewidth,height=.5\linewidth,keepaspectratio]{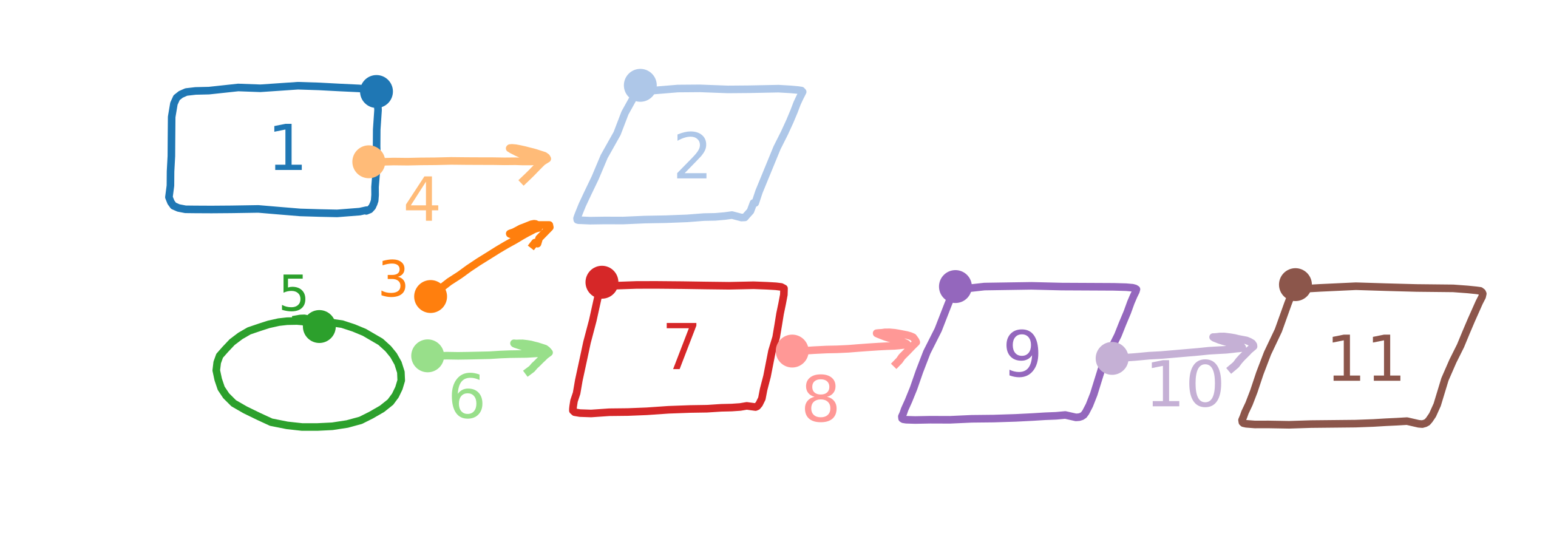}\\[-2ex]
\modelemb\\
\includegraphics[width=\linewidth,height=.6\linewidth,keepaspectratio]{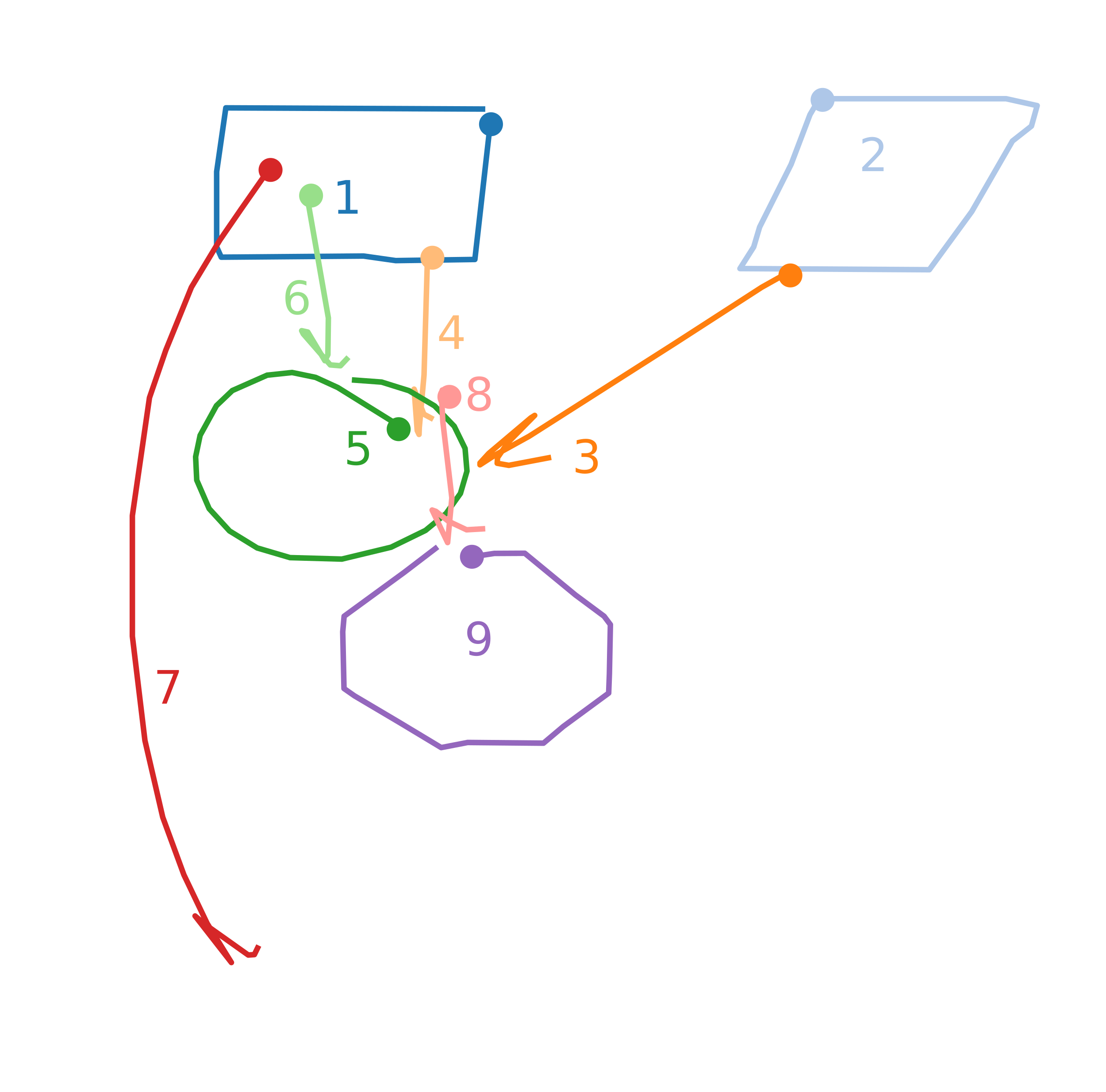}\\[-2ex]
\sketchrnn\\
\end{minipage}}
\vspace{1ex}
\caption{{\bf Comparison with \sketchrnn} -- For each pair of samples the first two strokes (in blue) are given as context, the remaining strokes (in color) are model outputs, numbers indicate prediction step. While \sketchrnn produces meaningful completions for the first few predictions, its performance quickly decreases with increasing complexity. In contrast, \modelemb is capable of predicting plausible continuations even over long prediction horizons.}
\label{fig:sketchrnn}
\end{figure}

We also evaluate our full model with additional positional encoding in the relational \modelemb-$\relational$~(Ord.). The results support our hypothesis that an order-invariant model is beneficial for the task of modelling compositional structures. It is also observed in sequential modelling of the stroke embeddings by using an RNN (row 3).
Similarly, our model outperforms \sketchrnn which treats drawings as sequence. We show a comparison of flowchart completions by \sketchrnn and \modelemb in \Fig{sketchrnn}. Our model is more robust to make longer predictions.

% %

\subsection{Stroke embedding}
\label{sec:exp_emb_model}
Our analysis in \Sec{exp_full_model} revealed that good reconstruction accuracy is not necessarily indicative of an embedding space that is useful for fully auto-regressive predictions. 
We now investigate the structure of our embedding space in qualitative and quantitative measures by analyzing the performance of clustering algorithms on the embedded data.
Since there is only a limited number of shapes that occur in diagrams, the expectation is that a well shaped latent space should form clusters consisting of similar shapes, while maintaining sufficient variation.

\paragraph{Silhouette Coefficient (SC)} 
This coefficient is a quantitative measure that assesses the quality of a clustering by jointly measuring tightness of exemplars within clusters vs.\ separation between clusters~\cite{silhouettecoefficient}. 
It does not require ground-truth cluster labels (\eg whether a stroke is a box, arrow, arrow tip), and takes values between $[-1, 1]$ where a higher value is an indication of tighter and well separated clusters. The exact number of clusters is not known and we therefore compute the SC for the clustering result of $k$-means and spectral clustering~\cite{shi2000normalized} with varying numbers of clusters ($\{5, 10, 15, 20, 25\}$) with both Euclidean and cosine distance on the embeddings of all strokes in the test data. This leads to a total of 20 different clustering results. 
In \Table{embedding_table}, we report the average SC across these 20 clustering experiments for a number of different model configurations along with the Chamfer distance (CD) for stroke reconstruction and prediction. Note, the Pearson correlation between the SC and the prediction accuracy is 0.92 indicating a strong correlation between the two (see \Sec{sc} in supplementary).

\paragraph{Influence of the embedding dimensionality ($D$)}
We performed experiments with different values of $D$ -- the dimensionality of the latent codes. \Table{embedding_table} shows that this parameter directly affects all components of the task: While a high-dimensional embedding space improves reconstructions accuracy, it is harder to predict valid embeddings in such a high-dimensional space and in consequence both the prediction performance and SC deteriorate. We observe a similar pattern with sequence-to-sequence architectures which benefit most from the increased embedding capacity by achieving the lowest reconstruction error (Recon. CD for seq2seq, D=32). However, it also leads to a significantly higher prediction error. Higher-dimensional embeddings result in less compact representation space, making the prediction task more challenging.

\begin{figure}[t]
\centering\small
\includegraphics[width=.32\columnwidth, trim={0pt 0pt 0pt 0pt}]{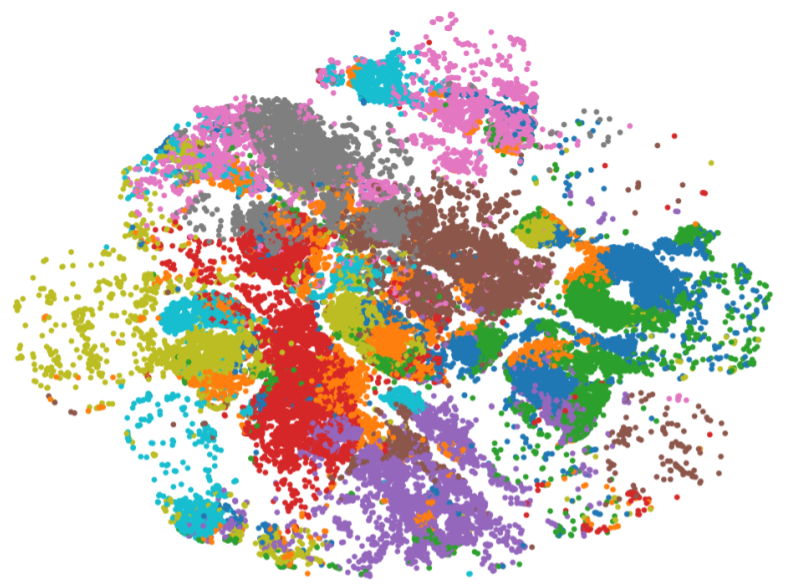}
\includegraphics[width=.32\columnwidth, trim={0pt 0pt 0pt 0pt}]{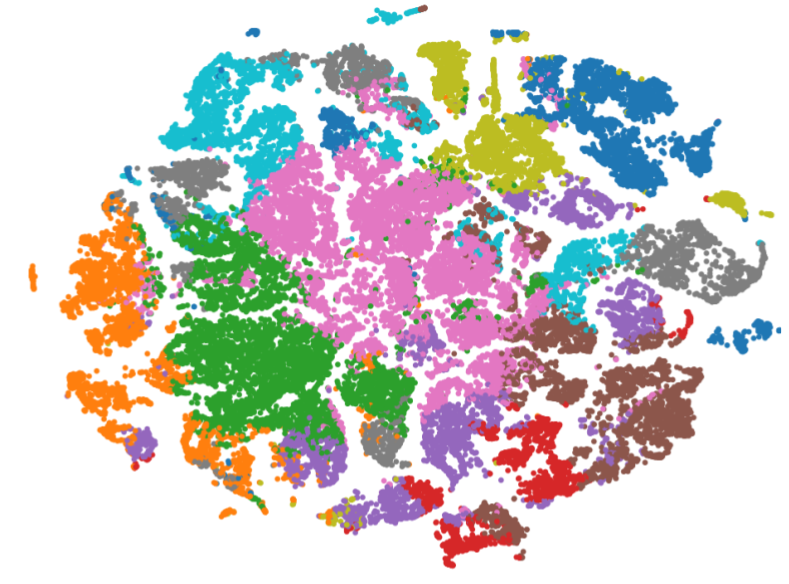}
\includegraphics[width=.32\columnwidth, trim={0pt 0pt 0pt 0pt}]{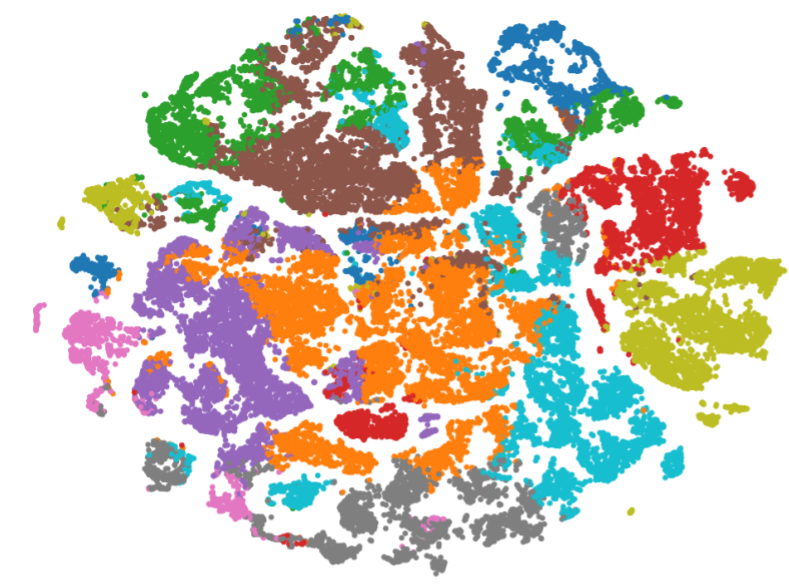}\\
\rule{0pt}{0pt}\hfill \modelemb-VAE, SC=$0.191$ \hfill\hfill \modelseq, SC=$0.270$\hfill\hfill \modelemb, SC=$0.322$\hfill\rule{0pt}{0pt}
\vspace{.1in}
\caption{\textbf{tSNE Embedding} -- Visualization of the latent spaces for different models (for quantitative analysis see \Table{embedding_table}). 
We employ $k$-means in latent space ($k{=}10$), and color by cluster ID. While a VAE regularized objective leads to an overall compact latent-space, clusters are not well separated, ours produces the most compact clusters (from left to right) which we show to be correlated with prediction quality. 
}
\label{fig:new_tsne}
\end{figure}

\begin{table*}[b]
\begin{minipage}[t]{0.42 \textwidth}
\strut\vspace*{-.1\baselineskip}\newline
\small
\caption{\textbf{Embedding space analysis} --~
(Top) Variants of our model with different embedding dimensionalities and a variant of our model with VAE.
(Bottom) Results for a sequence-to-sequence stroke autoencoder (seq2seq) and its variational (VAE) and/or auto-regressive (AR) variants. All stroke embedding models use our Transformer relational model $\relational$.
$D$ indicates the dimensionality of the embedding space. CD and SC denote Chamfer Distance and Silhouette Coefficient, respectively.}
\label{table:embedding_table}
\end{minipage}
\hfill
\begin{minipage}[t]{0.55\textwidth}
    \strut\vspace*{-.5\baselineskip}\newline
    \strut\hfill
    \renewcommand{\arraystretch}{0.9}
    \setlength\tabcolsep{2.5pt}%
    \small
    \begin{tabular}{lHrccc}
    $\encoder / \decoder$                 & $\relational$ & D  & Recon.\ CD $\downarrow$ & Pred. CD$\downarrow$ & SC $\uparrow$  \\
    \toprule                                                                 
    \modelemb-$\encoder$/$\decoder$~(\Tab{res_table})    & TR            & 8   & 0.0136                  & \bf 0.0442           & \bf 0.361     \\
    \modelemb-$\encoder$/$\decoder$                             & TR            & 16  & 0.0091                  & 0.0481               & 0.335         \\
    \modelemb-$\encoder$/$\decoder$                             & TR            & 32  & 0.0081                  & 0.0511               & 0.314         \\
    \midrule
    \modelemb-$\encoder$/$\decoder$-VAE                         & TR            & 8   & 0.0198                  & 0.0953               & 0.197         \\
    \midrule
    \midrule
    \modelseq~(\Tab{res_table}) & TR            & 8   & 0.0138                  & 0.0540               & 0.276         \\
    \modelseq & TR            & 16   & 0.0076                  & 0.0783               & 0.253         \\
    \modelseq & TR            & 32   & \textbf{0.0047}                  & 0.0848               & 0.261         \\
    \midrule
    \modelseq-VAE                         & TR            & 8   & 0.0161                  & 0.0817               & 0.180         \\
    \modelseq-AR                          & TR            & 8   & 0.0432                  & 0.0855               & 0.249         \\
    \modelseq-AR-VAE                      & TR            & 8   & 0.2763                  & 0.1259               & 0.151         \\
    \bottomrule
    \end{tabular}
\end{minipage}
\end{table*}

\paragraph{Architectural variants}
In order to obtain a smoother latent space, we also introduce a KL-divergence regularizer~\cite{kingma2013auto} and follow the same annealing strategy as Ha et al.~\cite{ha2017neural}. 
It is maybe surprising to see that a VAE regularizer (line \modelemb-VAE) \textit{hurts} reconstruction accuracy and interpretability of the embedding space. 
Note that the prediction task does not require interpolation or latent-space walks since latent codes represent entire strokes that can be combined in a discrete fashion.    
The results further indicate that our architecture yields a better behaved embedding space, while retaining a good reconstruction accuracy. This is indicated by
i) the increase in reconstruction quality with larger $D$ yet prediction accuracy and SC deteriorate;
ii) \modelemb obtains much better prediction accuracy and SC at similar reconstruction accuracy;
iii) smoothing the embedding space using a VAE for regularization hurts reconstruction accuracy, prediction accuracy and SC;
iv) autoregressive approach hurts reconstruction and prediction accuracy and SC - this is because autoregressive models tend to overfit to the ground-truth data (i.e., teacher forcing) and fail when forced to complete drawings based on their own predictions. 

\paragraph{Visualizations}
To further analyse the latent space properties, we provide a t-SNE visualization~\cite{maaten2008visualizing} of the embedding space with color coding for cluster IDs as determined by $k$-means with $k=10$ in \Figure{new_tsne}. 
The plots indicate that the VAE objective encourages a latent space with overlapping clusters, whereas for \modelemb, the clusters are better separated and more compact. 
An interesting observation is that the smooth and regularized VAE latent space does not translate into improved performance on either reconstruction or inference, which is inline with prior findings on the connection of latent space behavior and down-stream behavior~\cite{locatello2018challenging}.
Clearly, the embedding spaces learned using a \modelemb model have different properties and are more suitable for predictive tasks that are conducted in the embedding space. This qualitative finding is inline with the quantitative results of the SC and correlating performance in the stroke prediction task (see also \Sec{embedding_pred} in supplementary). 

\begin{figure}[b]
\centering\small
\includegraphics[width=0.8\columnwidth, trim={0pt 0pt 0pt 0pt}]{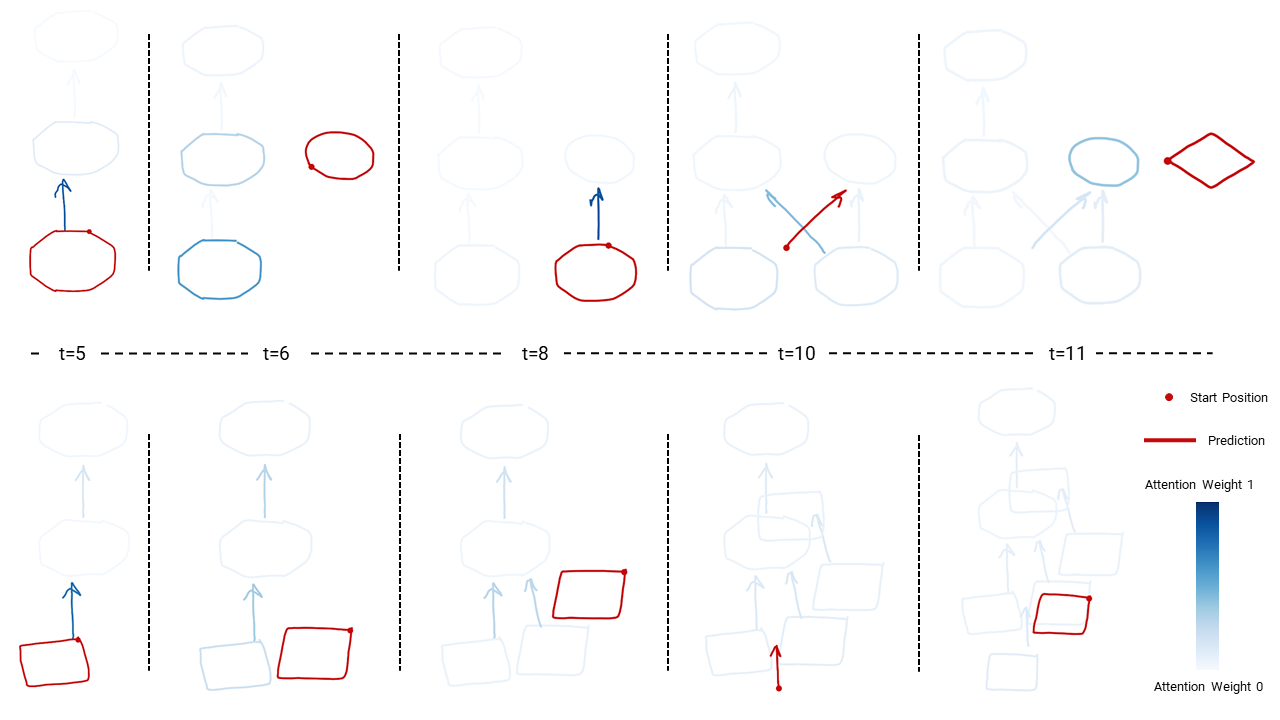}
\caption{{\bf Attention visualization over time} --
(top) with and (bottom) without conditioning on the start position to make a prediction for the next stroke's (in red) embedding. Attention weights correspond to the average of all attention layers across the network.}
\label{fig:attention_ablation}
\end{figure}

\subsection{Ablations}
\label{Sec:ablations}

\paragraph{Conditioning on the start position}
The factorization in Eq. \ref{eq:stroke_pos_factorization} allows our model to attend to a relatively local context. To show the importance of conditioning on the initial stroke positions, we train a model without this conditioning. Fig. \ref{fig:attention_ablation} shows that conditioning on the start position helps to attend to the local neighborhood, which becomes increasingly important as the number of strokes gets larger. Moreover, the Chamfer Distance on the predictions nearly double from 0.0442 to 0.0790 in the absence of the starting positions.

\begin{wrapfigure}{R}{0.5\textwidth}
\vspace{0.1em}
\centering
\includegraphics[width=0.95\linewidth, fbox, trim={0pt 0pt 0pt 0pt}]{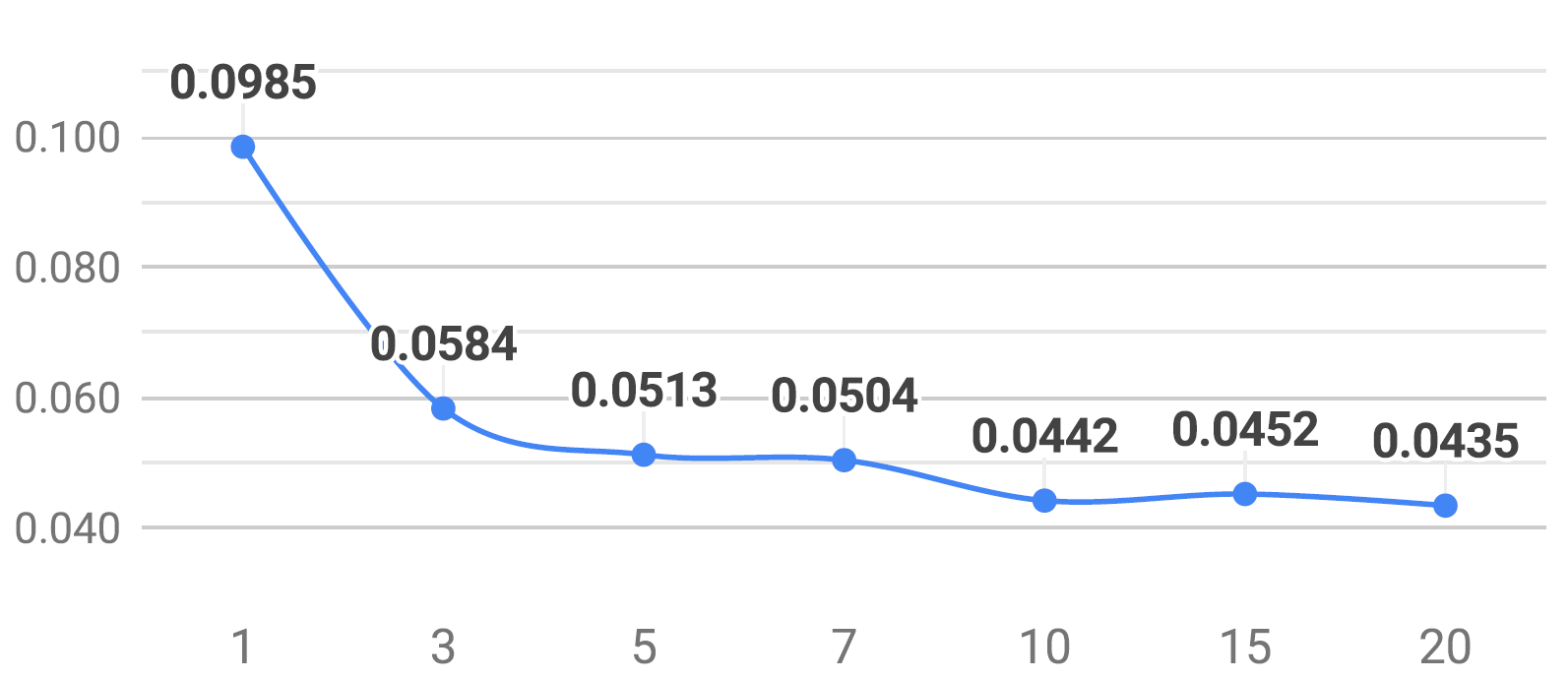}
\vspace{.5em}
\caption{Pred. CD performance of our model $\relational$ by using different number of components in the GMM for embedding predictions. }
\label{fig:gmm_k_ablation}
\vspace{-1em}
\end{wrapfigure}

\textbf{Number of GMM components.} Using a multi-modal distribution to model the stroke embedding predictions significantly improves our model's performance (cf. Fig. \ref{fig:gmm_k_ablation}). It is an important hyper-parameter as it is the only source of stochasticity in our relational model $\relational$. We observe that using $10$ or more components is sufficient. Our results presented in the paper are achieved with 10 components.

\textbf{Back-propagating $\relational$ gradients.} Since we aim to decouple the local stroke information from the global drawing structure, we train the embedding model \modelemb-$\encoder$/$\decoder$ via the reconstruction loss only, and do not back-propagate the relational model’s gradients. We hypothesize that doing so would force the encoder to use some capacity to capture global semantics. When training our best model with all gradients flowing to the encoder \modelemb-$\encoder$, increases the reconstruction error (Recon. CD) from 0.0136 to 0.0162 and the prediction error (Pred. CD) from 0.0442 to 0.0470.

\subsection{Qualitative Results}
The quantitative results from \Table{res_table} indicate that our model performs better in the predictive modelling of complex diagrams compared to the baselines. \Fig{sketchrnn} provides further indication that this is indeed the case. We show predictions of SketchRNN~\cite{ha2017neural} which performs well on structures with very few strokes but struggles to predict more complex drawings. In contrast ours continues to produce meaningful predictions even for complex diagrams. 
This is further illustrated in \Fig{gallery} showing a number of qualitative results from a model trained on the \didi (left), \iamondb (center) and the \quickdraw (right) datasets, respectively. Note that all predictions are in the auto-regressive setting, where only the first stroke (in light blue) is given as input. All other strokes are model predictions (numbers indicate steps).

%% file: fig/res_table.tex
\begin{table*}[b]
\centering
\begin{minipage}[t]{0.38\textwidth}
\strut\vspace*{-.5\baselineskip}\newline
\caption{\textbf{Stroke prediction} -- 
We evaluate reconstruction (\ie $\text{CD}(\decoder(\encoder(\stroke)), \stroke)$ and prediction 
(\ie $\text{CD}(\relational(\lambda_{<k}), \stroke_k)$ for a number of different models.
Note that performing well on reconstruction does not necessarily correlate with good prediction performance.}

\label{table:res_table}
\end{minipage}
\hfill
\begin{minipage}[t]{0.6\textwidth}
\strut\vspace*{-.5\baselineskip}\newline
\small
\begin{tabular}{cccc}
$\encoder / \decoder$ & $\relational$ & Recon. CD$\downarrow$ & Pred. CD$\downarrow$ \\
\toprule
\modelseq                                   & RNN                                & 0.0144                & 0.0794          \\
\modelseq                                   & \modelemb-$\relational$            & 0.0138                & 0.0540           \\
\midrule
\modelemb-$\encoder$/$\decoder$             & RNN                                & 0.0139                & 0.0713           \\
\modelemb-$\encoder$/$\decoder$             & \modelemb-$\relational$~(Ord.)     & 0.0143                & 0.0696           \\
\modelemb-$\encoder$/$\decoder$             & \modelemb-$\relational$            & \textbf{0.0136}                & \textbf{0.0442}  \\ 
\midrule  
% \hline
\multicolumn{2}{c}{\sketchrnn Decoder \cite{ha2017neural}}   & N/A & 0.0679        \\ 
\bottomrule
\end{tabular}
\end{minipage}
\end{table*}

%% file: 8_conclusion.tex
% !TEX root = ../main.tex

\begin{figure}[h]
\centering
{
\begin{minipage}[t]{0.50\textwidth}
\strut\vspace*{-\baselineskip}\newline
\newcommand{\diagram}[1]{\includegraphics[width=.25\linewidth,height=.3\linewidth,keepaspectratio=true,valign=c]{fig/fig6/diagrams/#1}}
\newcommand{\diagramw}[1]{\includegraphics[width=.38\linewidth,height=.3\linewidth,keepaspectratio=true,valign=c]{fig/fig6/diagrams/#1}}
\newcommand{\hww}[1]{\includegraphics[width=.6\linewidth,height=1.2\linewidth,keepaspectratio=true,valign=c]{fig/#1}}
\diagram{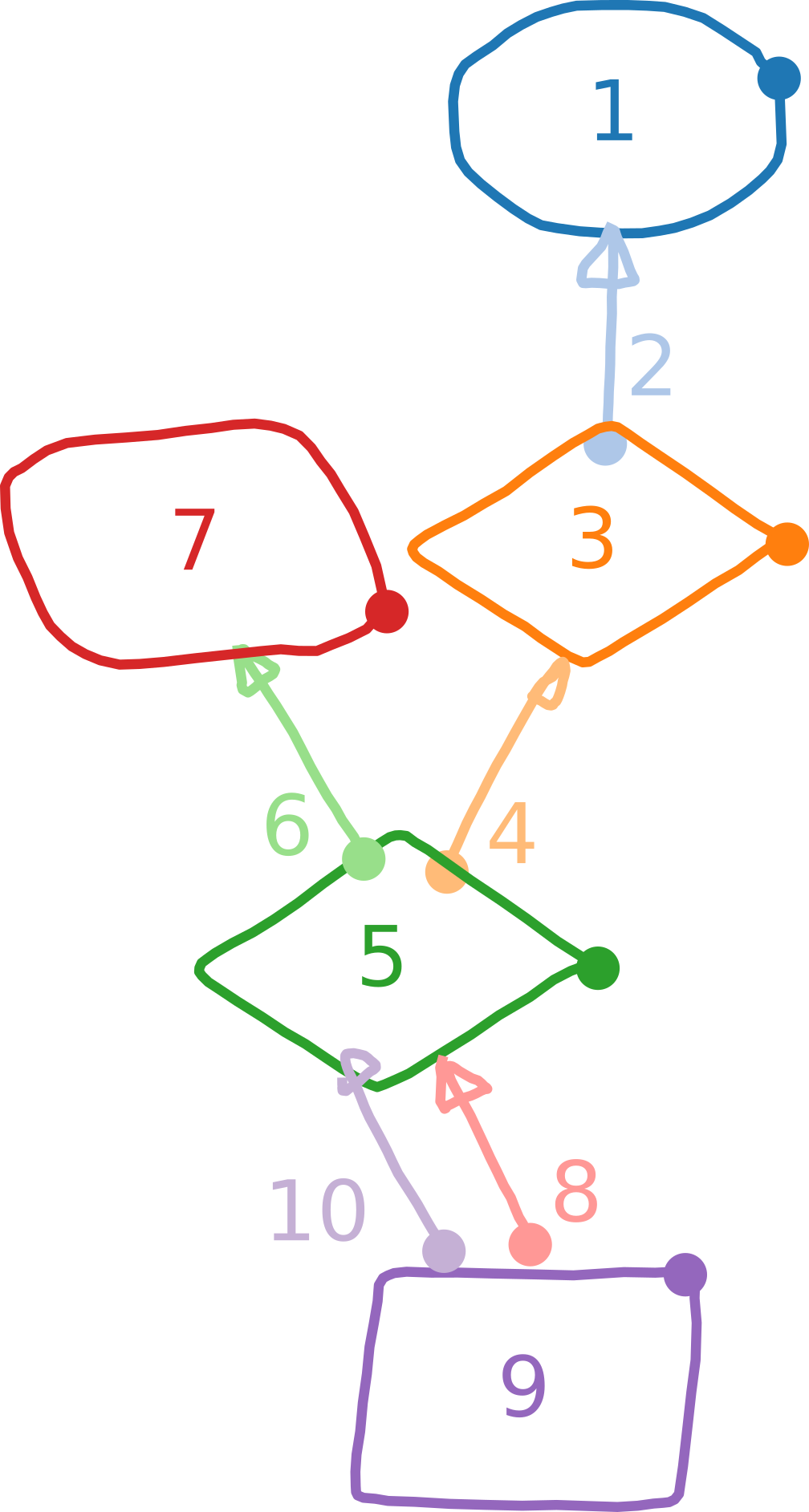}
\diagram{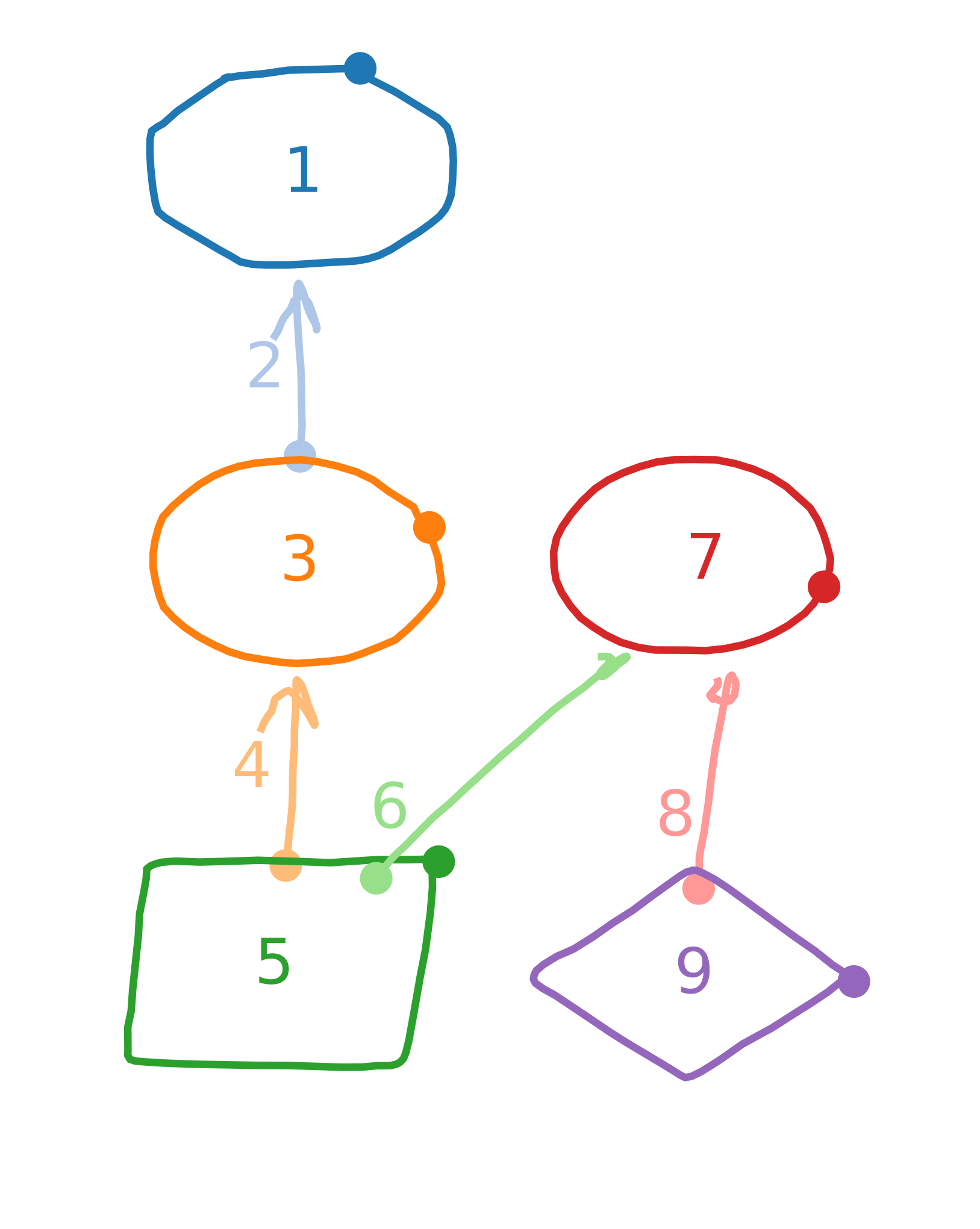}
\diagram{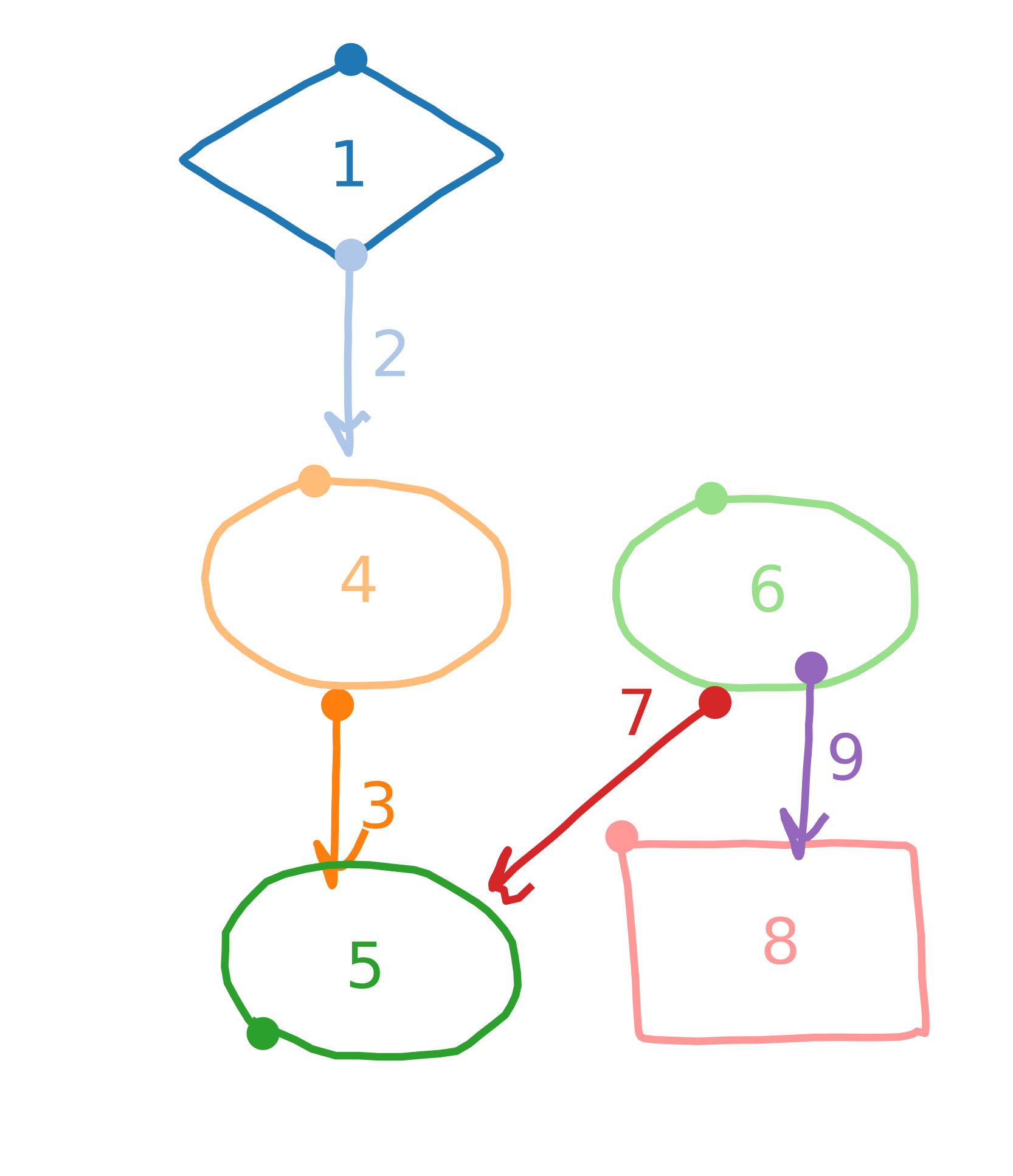}
%\diagram{64_pos_ar_heatmap_ordered_s9}
\diagram{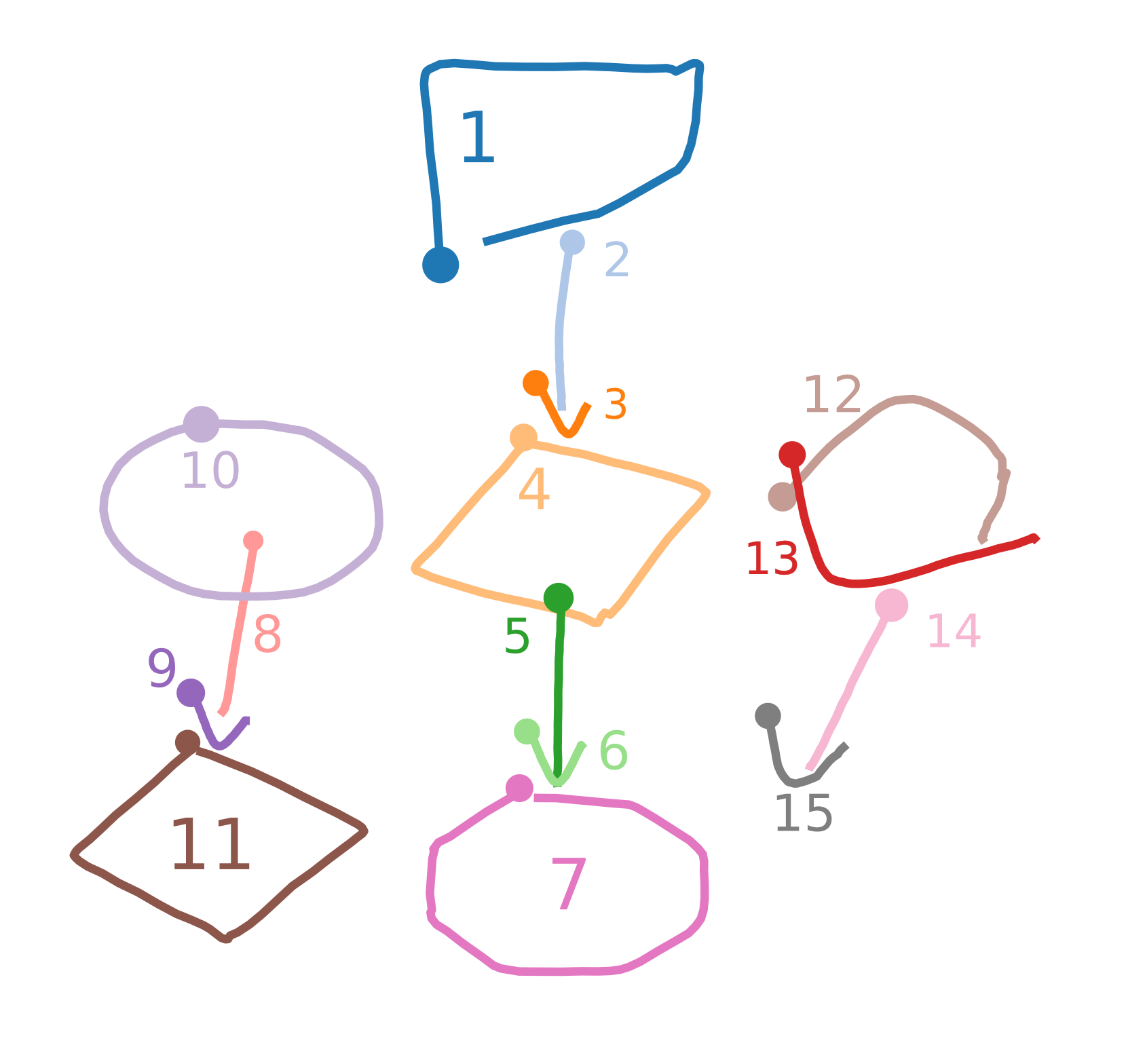}\\
%\diagram{set3_64_pos_ar_heatmap_ordered_s9}\\
\diagramw{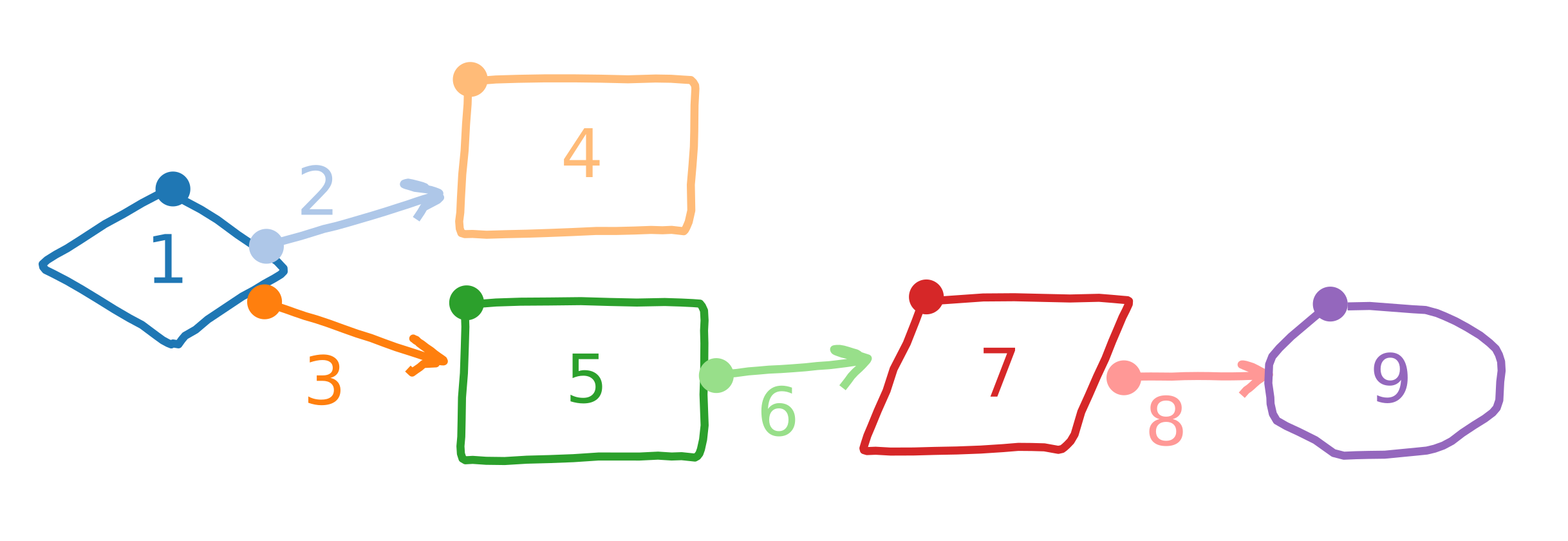} 
\hww{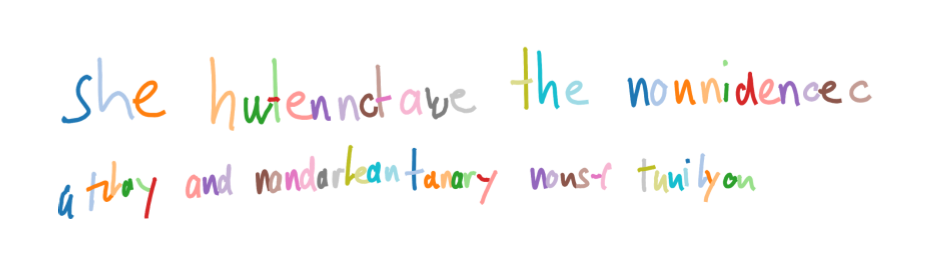} \\
%\diagramw{73_pos_ar_heatmap_ordered_s8}
%\diagramw{9_pos_ar_heatmap_ordered_s11}
%\diagramw{15_pos_ar_heatmap_ordered_s5} \\
\diagramw{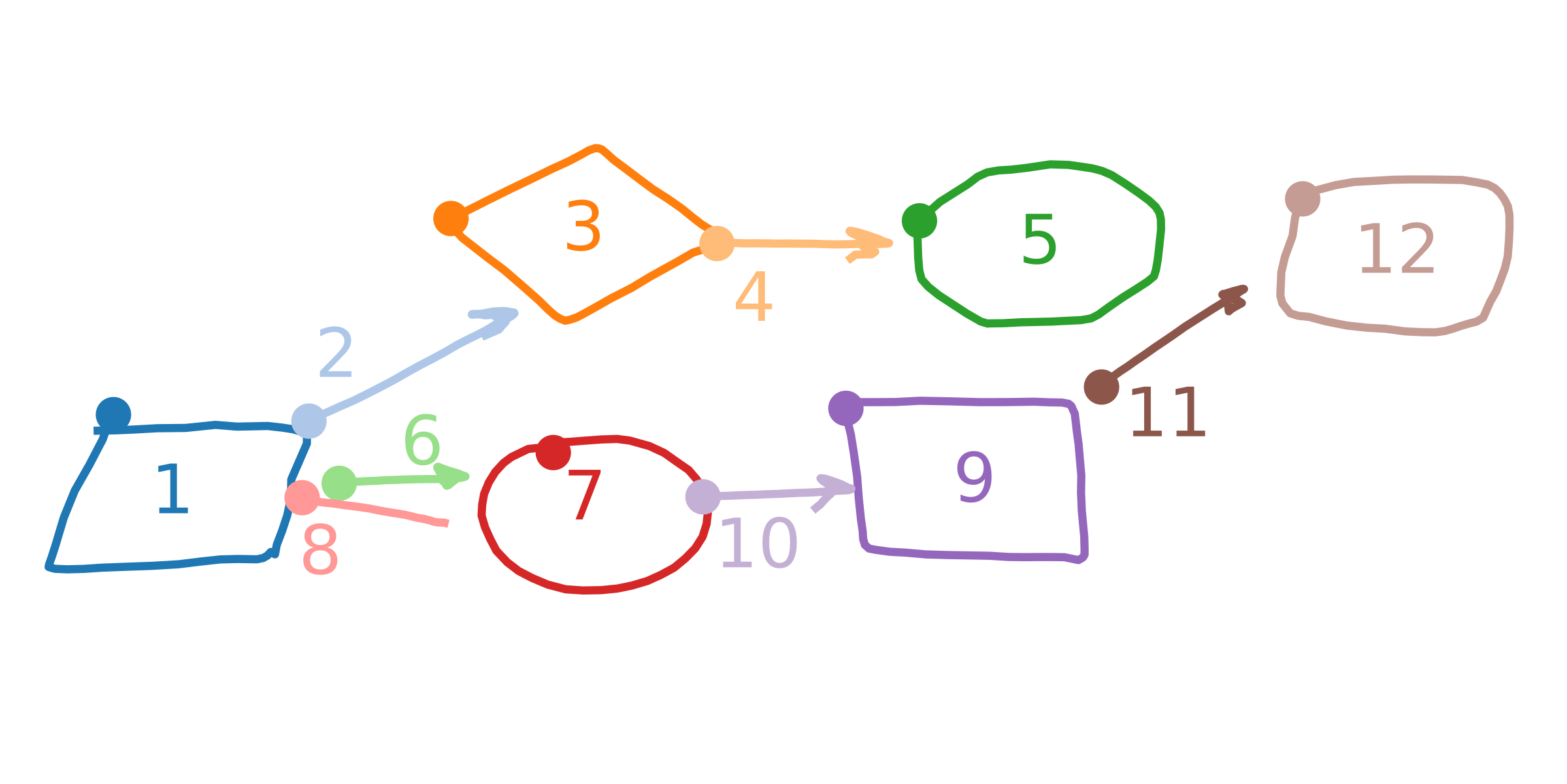}
%\diagramw{21_pos_ar_heatmap_ordered_s9} 
%\diagramw{4_pos_ar_heatmap_ordered_s6}
% \diagram{214_pos_ar_heatmap_ordered_s12}
% \diagram{236_pos_ar_heatmap_ordered_s8}
\hww{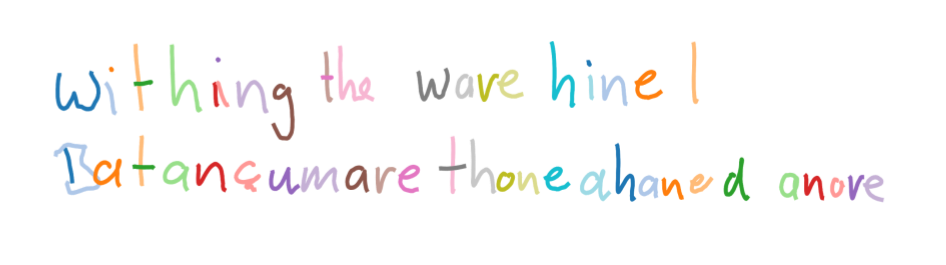}
\end{minipage}}~%
{
\begin{minipage}[t]{0.24\textwidth}
\strut\vspace*{-1.\baselineskip}\newline
\newcommand{\cat}[1]{\includegraphics[width=.48\linewidth,height=.48\linewidth,keepaspectratio=true]{fig/fig6/cats/#1.png}}
\cat{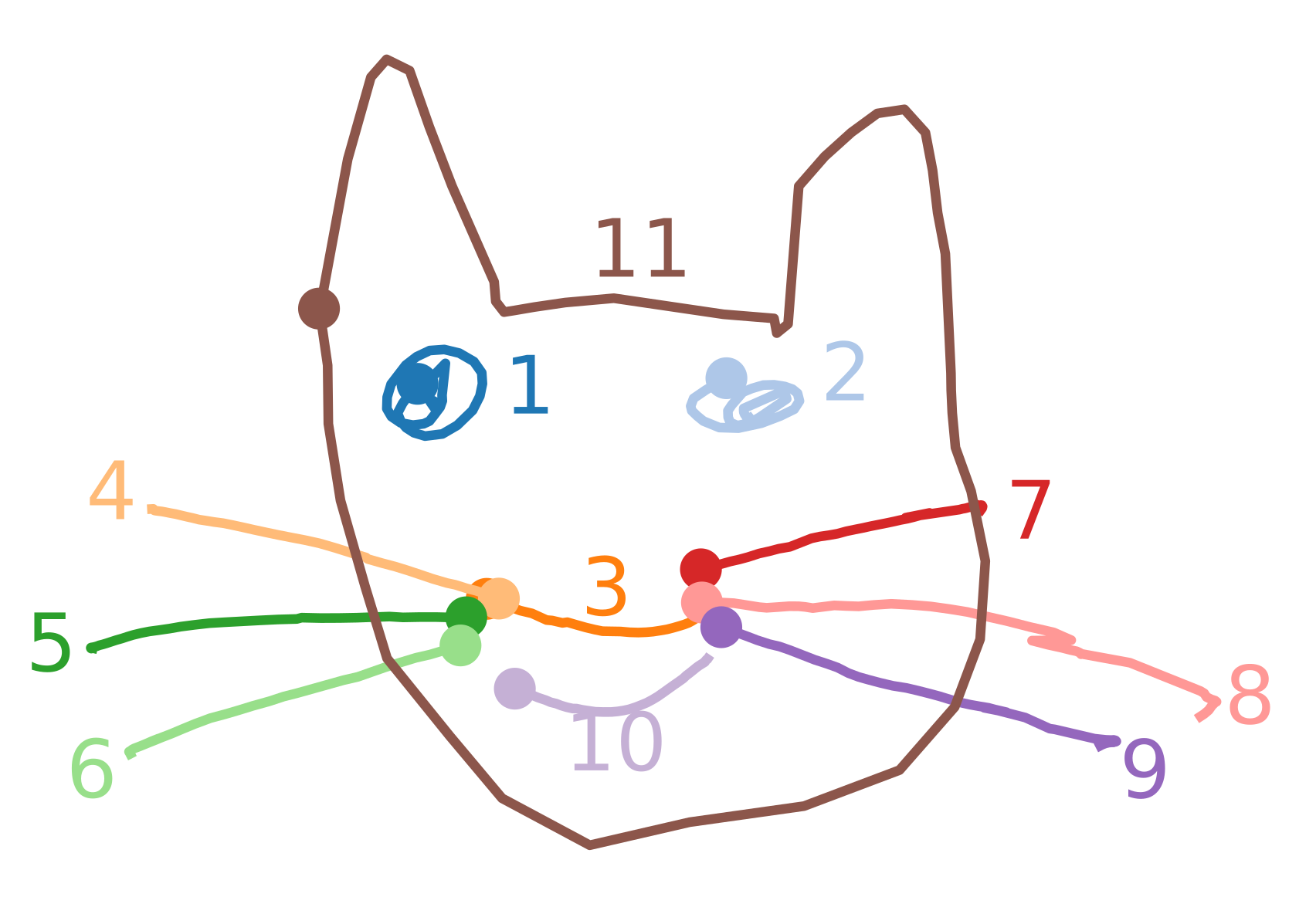}
\cat{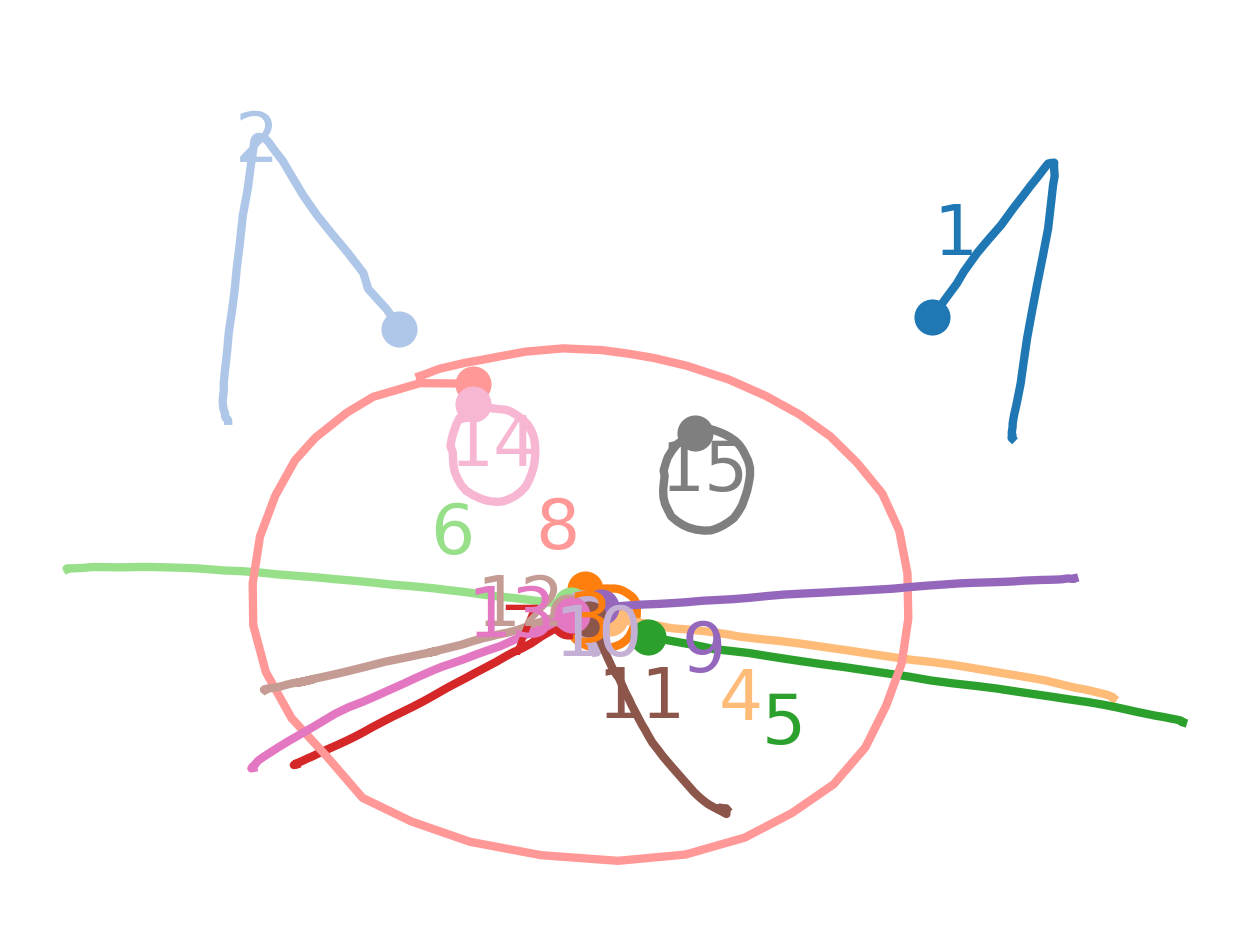}
\cat{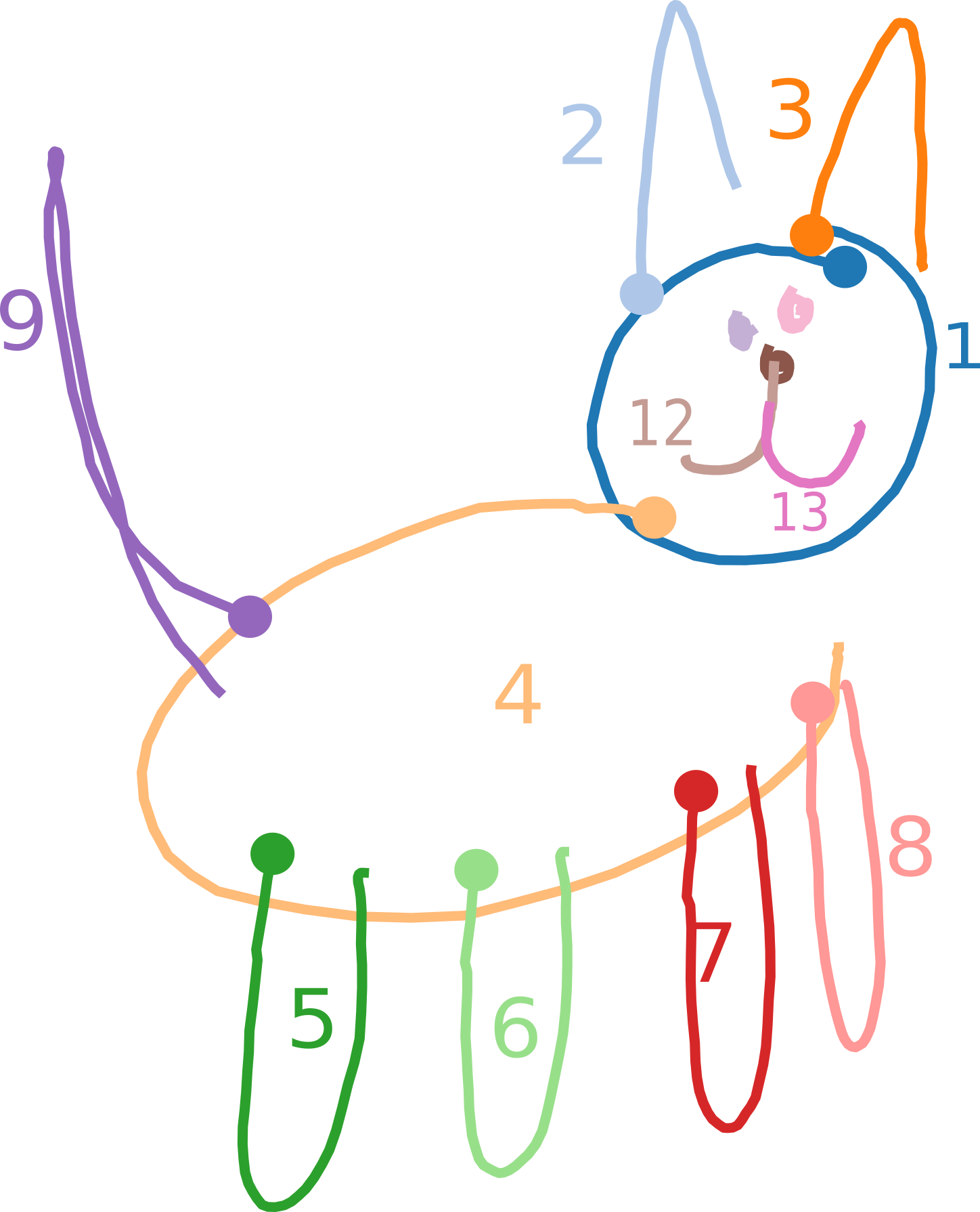}
\cat{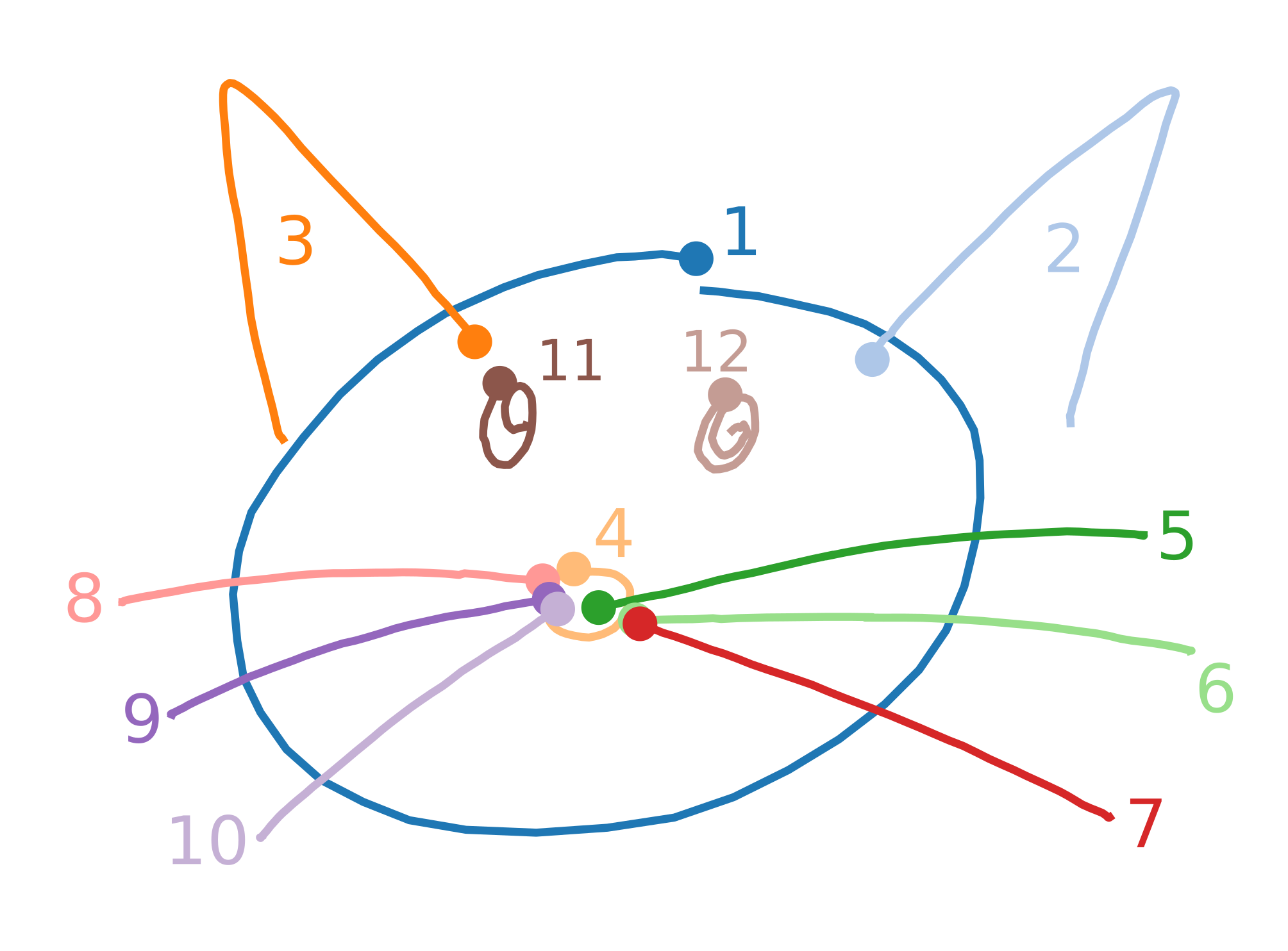}
\cat{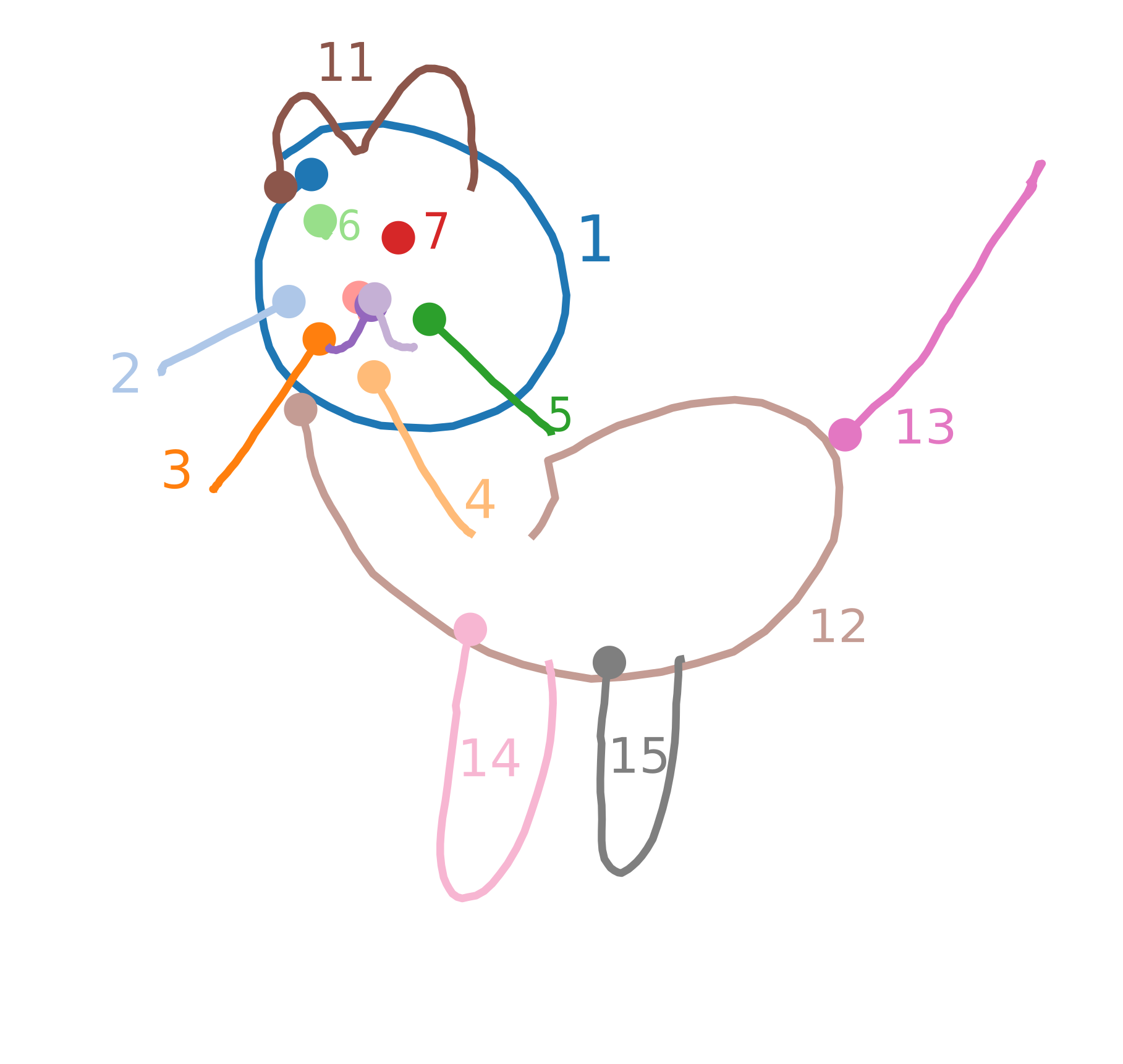}
\cat{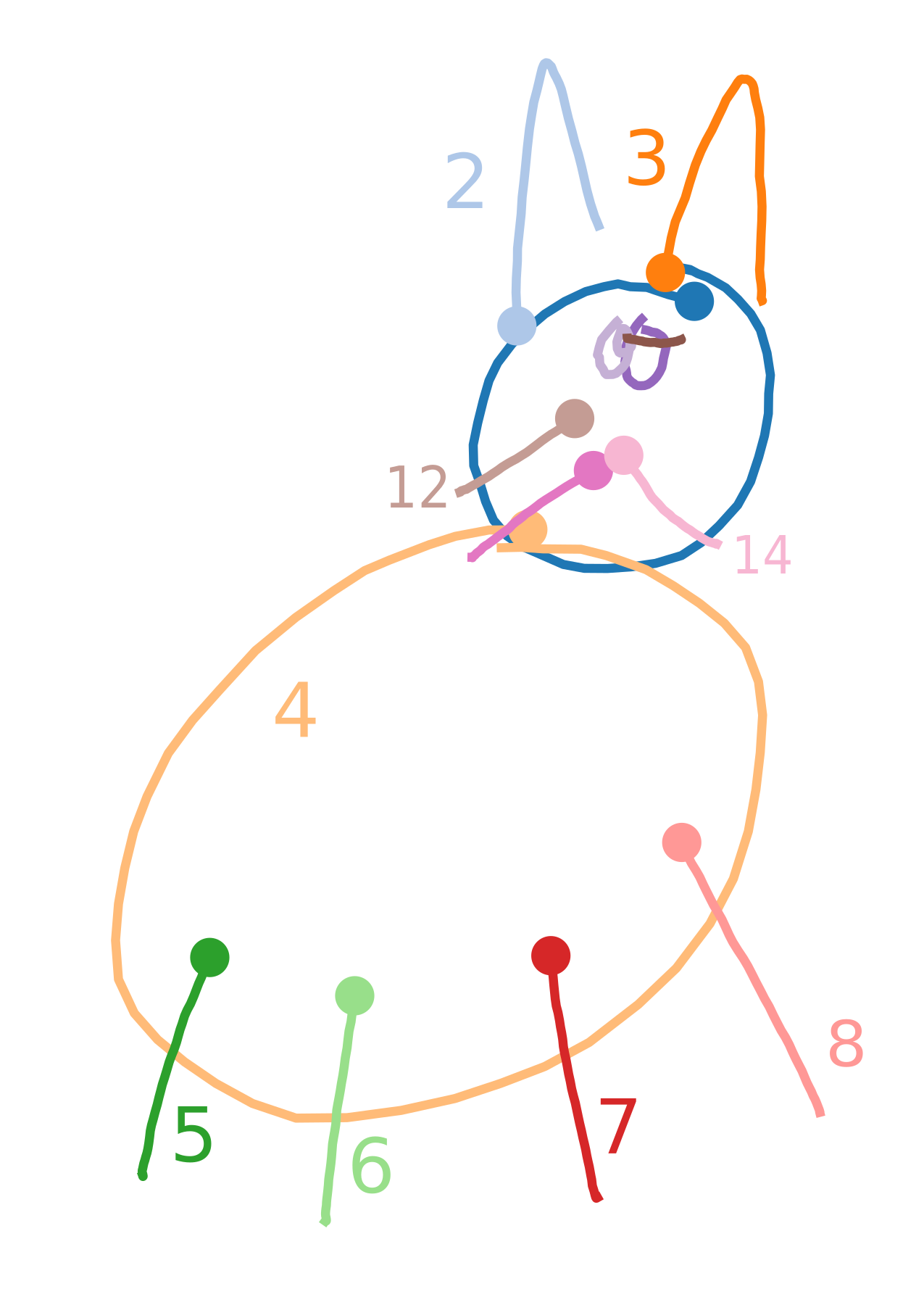}
\end{minipage}}~%
\begin{minipage}[t]{0.24\textwidth}
\strut\vspace*{-1.\baselineskip}\newline
\newcommand{\elephant}[1]{\includegraphics[width=.48\linewidth,height=.48\linewidth,keepaspectratio=true]{fig/fig6/elephants/#1.png}}
\elephant{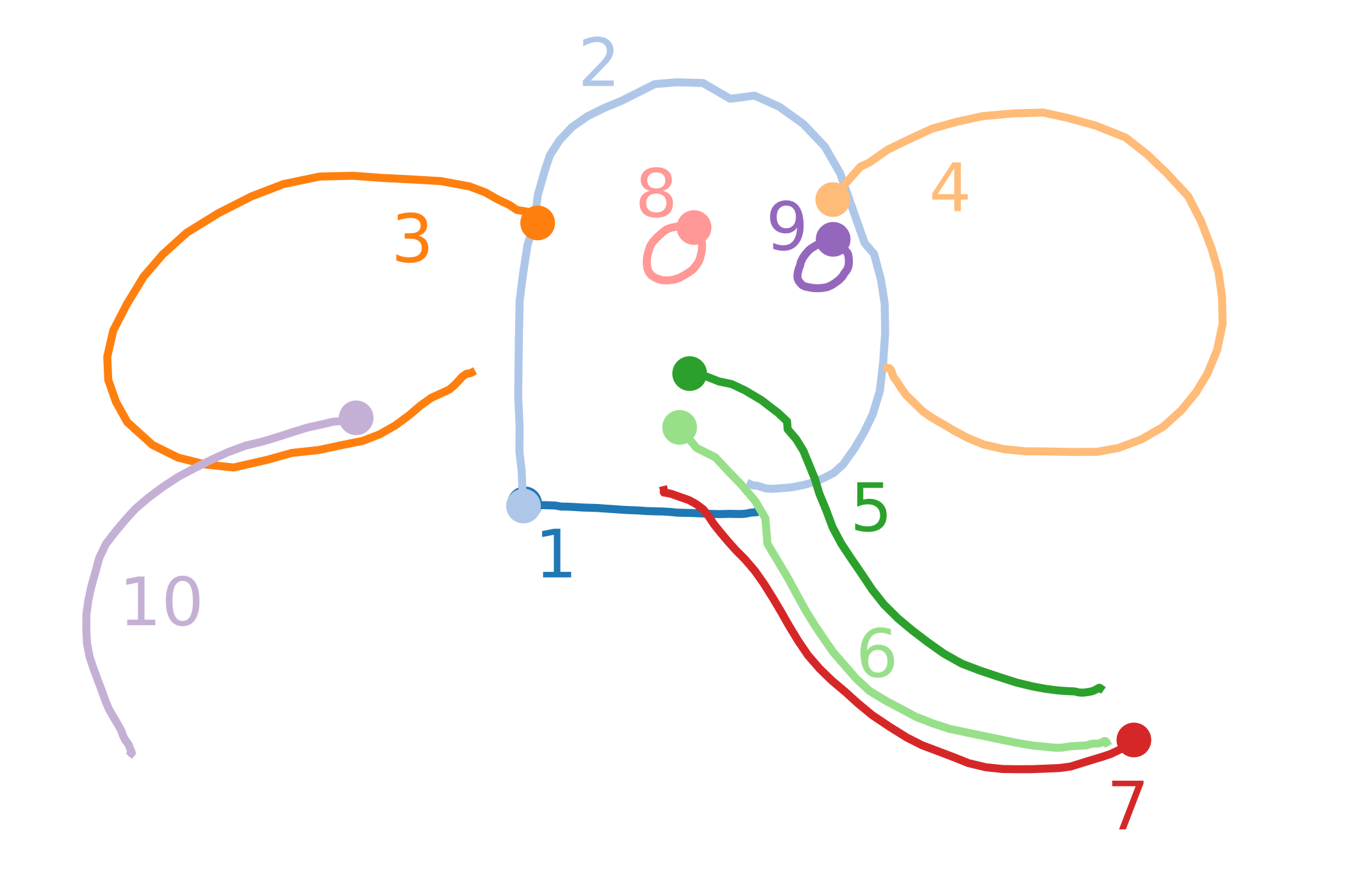}
\elephant{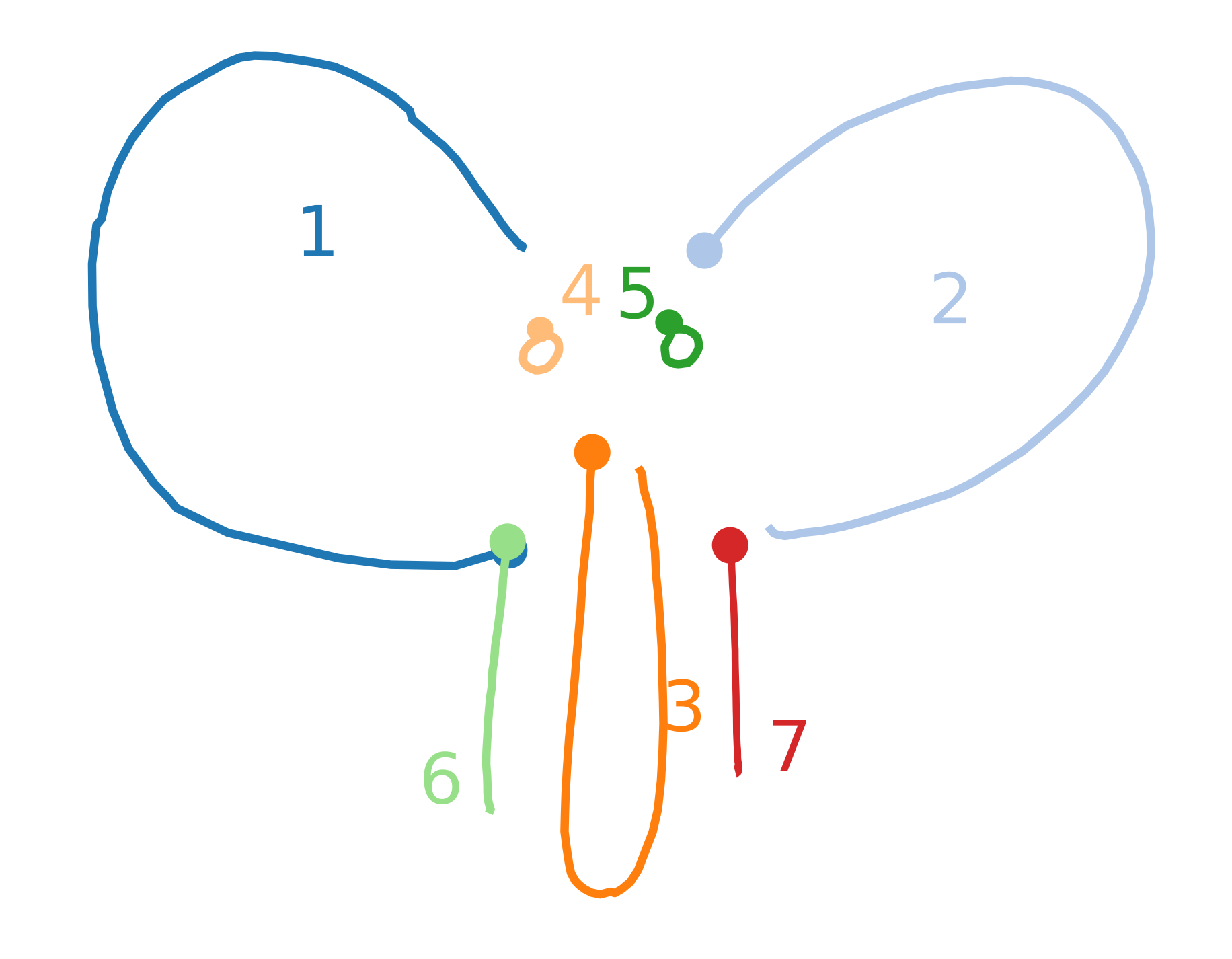}
\elephant{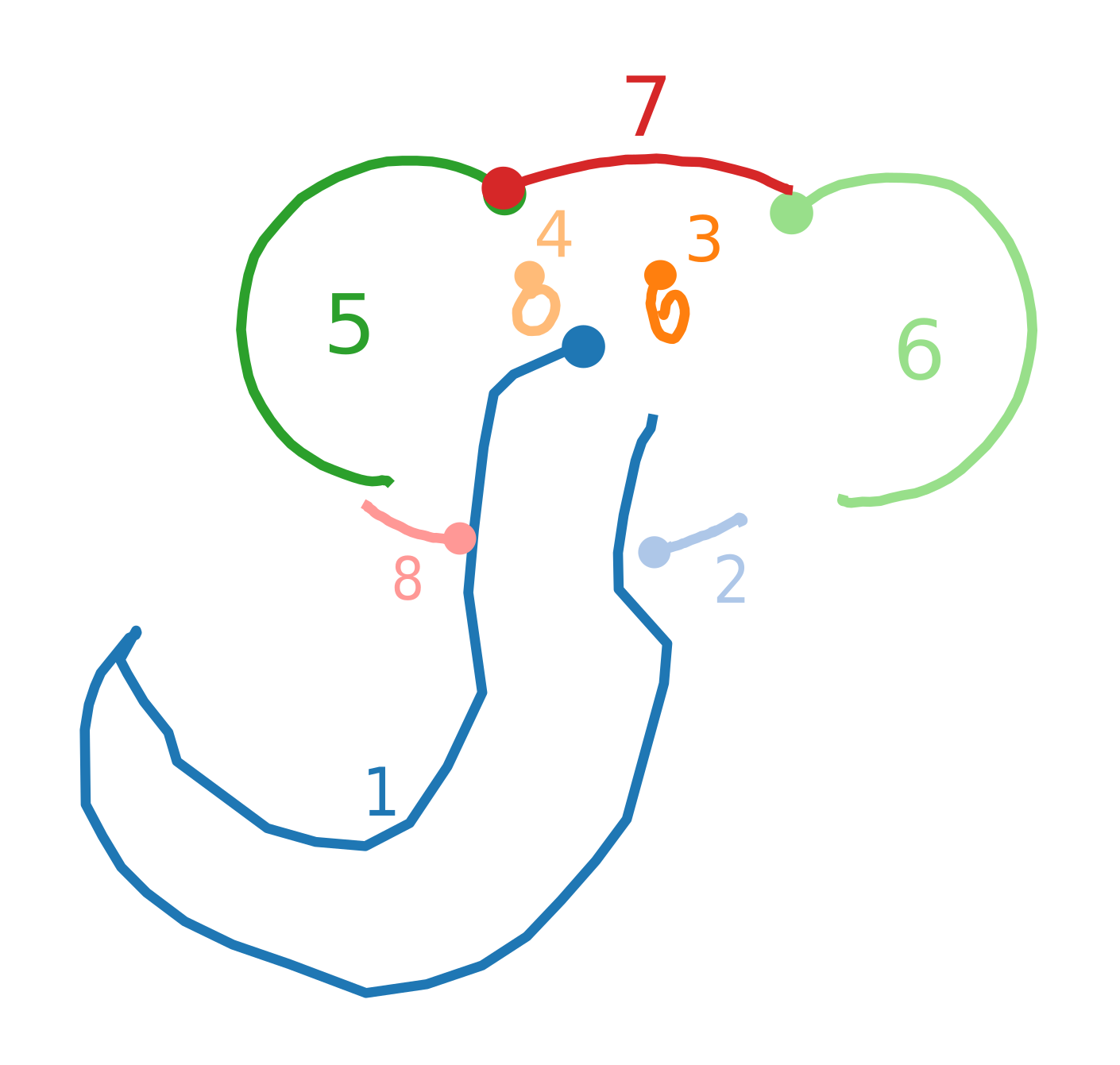}
\elephant{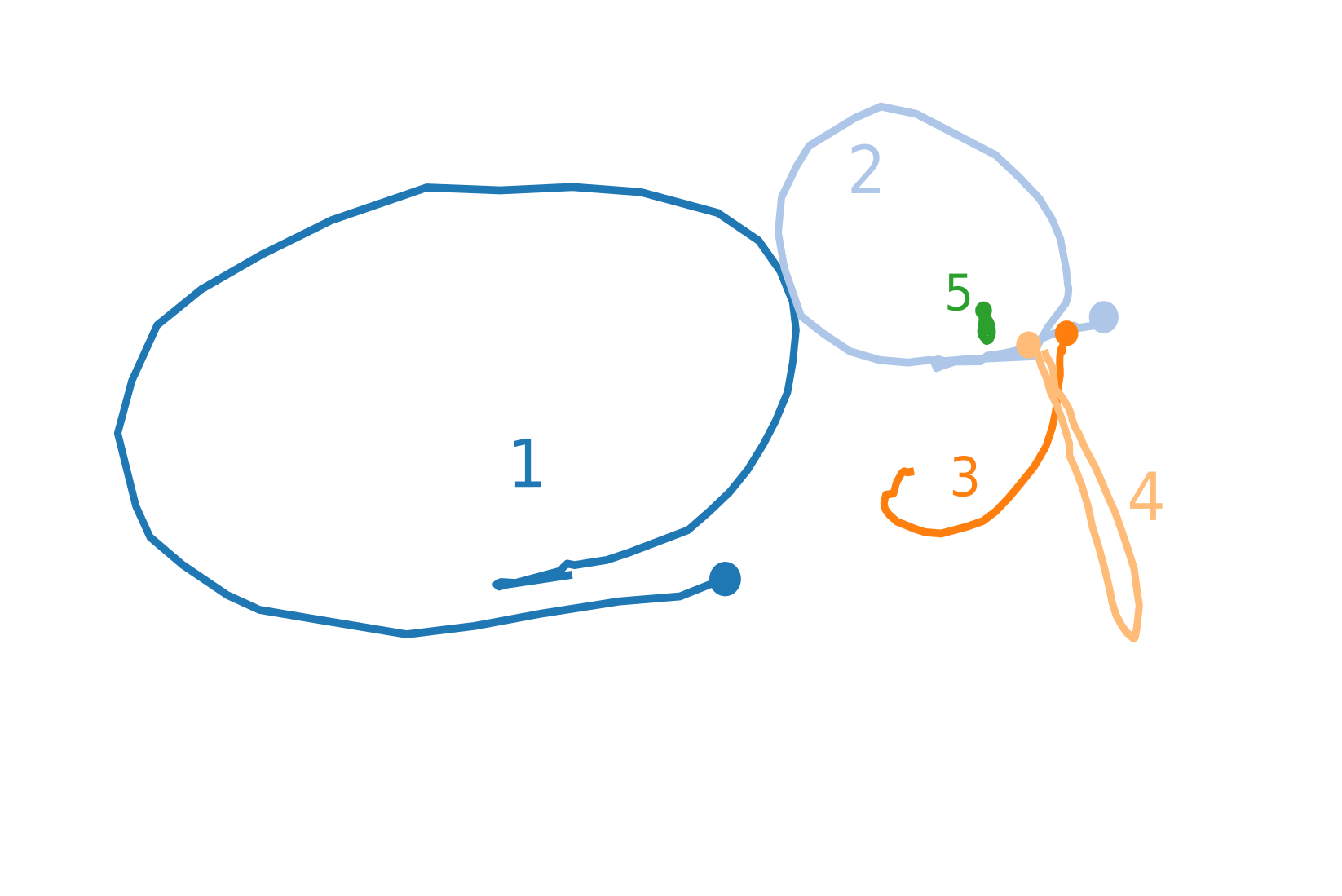}
\elephant{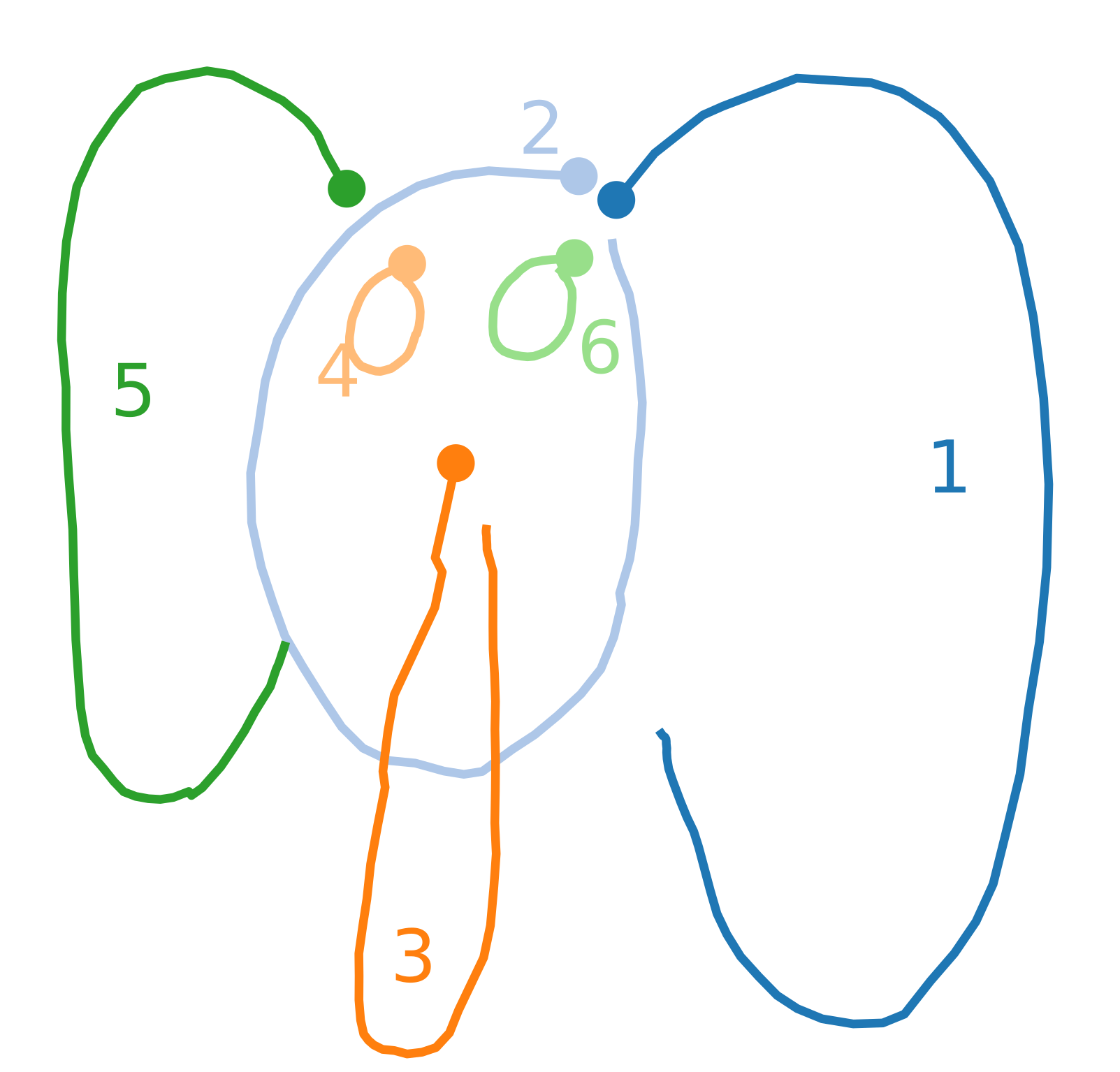}
\elephant{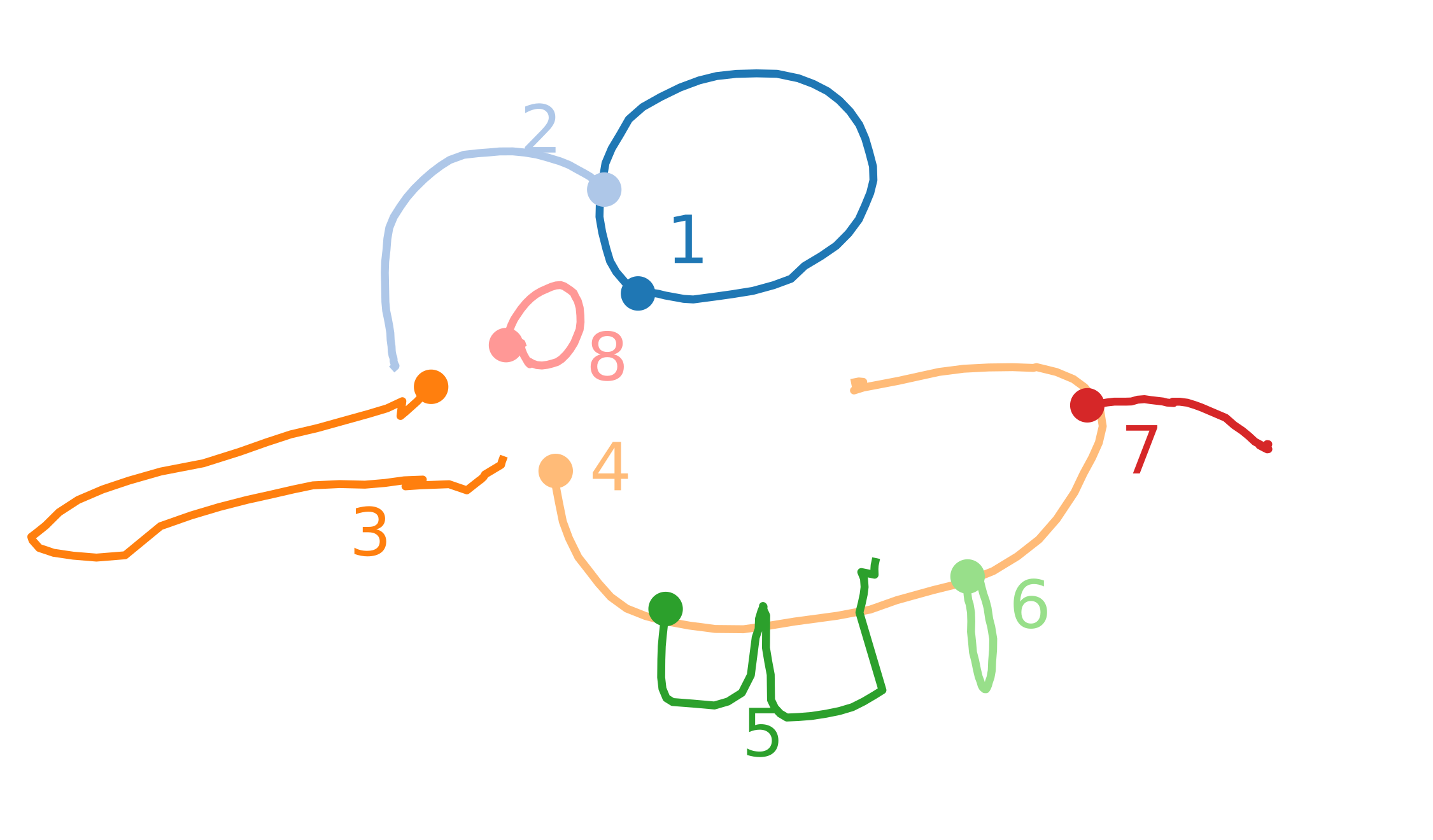}
\end{minipage}
\vspace{0em}
\caption{{\bf Qualitative examples from \modelemb} -- Drawings were sampled from the model given the \textcolor{firststroke}{\textbf{first stroke}}. 
}
\label{fig:gallery}
\end{figure}

\section{Conclusions}
We have presented a \emph{compositional generative model} for stroke data. 
In contrast to previous work, our model is able to model complex free-form structures such as those that occur in hand-drawn diagrams. A key insight is that by ignoring the ordering of individual strokes the complexity induced by the compositional nature of the data can be mitigated. 
This is achieved by decomposing our model into a novel auto-encoder that is able to create qualitatively better embedding spaces for predictive tasks and a relational model that operates on this embedding space.
We demonstrate experimentally that our model outperforms previous approaches on complex drawings and provide evidence that 
a) good reconstruction accuracy is not necessarily indicative of good predictive performance
b) the embedding space that our model learns strikes a good balance between these two and behaves qualitatively differently than the embedding spaces learned by the baselines models. We believe that in future work important concepts of our work can be directly applied to other tasks that have a compositional nature.

%% file: impact_statement.tex
% !TEX root = ../main.tex
\section*{Broader Impact} 
On the broader societal level, this work remains largely academic in nature, and does not pose foreseeable risks regarding defense, security, and other sensitive fields.
One potential risk associated with all generative modelling work, is the danger of creating digital content that can be used for malicious purposes. In the context of stroke-based data the forgery of handwriting and signatures in particular is the most immediate concern. However, an attacker would require sufficient training data and extrapolation from such data would remain challenging. 
Furthermore, automating tasks that are commonly associated with creativity or craftsmanship holds some danger of rendering jobs or entire occupations redundant. However, we do believe that the positive impact of this work, which is to build technology that allows for better interaction between humans and computers by enabling for a more immediate and natural way to create structured drawings, will have a larger positive impact, such as improved creativity and better communication channels between humans.

% PS: Also, Emre needs to finish his PhD.

%% file: 9_appendix.tex
\clearpage
\title{\mytitle \\ (Supplementary Material)}
\author{%
    Emre Aksan\\ETH Zurich\\\texttt{eaksan@inf.ethz.ch} 
    \And 
    Thomas Deselaers\\ Apple Switzerland\\\texttt{deselaers@gmail.com}
    \AND
    Andrea Tagliasacchi\\Google Research\\\texttt{atagliasacchi@google.com}
    \And
    Otmar Hilliges\\ETH Zurich\\\texttt{otmar.hilliges@inf.ethz.ch}
}
\date{}
\settitle
\maketitle

In this supplementary material accompanying the paper \emph{\mytitle}, we provide the following:

\begin{itemize}
    \item additional details about the model parameters and implementation. (\Sec{data}, \Sec{ablationarch});
    \item further description about the correlation between the Silhouette Coefficient and predictive capabilities (\Sec{sc});
    \item an additional visualization of the embedding space (\Sec{embedding_pred});
    \item a discussion of MSE as reconstruction objective (\Sec{appendix_MSE});
    \item a discussion of the limitation of our model (\Sec{limitation});
    \item additional visualization of predictions from our model (\Fig{suppzoo}).
\end{itemize}

In addition we also uploaded a video showing how a user interacts with our proposed model.

\section{Data preparation and augmentation}
\label{sec:data}
The common representation for drawings employed in the related work is $\vectr{x} = \{(x_t, y_t, p_t)\}_{t=1}^T$ -- a sequence of temporarily ordered triplets $(x, y, p)$ of length $T$ corresponding to pen-coordinates on the device screen $(x, y)$ and pen-up events $p$ -- i.e.~$p{=}1$ when the pen is lifted off the screen, and $0$ otherwise~\cite{graves_sequence_data, aksan2018deepwriting}. In \cite{ha2017neural}, the pen-event is extended to triplets of pen-down, pen-up and end-of-sequence events.

We treat a segment of points up to a pen-up event (i.e., $p=1$) as a stroke $\vectr{s}$. Writing a letter or drawing a basic shape without lifting the pen results in a stroke. An ink sample can be considered as a sequence or set of strokes depending on the context. While it is of high importance to preserve the order in handwriting~\cite{google-hwrlstm}, we hypothesize that in free-form sketches, the exact ordering of strokes is not as important.

In this work, we consider strokes as an atomic unit instead of following a point-based representation as in previous work \cite{graves_sequence_data, ha2017neural, aksan2018deepwriting}. We define the ink sample as $\vectr{x} = \{\vectr{s}_i\}_{i=1}^K$ where $K$ is the number of strokes and the stroke $\vectr{s}_i = \{(x_t,y_t)\}_{t=1}^{T_i}$ is a segment of $\vectr{x}$. The stroke length $T_i$ is determined by the pen-up event.

We are using the data pre-processing as proposed by the \didi dataset which consists of two steps: 
1) normalizing the size of the diagram to unit height;
2) resampling the data along time with step size of $20$ ms, meaning a stroke that took 1s to draw will be represented by 50 points.
Each drawing is then stored as a variable length sequence of strokes. In the strokes, we only retain the $(x,y)$ coordinates.

During training, we apply a simple data augmentation pipeline of 3 steps for the drawing sample $\vectr{x}$:

\begin{itemize}
    \item Rotate the entire drawing with a random angle -90\degree and 90\degree.
    \item Scale the entire drawing by a random factor between 0.5 and 2.5
    \item Shear the entire drawing by a random factor between -0.3 and 0.3
\end{itemize}

where each of these steps is executed with a probability of 0.3. 

\section{Architecture Details}
\label{sec:ablationarch}
Our pipeline consists of a stroke embedding model with an encoder (\modelemb-$\encoder$)  and decoder ( \modelemb-$\decoder$) and relational models to predict the stroke position and embeddings (\modelemb-$\relational$). %
In the paper we show results for an ablation experiment where we are replacing the individual components of our model with baselines, \ie, we replace the stroke embedding with a \modelseq model, and we replace the relational model by a standard autoregressive RNN model. In the following we give details on the architectures used.

We aim to keep the number of trainable parameters similar in our experiments. 
All models have $2$-$3$M parameters. 
We also run an hyper-parameter search for the reported hyper-parameters. Training of our full model takes around $20$ hours on a GeForce RTX 2080 Ti graphics card. 

We follow the notation of \cite{transformer} to describe the hyper-parameters of Transformer-based building blocks.

\subsection{\modelemb-$\encoder$}
\begin{figure}[t]
\centering
\strut\hfill
\includegraphics[width=\columnwidth]{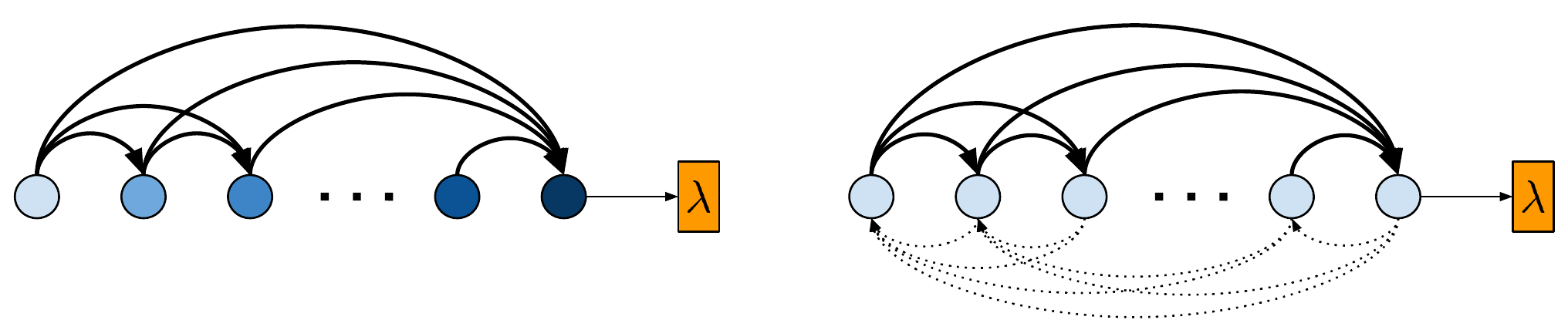}
\caption{\textbf{Transformer-based Stroke Encoders} Figures illustrate sequence- and set-based concepts for encoding the variable-length strokes $\stroke$ into $\strokecode$.
Each circle corresponds to a point in the stroke. 
Shading represents positional encoding to inject sequential information. 
Solid and dashed arrows denote access to information from past to the future and from the future to the past, respectively. 
(left) Our \modelemb-$\encoder$ with using a \emph{sequence} interpretation of strokes and (right) its counterpart using \emph{sets}. Note that we stack a number of this layers in our model and the output of the top-most layer for the last input step is used as the stroke embedding $\strokecode$.
}
\label{fig:transformer_encoder}
\end{figure}

Our encoder \modelemb-$\encoder$ follows the Transformer encoder architecture \cite{transformer}. 
We experiment with the following variants (see \Fig{transformer_encoder}).
\begin{enumerate}
    \item With positional encoding and look-ahead masks,
    \item With positional encoding and without look-ahead masks in a bi-directional fashion,
    \item Without positional encoding and without look-ahead masks, which corresponds to modelling the points in a stroke as set of points.
\end{enumerate}

We empirically find that treating the stroke as a sequence rather than as a set yields significantly better performance. 
Similarly, restricting the attention operation from accessing the future points in the stroke is beneficial.
This implies that the temporal information is informative for capturing local patterns in the stroke level. 
We further show that at the drawing level, the temporal information is not important.

For a given stroke sequence, we first feed it into a dense layer to get a $d_{model}$-dimensional representation for the transformer layers, which is followed by adding positional embeddings. 
We stack $6$ layers with an internal representation size of $d_{model} = 64$ and feed-forward representation size of $d_{ff}=128$. 
We use multi-head attention with $4$ heads and omit dropout in the transformer layers. 

To compute the stroke embedding $\strokecode$ we use the output of the top-most transformer layer at the last time-step. The stroke embedding $\strokecode$ is obtained by feeding this 64-dimensional vector into a single dense layer with $D$ output nodes without activation function.

\subsection{\modelemb-$\decoder$}
Our decoder consists of $4$ dense layers of size $512$ with ReLU activation function. 
The output layer parameterizes a GMM with $20$ components from which we then sample the $(x,y)$ positions for a given stroke embedding $\strokecode$ and $t$. 

During training, we pair a stroke embedding $\strokecode$ with $4$ random $t$ values and minimize the negative log-likelihood of the corresponding $s(t)$ targets. We map a stroke to the range $[0, 1]$ and then interpolate $s(t)$ for $t$ values that are not corresponding to a point.

\subsection{\modelemb-$\relational$}
We consider a drawing as a set of strokes and aim to model the relationship between strokes on a $2$-dimensional canvas. 
We use the self-attention concept which explicitly relates inputs with each other. 
Our model follows the Transformer decoder architecture \cite{transformer}. 
In order to prevent the model from having any sequential biases we do not apply positional encoding and look-ahead masks (similar to \Fig{transformer_encoder} right), enabling us to model strokes as a set.

We use separate models to make position and embedding predictions for the next stroke. 
We concatenate the given stroke embeddings and their corresponding start positions and feed into the position prediction model. 
The embedding prediction model additionally takes the start position of the next stroke. 
It is appended to the every given stroke. At inference time, we use the predicted start positions while they are the ground-truth positions during training. 
In summary, input size of the position prediction model is $10$ consisting of $8$D stroke embedding $\strokecode$ and $2$D position. For the embedding prediction model, it is $12$ with an additional $2$D start position of the next stroke. 
Similar to the \modelemb-$\encoder$, the model representation of the top layer for the last input stroke is used to make a prediction. Both models predict a GMM with $10$ components to model position and stroke embedding predictions. 

We use the same configuration for both models. We stack $6$ layers with an internal representation size of $d_{model} = 64$ and feed-forward representation size of $d_{ff}=256$. We use multi-head attention with $4$ heads and omit dropout in the transformer layers. 

For a given drawing sample, we create $32$ subsets of strokes randomly. $16$ of them preserves the drawing order while the remaining $16$ are shuffled (see \Sec{training}).

Our model's sequential counterpart (i.e., \modelemb-$\relational$~(Ord.) in \Tab{res_table}) uses positional encodings to access the temporal information. It is also trained by preserving the order of strokes in the input subsets.

\subsection{Ablation models}
We evaluate our hypothesis by replacing our model's components with fundamental architectures for the underlying task. In all experiments, we follow the same training and evaluation protocols as with our model. 

\subsubsection{Seq2Seq Stroke Embedding Model}
In our ablation study, we experiment with a \modelseq model to encode and decode variable-length strokes. We use LSTM cells of size $512$ in the encoder and decoder. We find that processing the input stroke in both directions gives a better performance. For reconstruction, the decoder takes the stroke embedding at every step. In the \modelseq-AR counterpart, we feed the decoder with the prediction of the previous step as well. 

Similar to our \modelemb-$\encoder$/$\decoder$, the outputs are modeled with a GMM and the model is trained with negative log-likelihood objective.

\subsubsection{VAE Extension}
We apply KL-divergence regularization to learn a smoother and potentially disentangled latent space. 
The encoder simply parameterizes a Normal distribution corresponding to the approximate posterior distribution. In addition to the reconstruction loss, we apply an KL-divergence loss between the approximate posterior and a standard Gaussian prior $\mathcal{N}(0, I)$. We follow the same annealing strategy as in \cite{ha2017neural}.

\subsubsection{RNN Prediction Model}
We use LSTM cells of size $512$ for the position and embedding prediction models. We observe that training the RNN prediction models with both ordered and random subsets of stroke embeddings does not make a significant difference. In other works, the RNN models do not benefit from using set of strokes.

\subsection{\sketchrnn Baseline}
We train the decoder-only \sketchrnn model by following the instructions in \cite{ha2017neural} on the publicly-available codebase. We use an LSTM of size $1000$ and use the default hyper-parameters. Similar to the quantitative evaluation of our models, we predict $10$ samples from the \sketchrnn model by conditioning the model on the context strokes and ground-truth start position. We run the quantitative evaluation with various sampling temperatures in $[0.1, 0.3, 0.5, 0.7, 0.9]$ and report the best result.

\subsection{Training}
For the training of models with RNN components we find that an initial learning rate of $1e^{-3}$, annealed with an exponential decay strategy works the best. We use the Transformer learning rate scheduling proposed in \cite{transformer} for the Transformer based models.

All models are trained with a batch size of $128$. The stroke embedding models are trained by treating strokes independently. For the relational models, we create a new batch on-the-fly consisting of the random subsets of stroke embeddings. 

\begin{wrapfigure}{R}{0.45\textwidth}
\vspace{-1.5cm}
\centering
\includegraphics[width=0.45\columnwidth]{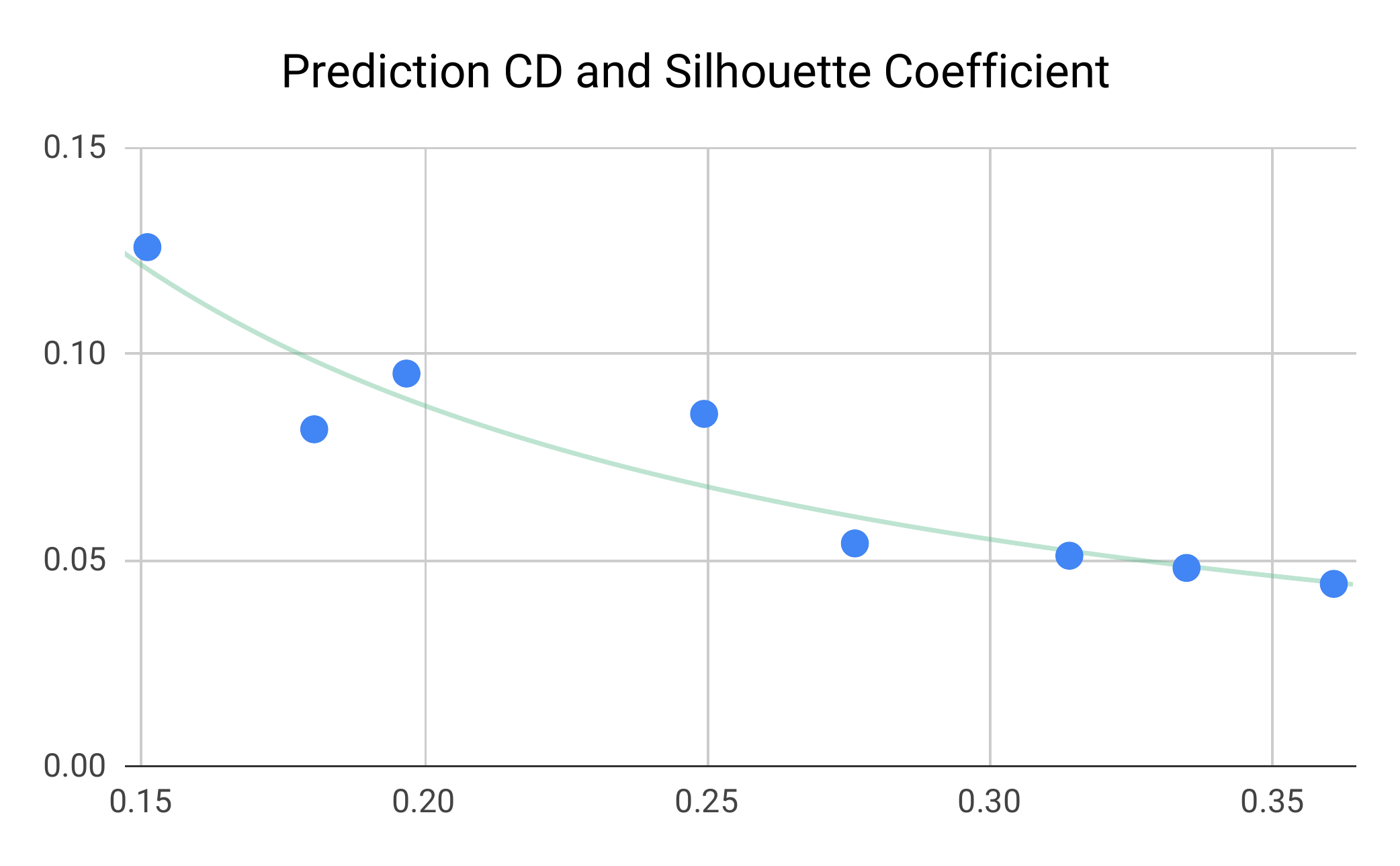}
\caption{Prediction Chamfer Distance (CD) (y-axis) and Silhouette Coefficient (SC) (x-axis) for the results presented in \Table{embedding_table}.}
\label{fig:sc_plot}
\vspace{-1cm}
\end{wrapfigure}

\section{Silhouette Coefficient}
\label{sec:sc}
\Fig{sc_plot} plots the silhouette coefficient (SC) and prediction chamfer distance (CD) results reported in \Table{embedding_table}. As we note in the main paper, the Pearson correlation between the SC and the prediction metrics is 0.92 indicating a strong
correlation between the two. We would like to note that we make this comparison among the models with our relation model \modelemb-$\relational$. Only the underlying embedding models vary.

\section{Embedding Predictions}
\label{sec:embedding_pred}

\begin{figure}[h]
\vspace{-0.2cm}
\centering\small
\includegraphics[width=.32\columnwidth, trim={10pt 10pt 10pt 10pt}]{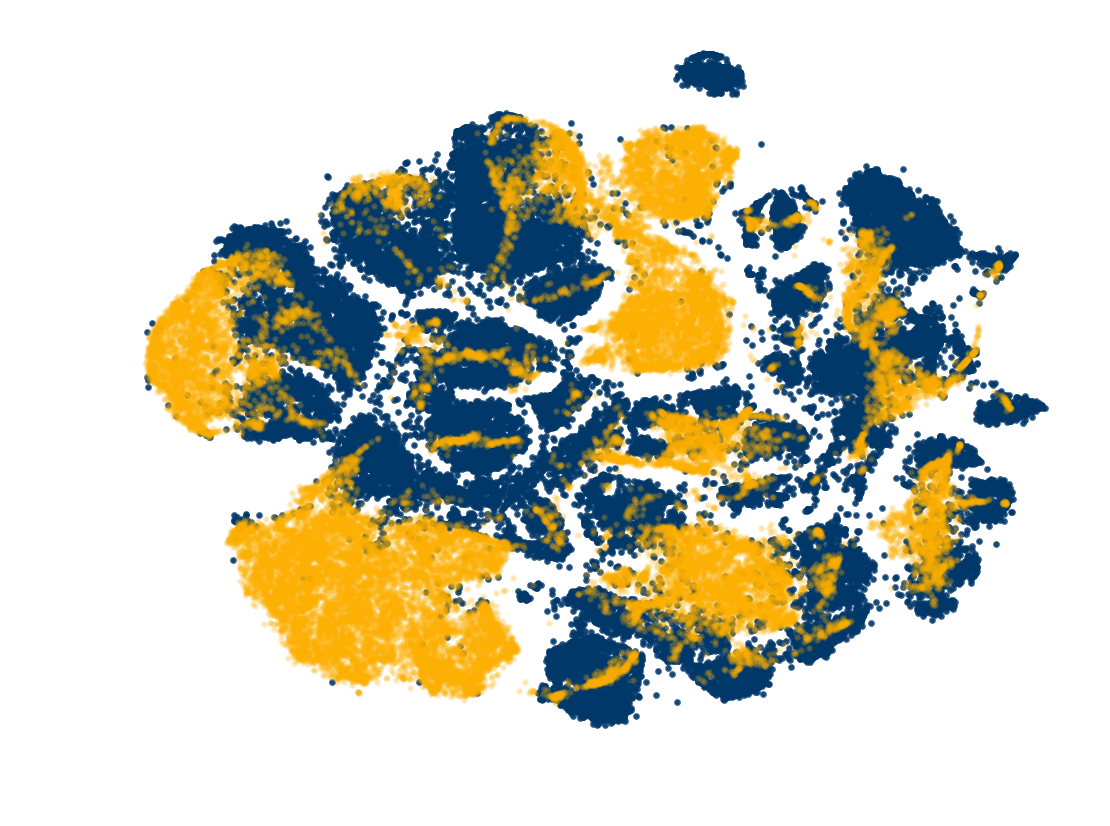}
\includegraphics[width=.32\columnwidth, trim={10pt 10pt 10pt 10pt}]{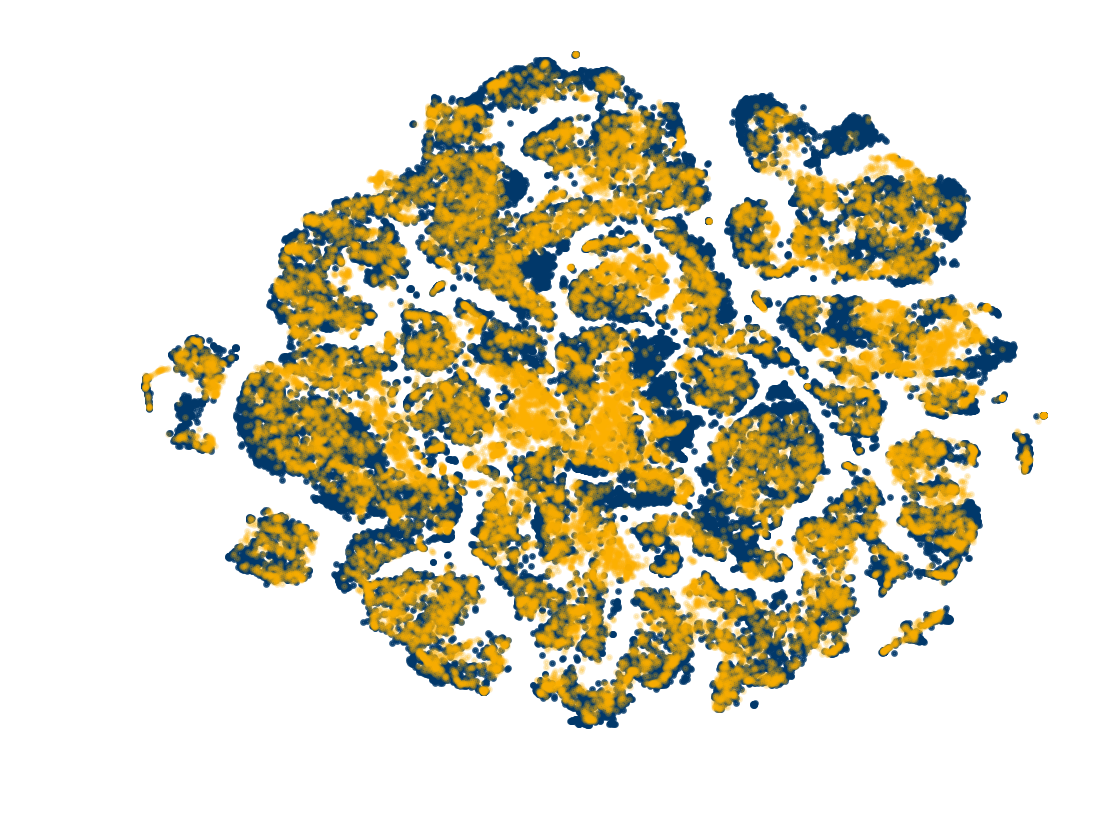}
\includegraphics[width=.32\columnwidth, trim={10pt 10pt 10pt 10pt}]{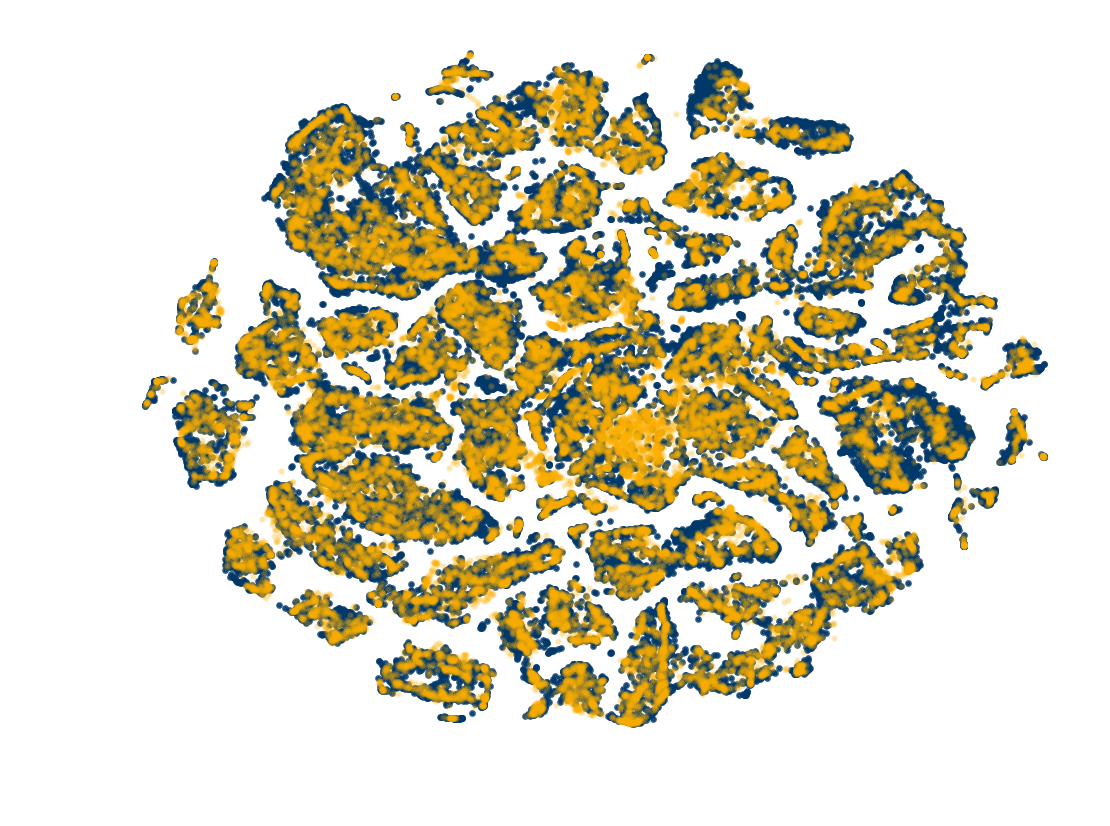}
\caption{\textbf{tSNE Embedding} -- 
\textcolor{gtemb}{Blue points correspond to embeddings computed from the original data}, 
\textcolor{predemb}{yellow points correspond to embeddings predicted by our relational model $\relational$}.
(Left) Our model with KL-divergence regularization on the latent space (i.e., \modelemb-$\encoder$/$\decoder$+VAE in \Table{embedding_table}), 
(middle) our model trained in a sequence-based fashion (i.e., \modelemb-$\relational$~(Ord.) in \Tab{res_table}), (right) our model (i.e., \modelemb-$\relational$ in \Tab{res_table}).
}
\label{fig:tsne_collapse}
\end{figure}

In \Fig{tsne_collapse} we provide an additional analysis of the embedding spaces obtained from different models. 

Each subplot shows a tSNE visualization of the embedding spaces, where blue points correspond to the embedding of strokes from the original data and yellow points correspond to embeddings predicted from our relational model $\relational$. 
This clearly demonstrates that our model is able to predict much more ``\emph{natural}'' embeddings in the embedding space from \modelemb-$\encoder/\decoder$ than from the other two models.
Our proposed model achieves a higher amount of overlapping with the real data. 
When we ignore our set assumption and model the strokes as a sequence, we start observing non-overlapping regions in the visualizations. 
Finally, our model fails to operate well in the latent space governed by the KL-divergence regularization. The underlying stroke embedding model is underfitting due to the strong KL-divergence regularizer, which further degrades the prediction performance.
We also quantify this effect by calculating the Earth-Mover distance (EMD) between the two embedding distributions. Our model $\modelemb$-$\relational$ achieves an EMD of 155 while the sequential and the VAE counterparts result in EMDs of 1797 and 251, respectively. The EMD decreases as the GT and predicted distributions become more similar.
We would like to note that our analysis on the embeddings also agrees with the prediction metric results reported in the main paper.

\section{MSE Reconstruction Objective}
\label{sec:appendix_MSE}
We compare our probabilistic reconstruction objective (i.e., log-likelihood with a GMM) with a deterministic mean-squared error (MSE). 
The model configuration is the same in both experiments with a difference in the reconstruction objective and the number of $t$ samples per stroke. 
We use $100$ $t$ samples for the model trained with MSE whereas it is only $4$ for the model with GMM predictions.

\begin{figure}[t]
\centering
\newcommand{\diagram}[2]{\hfill\includegraphics[width=#1\linewidth,keepaspectratio=true,valign=c]{fig/supp_gmm_vs_mse/#2}\hfill}

\diagram{0.2}{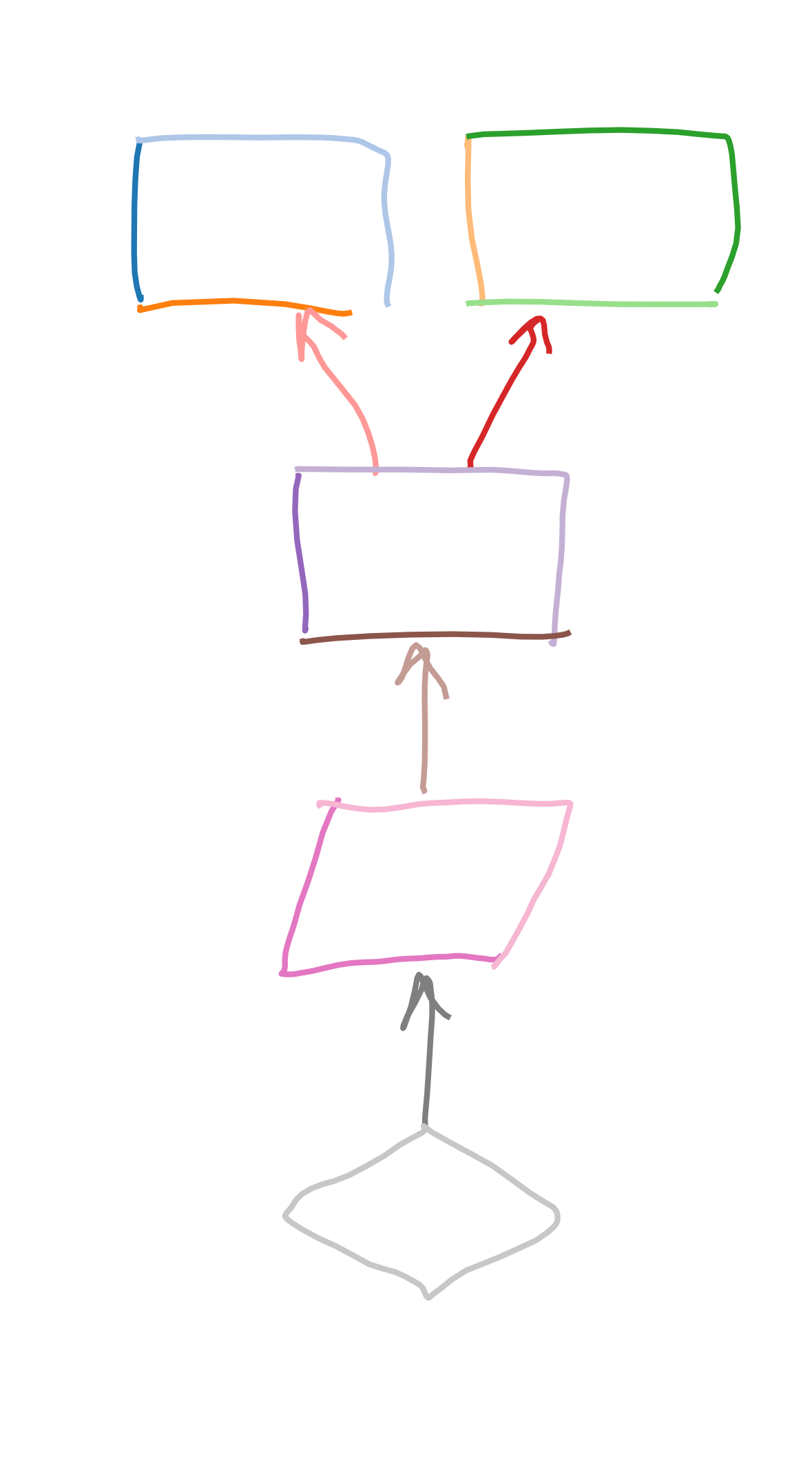}
\diagram{0.2}{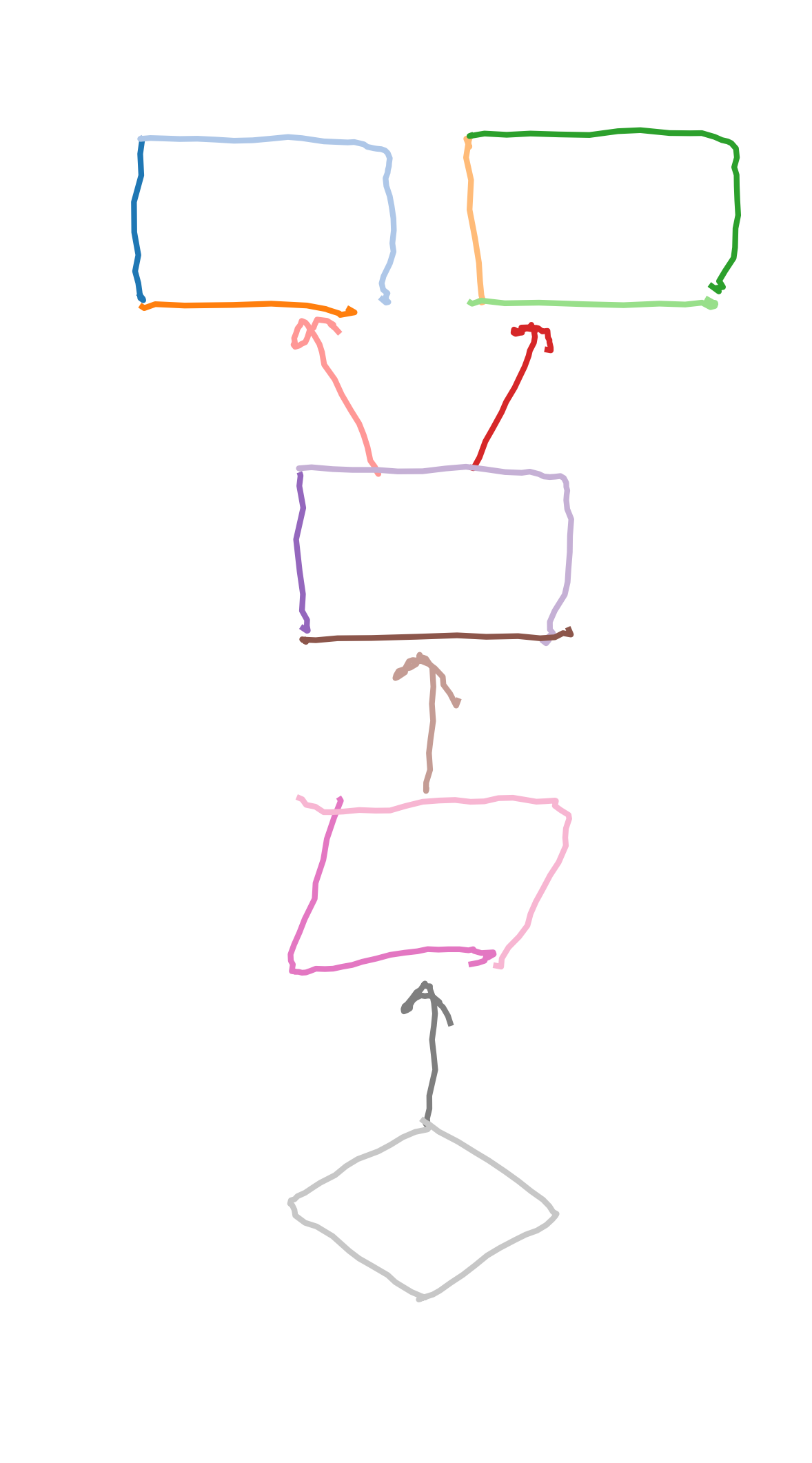}
\diagram{0.2}{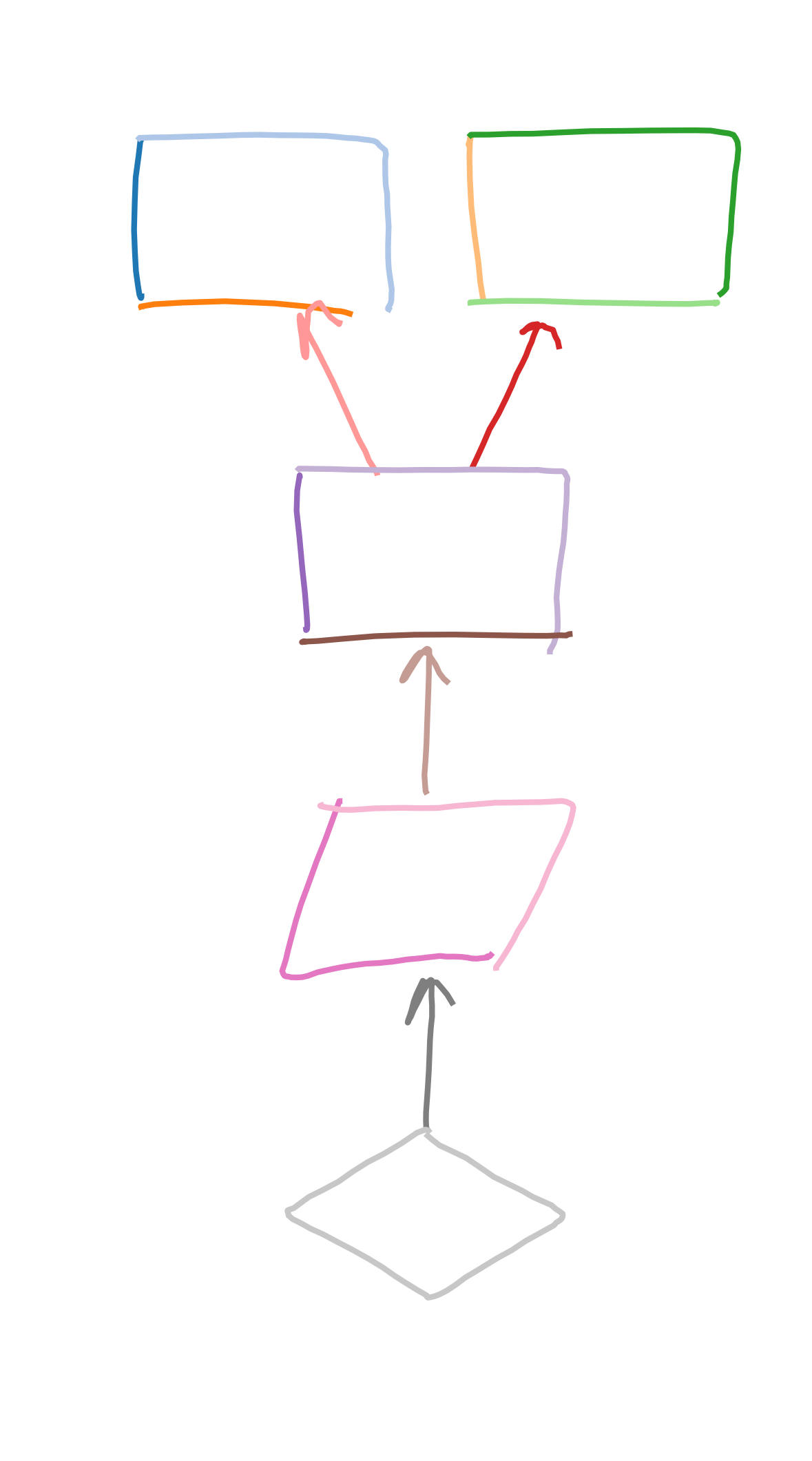} \strut\\
\diagram{0.15}{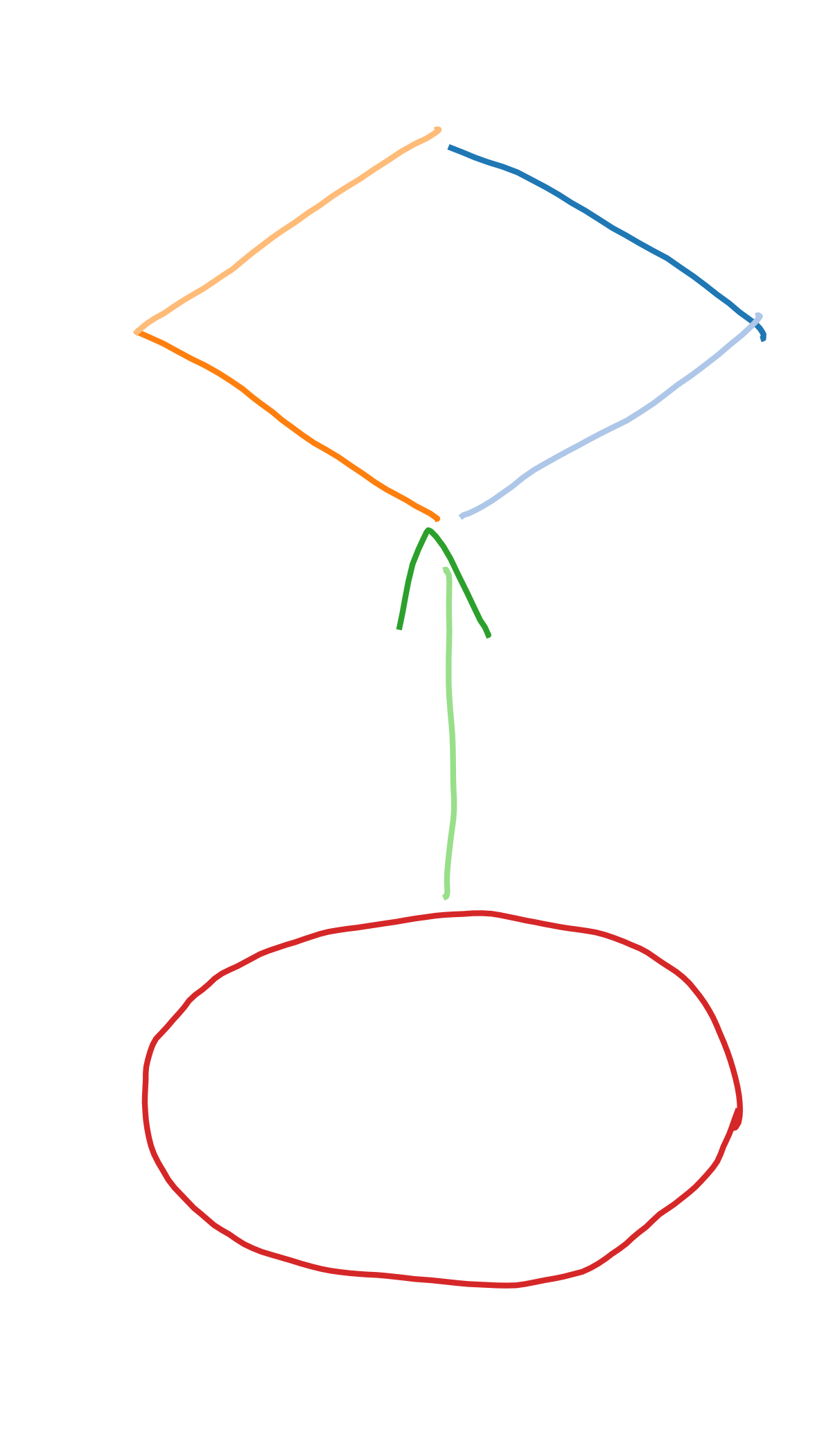}
\diagram{0.15}{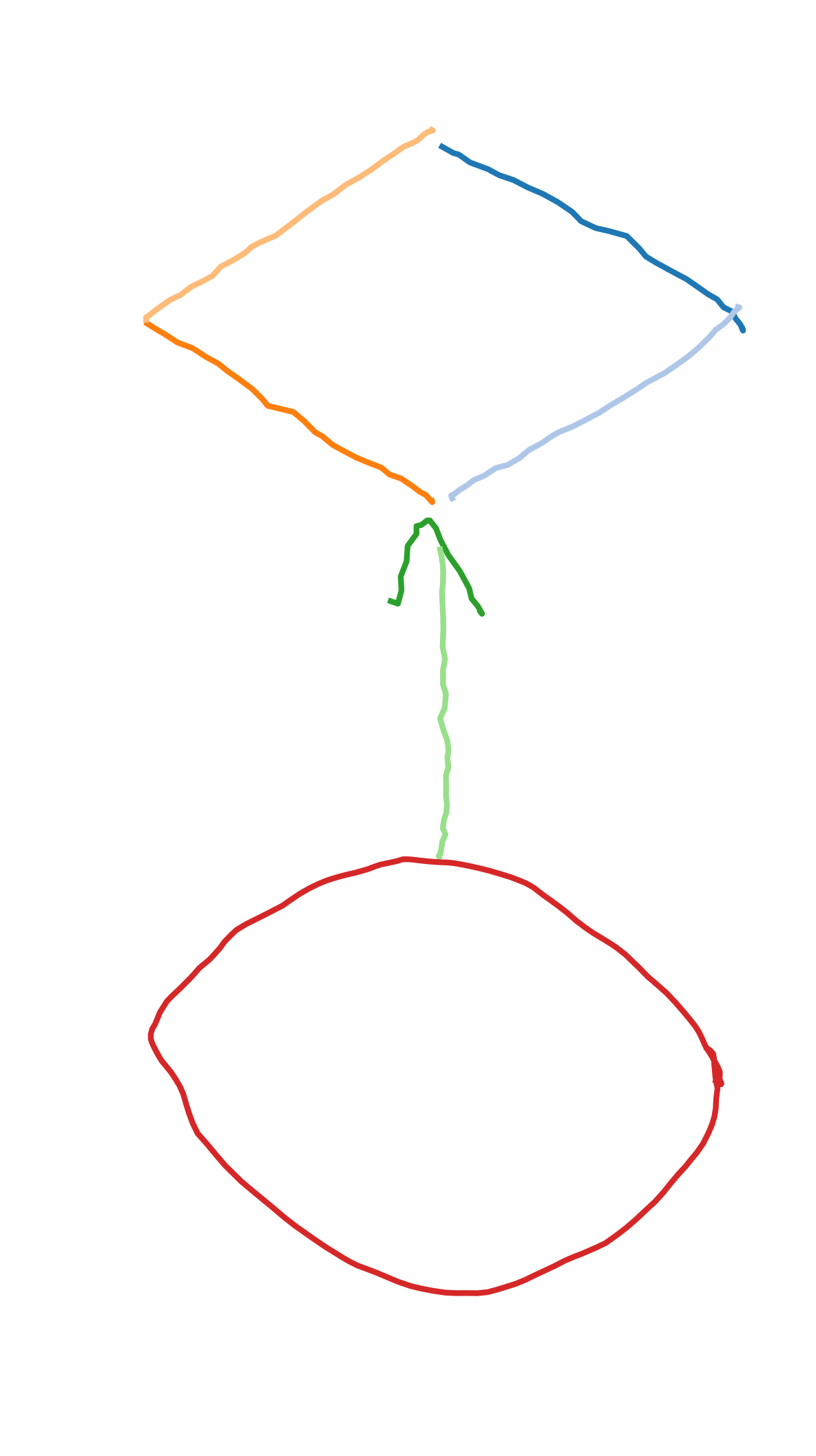}
\diagram{0.15}{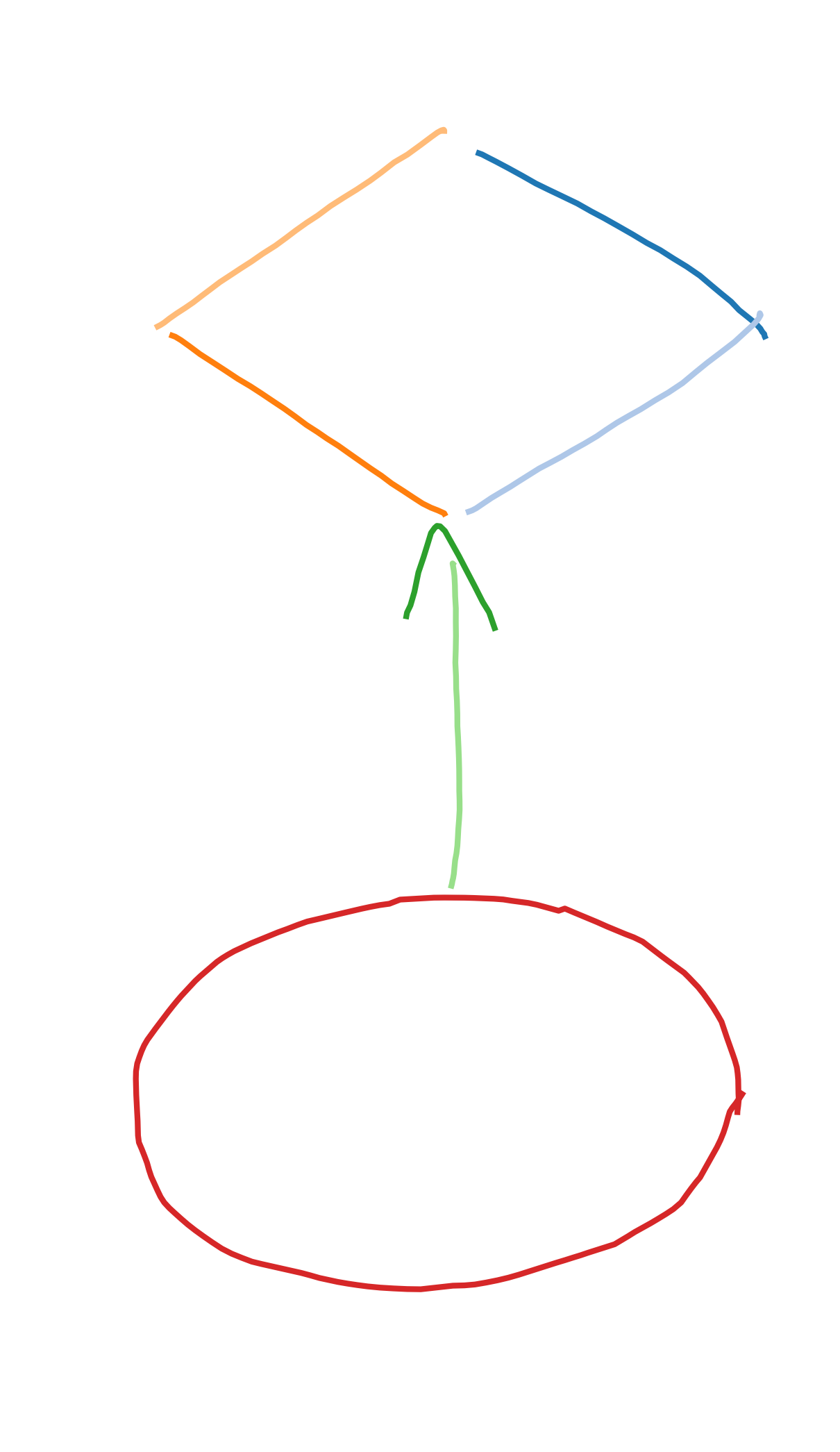} \strut\\
\diagram{0.25}{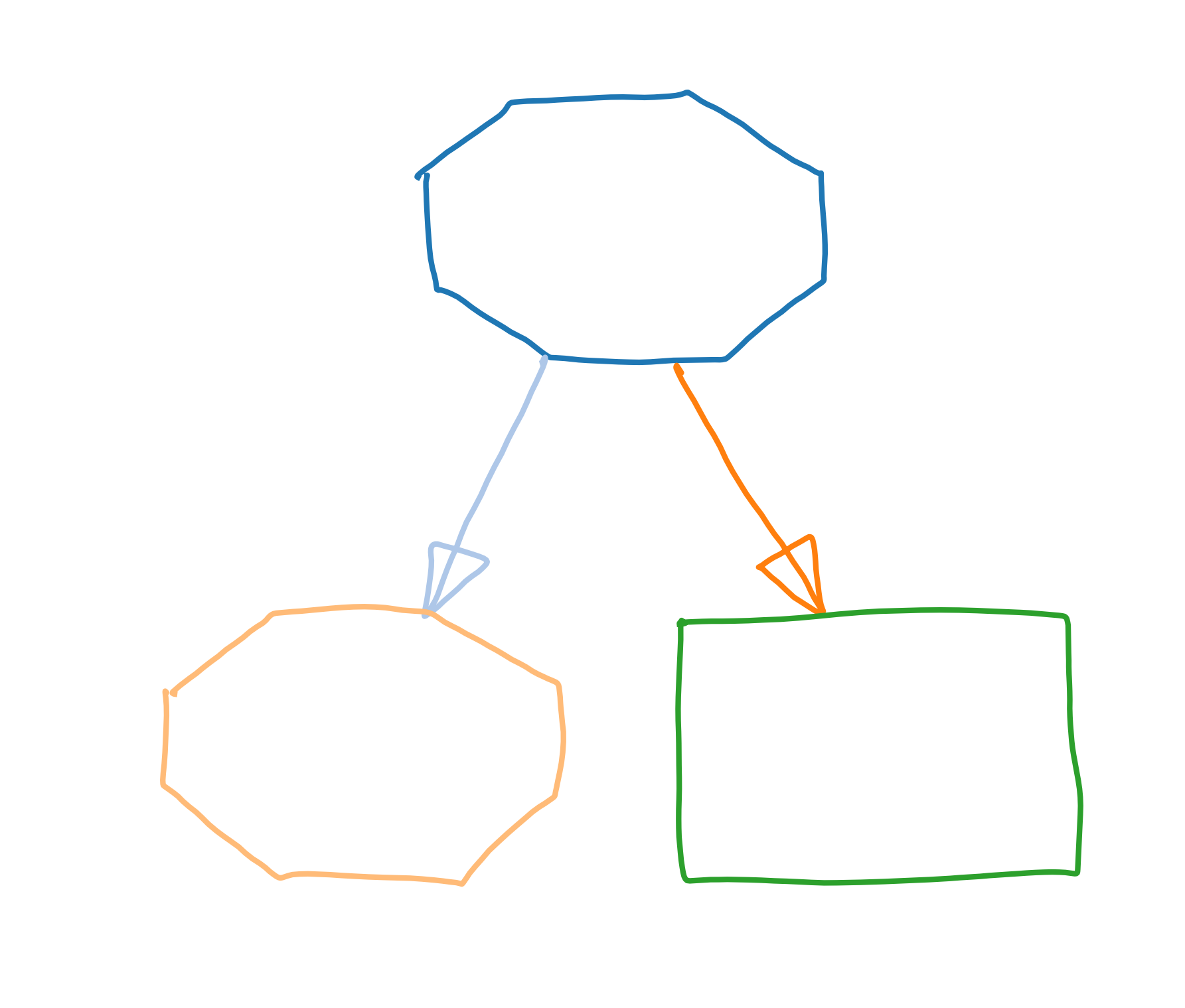}
\diagram{0.25}{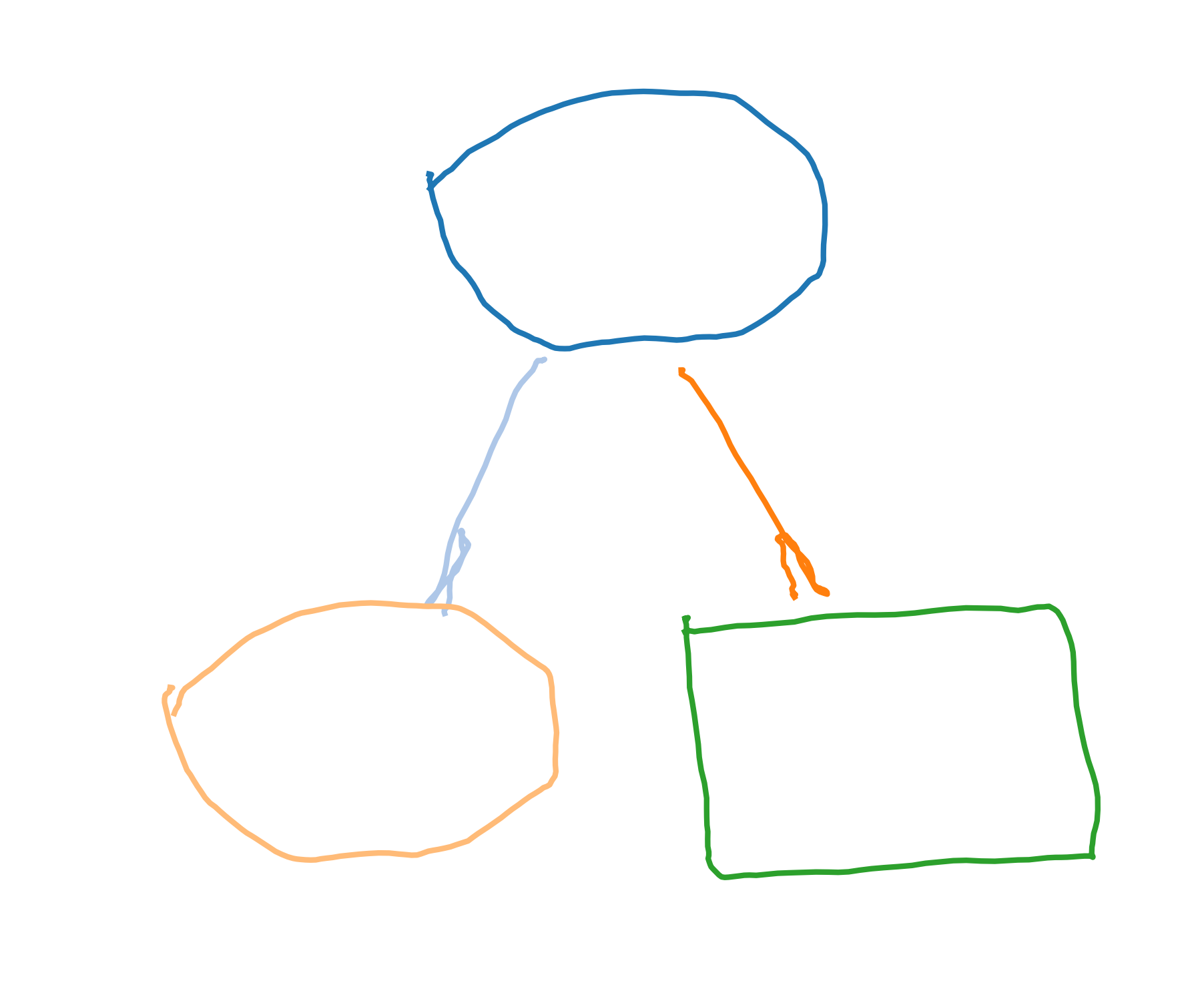}
\diagram{0.25}{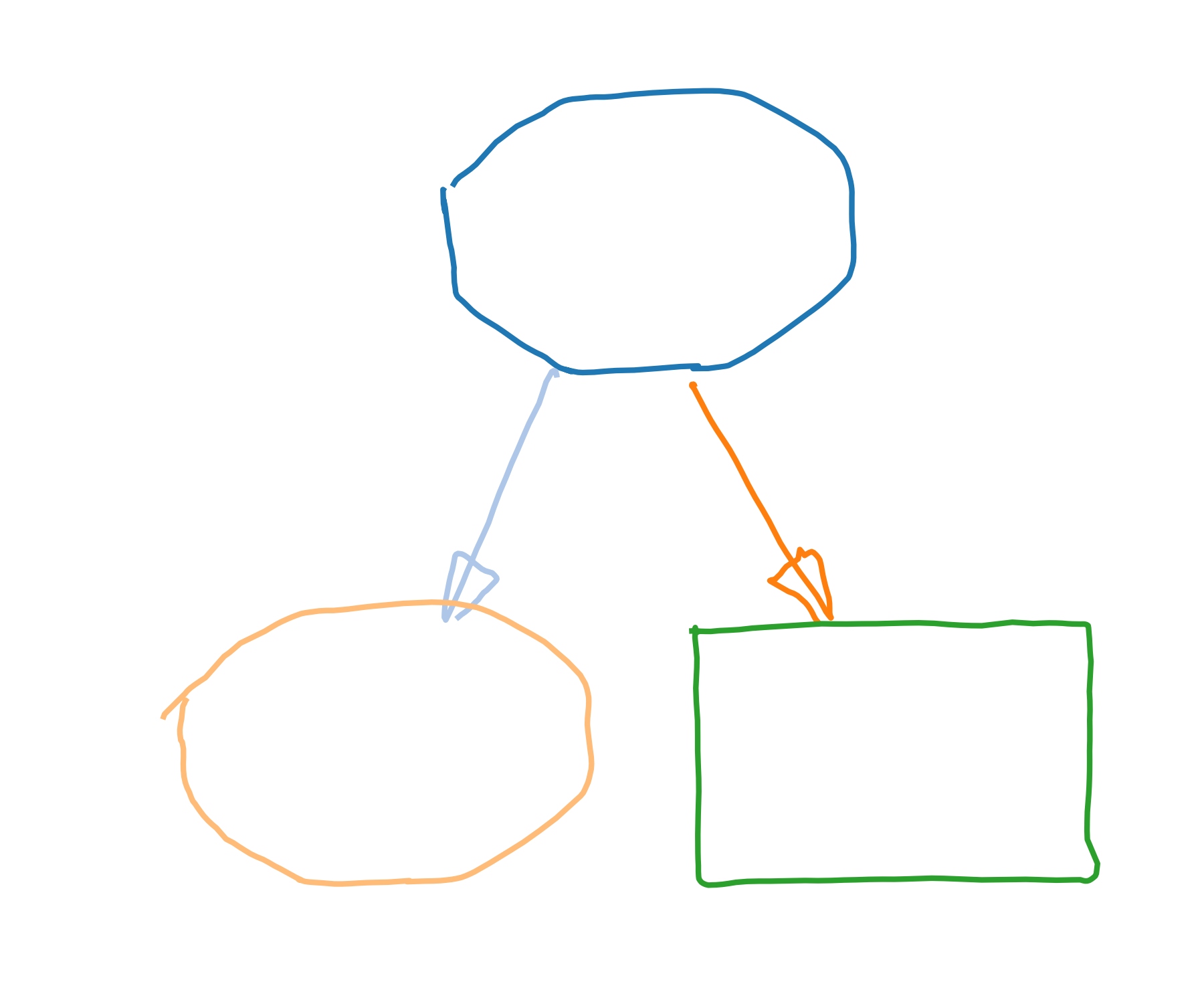} \strut\\

\caption{{\bf Reconstruction performance of deterministic and probabilistic objectives} (left) Ground-truth sample, reconstruction performance of the model trained with MSE (middle) and trained with log-likelihood objective and GMM output model (right).}
\label{fig:gmm_vs_mse}
\end{figure}

The MSE objective results in a competitive stroke reconstruction loss ($0.018$ compared to $0.014$ of the model with GMM log-likelihood). 
\Fig{gmm_vs_mse} shows qualitatively that the reconstructions are noisy although the overall shape information is preserved. 

\begin{figure}[h]
\centering\small
\includegraphics[width=.25\columnwidth, trim={0pt 0pt 0pt 0pt}]{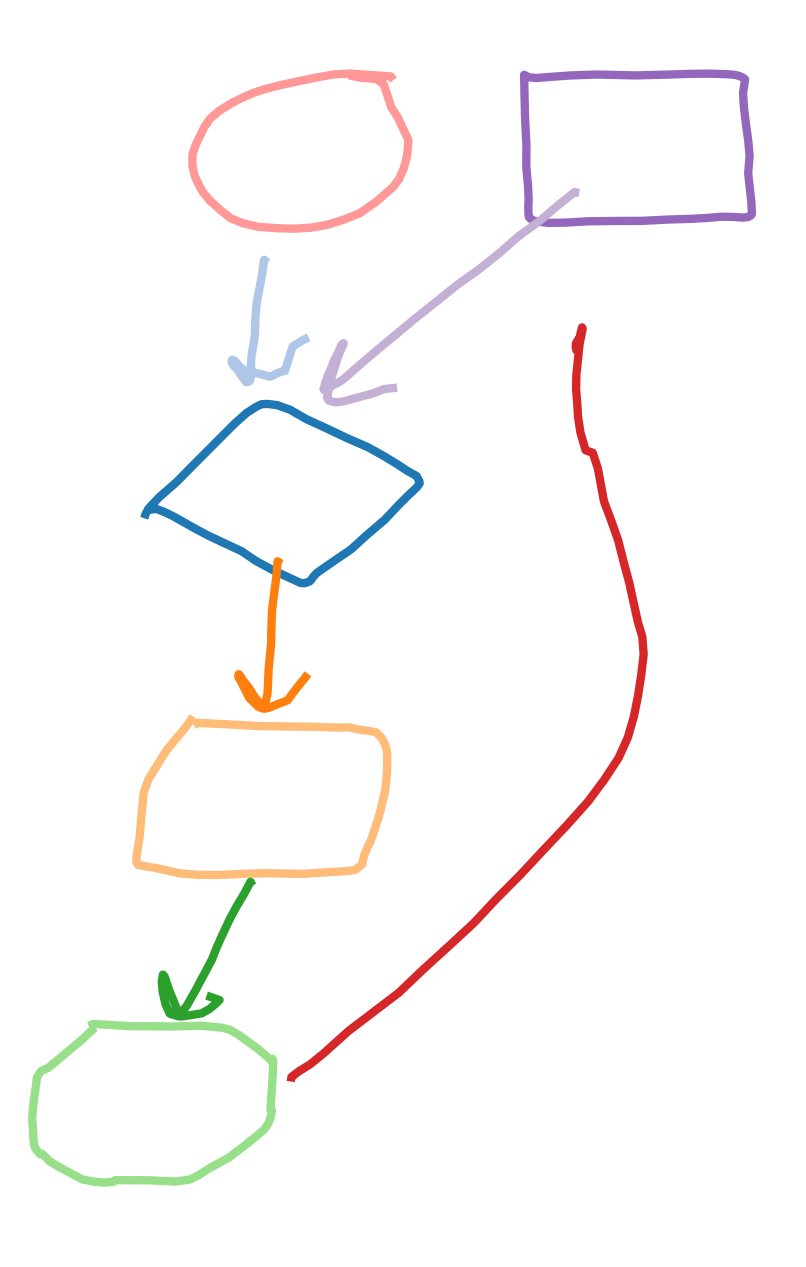}
\includegraphics[width=.25\columnwidth, trim={0pt 0pt 0pt 0pt}]{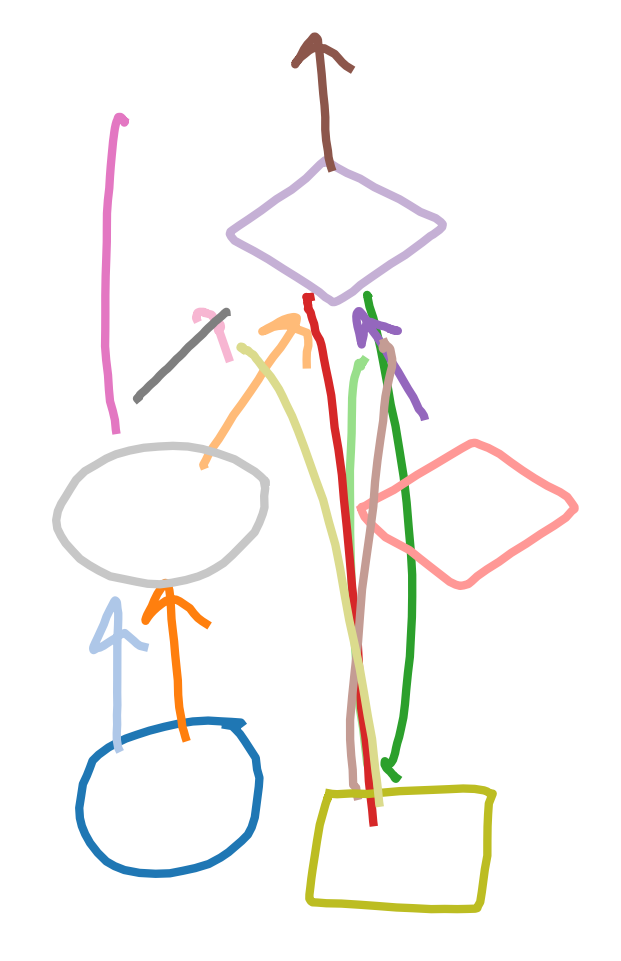}
\includegraphics[width=.45\columnwidth, trim={0pt 0pt 0pt 0pt}]{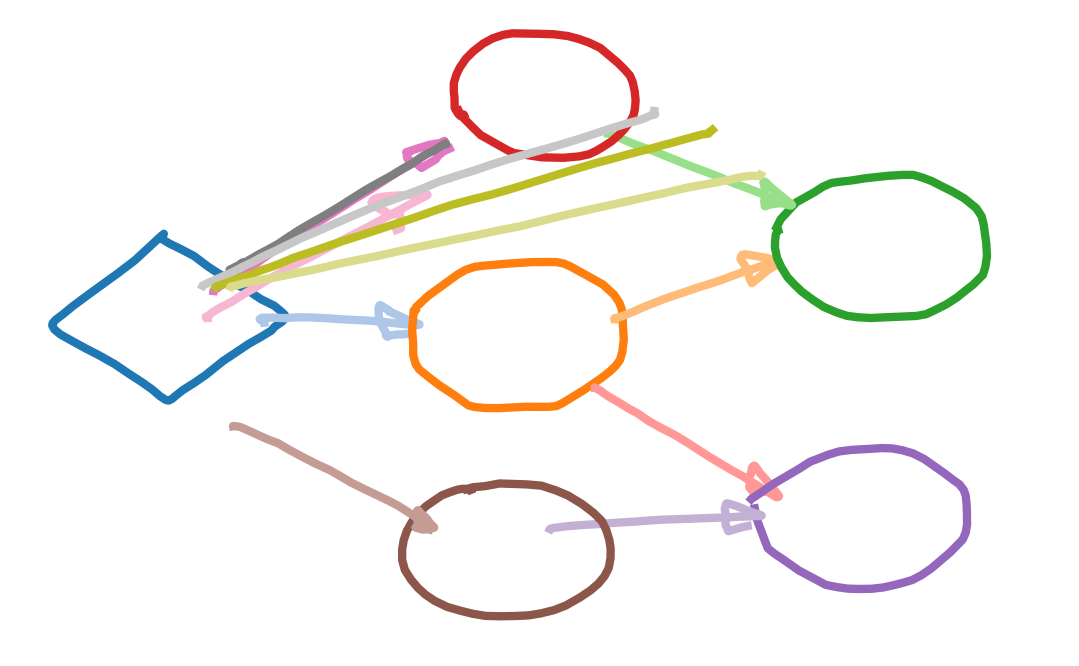}
\vspace{.1in}
\caption{\textbf{Failure cases} -- Given the \textcolor{firststroke}{\textbf{first}} and \textcolor{secondstroke}{\textbf{second}} strokes, failed predictions of our model. (Left) A problem with connecting distant shapes via a long arrow. (Middle-Right) With increasing number of predictions, our model may predict overlapping arrows.
}
\label{fig:failure}
\end{figure}

\section{Limitations}
\label{sec:limitation}
In our work, we consider strokes as an atomic unit of drawings. However, there exist cases where users draw a very long and complex stroke such as in cursive handwriting. 
Our stroke embedding model fails to capture details of such long and complex strokes. 
We argue that it can be resolved by following a simple heuristic where too long strokes are split into shorter ones. 
Then the position prediction model is expected to predict the start position consecutive to the ending of the previous split.

In \Fig{failure} we present common failure cases of our model. While it is straightforward to link the neighboring shapes, our model sometimes struggles to connect distant shapes via longer arrows.

We also observe that as the number of predicted strokes increases (i.e. $\sim15$), it becomes more likely to predict arrow-like shapes by ignoring the existing content. 

There are two main issues that may cause these failure cases. First, the number of shorter arrows connecting closer shapes is significantly higher in the training data. Hence, our model possibly fails to capture distant connection patterns. Second, our embedding space is not explicitly regularized to be smooth. Despite the fact that it works very well for our purpose, this may also cause our relational model to make predictions from regions with low density. 

\section{More visualization}
\label{sec:more visualization}
\begin{figure}[p]
\newcommand{\suppzooh}[1]{\hfill\includegraphics[width=.3\linewidth,height=.24\linewidth,keepaspectratio,valign=c]{fig/supp_zoo/#1}\hfill}

\newcommand{\suppzoos}[1]{\hfill\includegraphics[width=.22\linewidth,height=.24\linewidth,keepaspectratio,valign=c]{fig/supp_zoo/#1}\hfill}
\newcommand{\suppzoov}[1]{\hfill\includegraphics[width=.22\linewidth,height=.25\linewidth,keepaspectratio,valign=c]{fig/supp_zoo/#1}\hfill}

\suppzooh{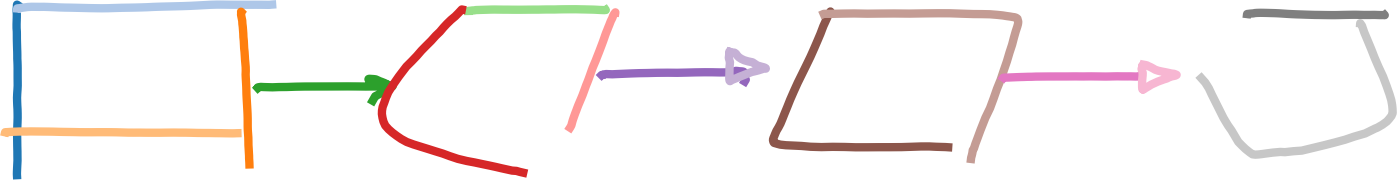}
\suppzooh{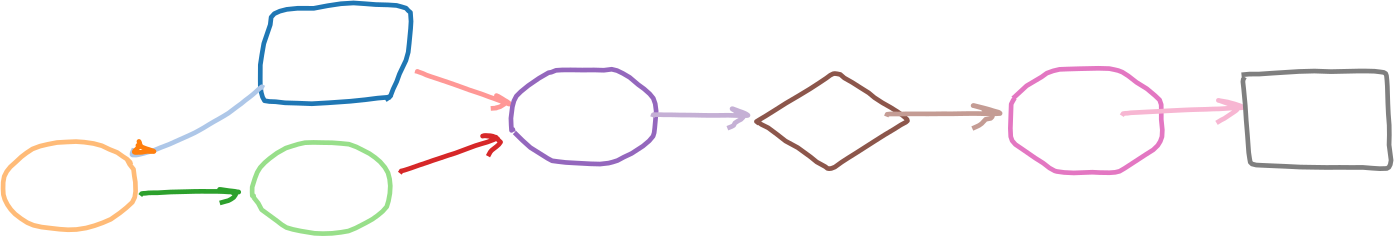}
\suppzooh{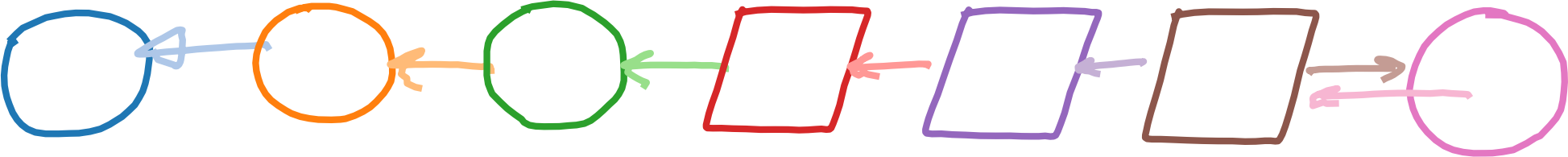}\\[2ex]
\suppzooh{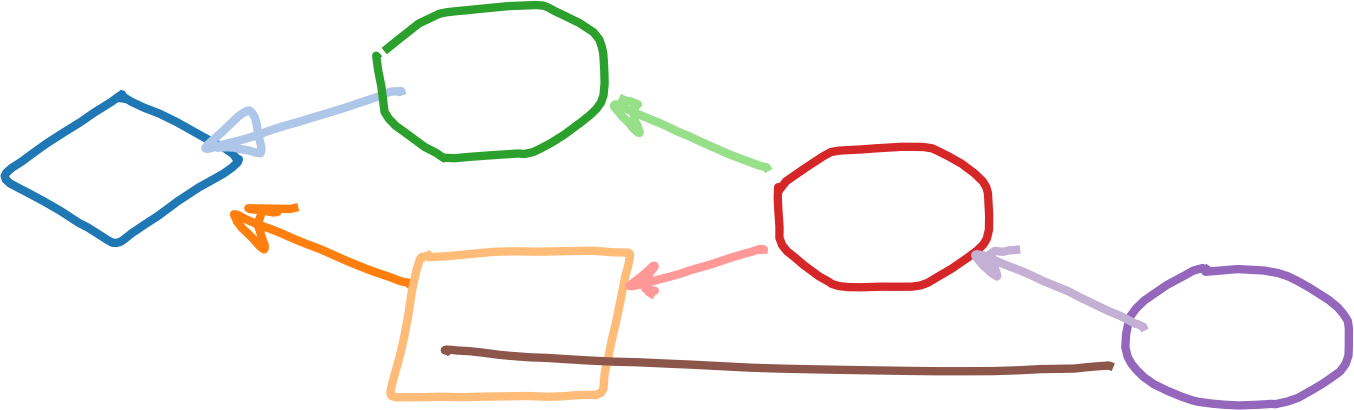}
\suppzooh{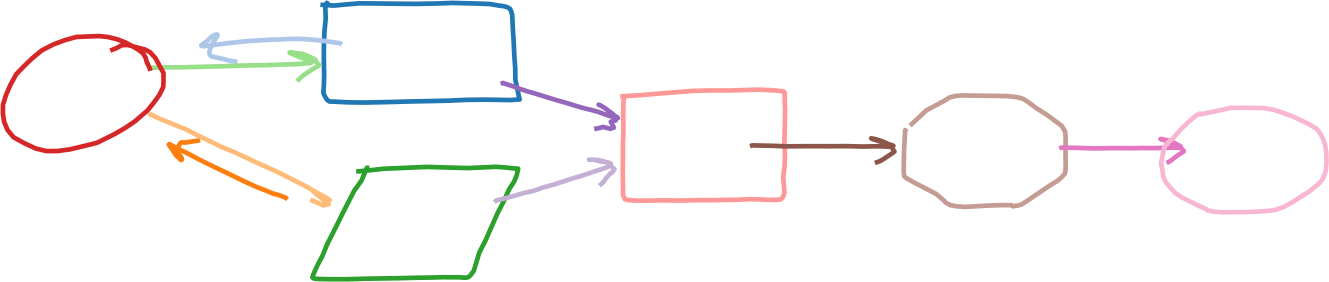}
\suppzooh{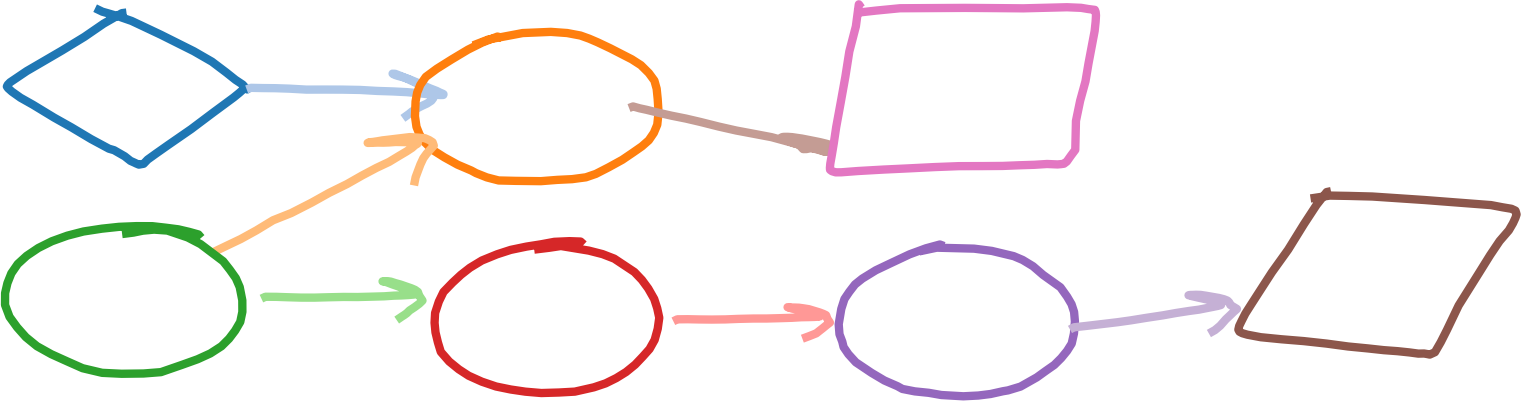}\\[2ex]
% \suppzooh{298_pos_ar_heatmap_ordered_s10_v2}
% \suppzooh{376_pos_ar_heatmap_ordered_s15}

\suppzoos{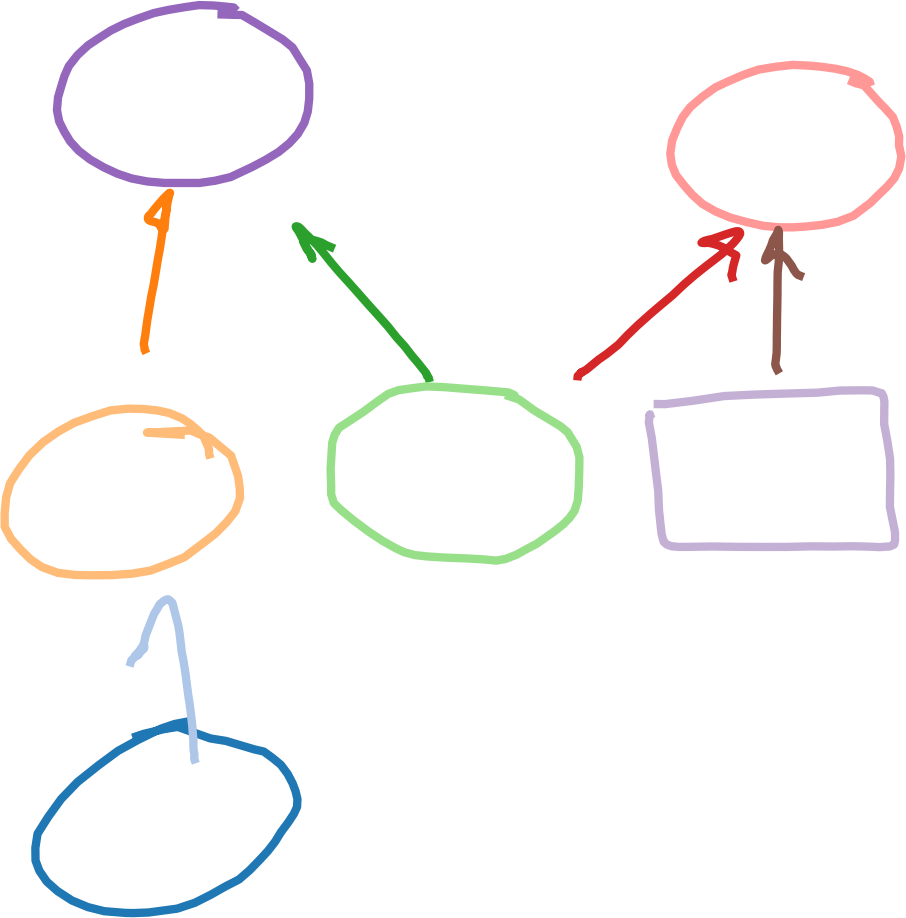}
\suppzoos{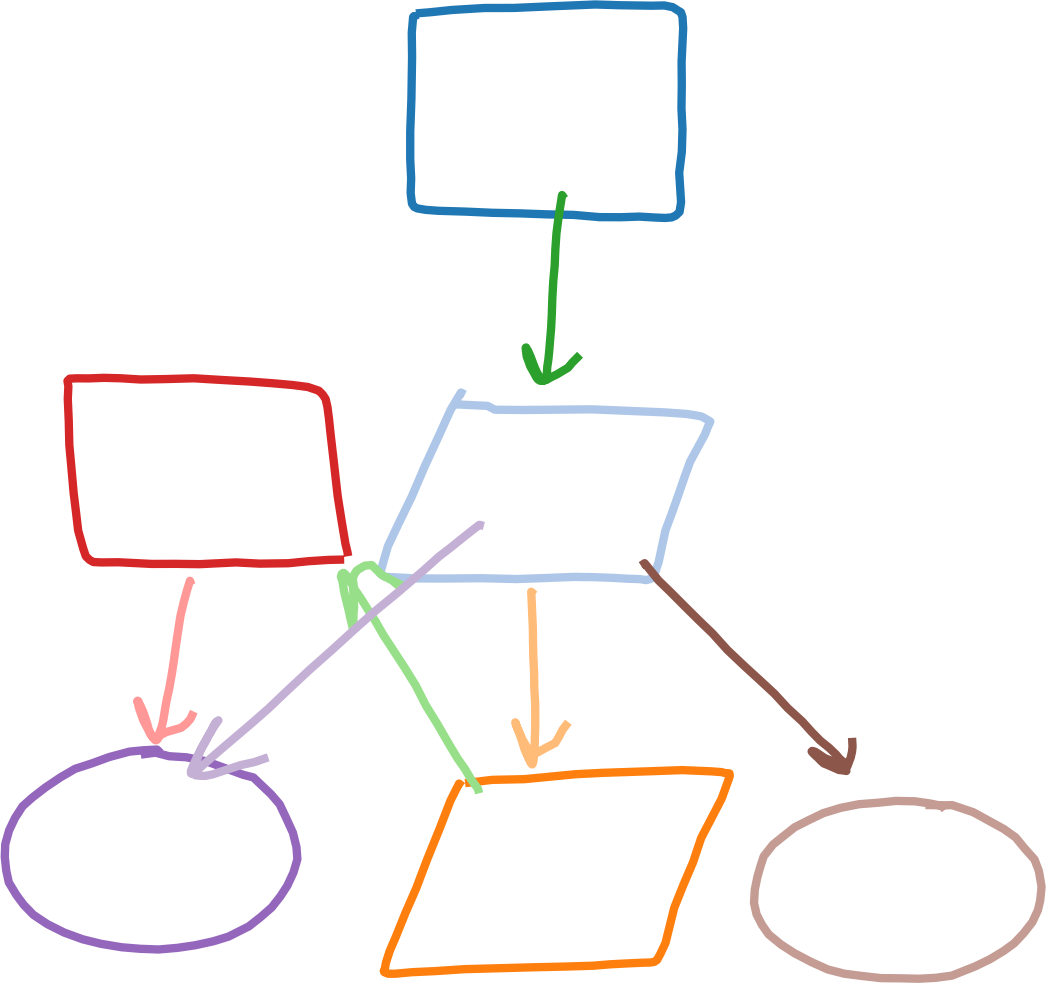}
\suppzoos{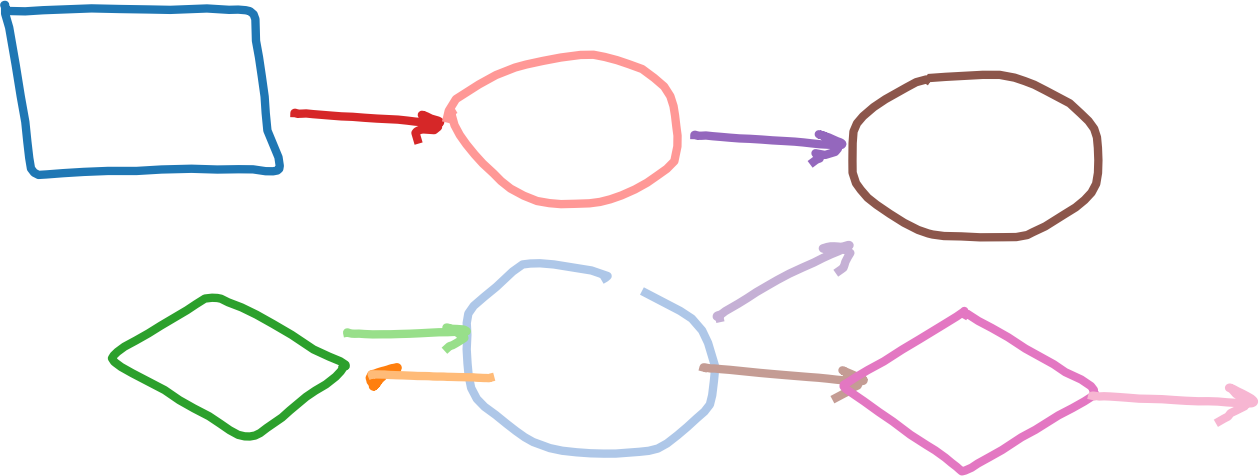}
\suppzoos{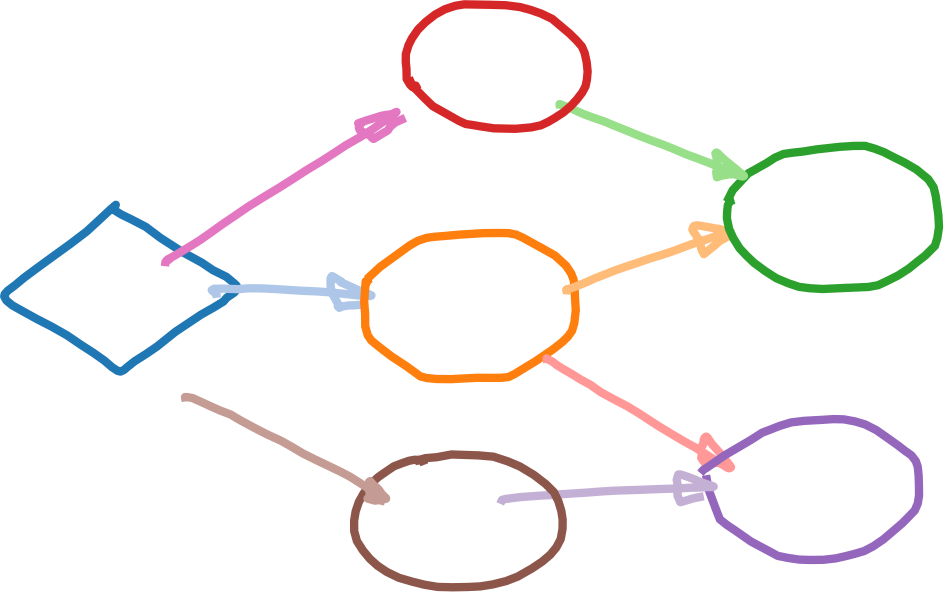}\\[2ex]
\suppzoos{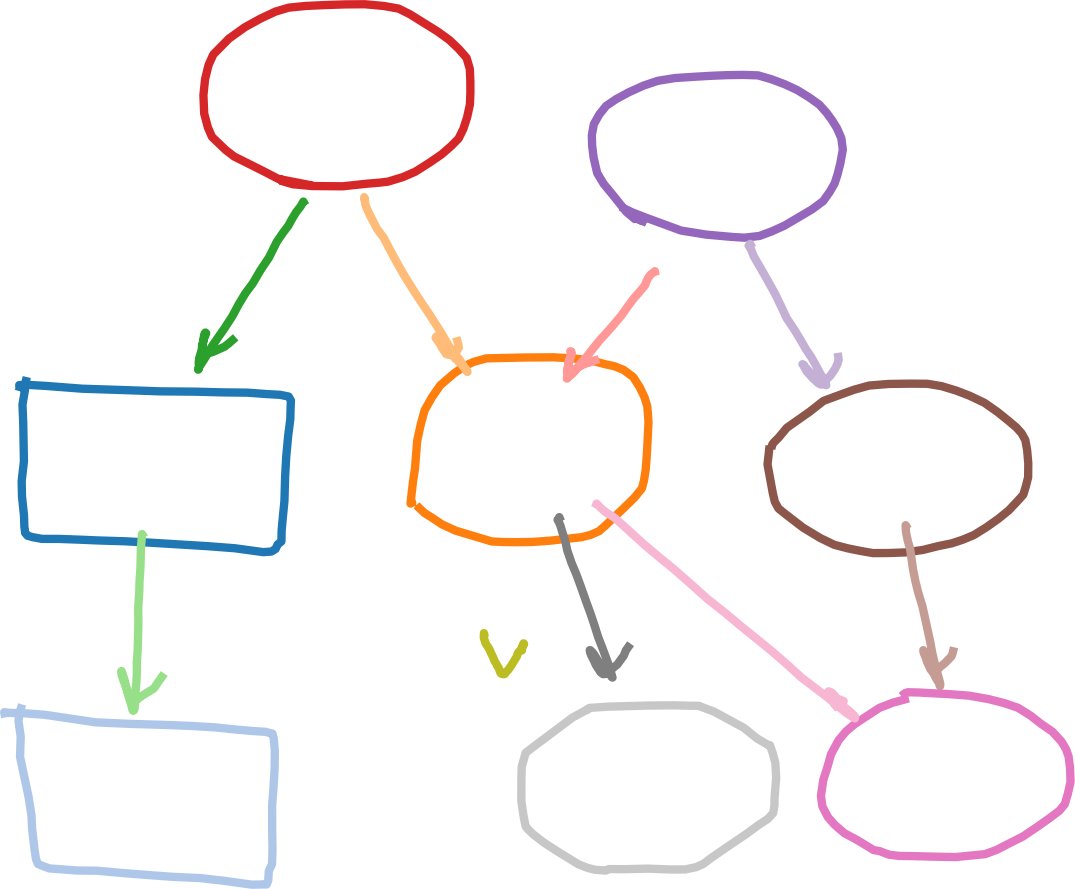}
\suppzoos{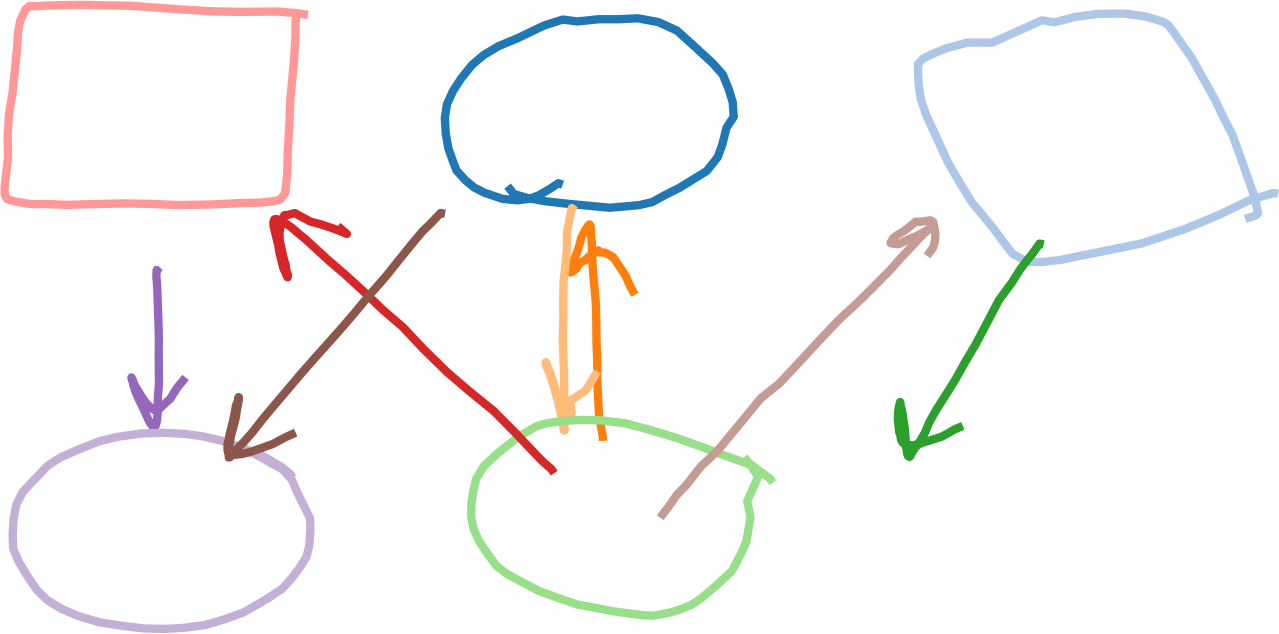}
\suppzoos{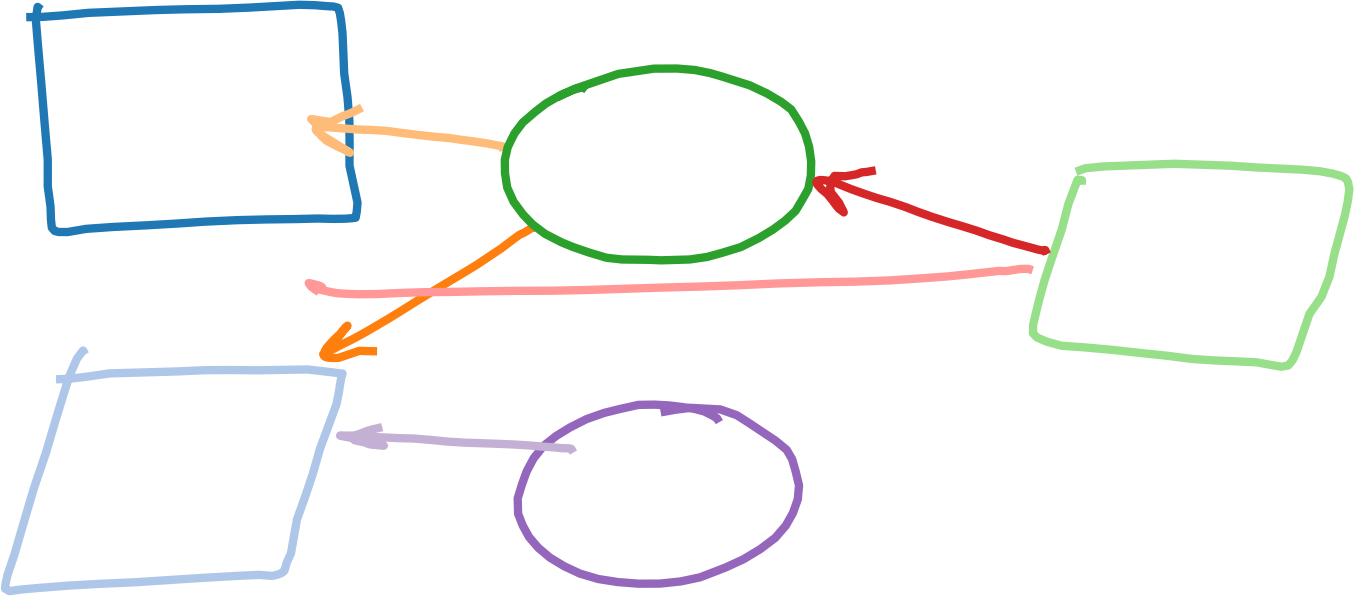}
\suppzoos{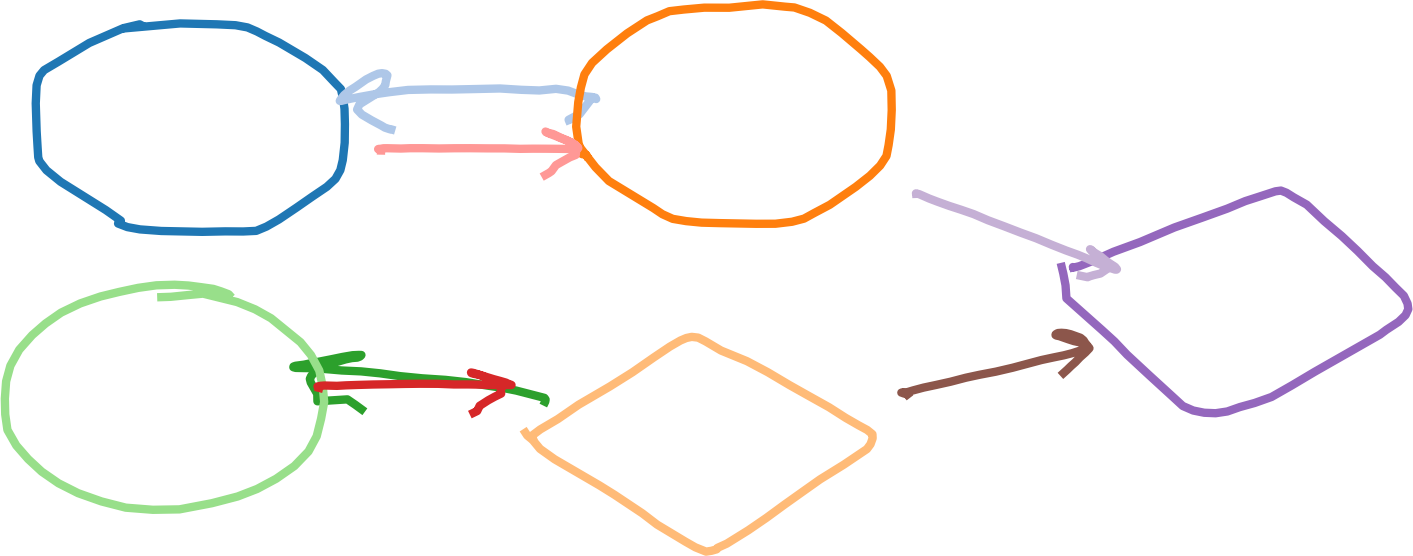}\\[2ex]

% \suppzoos{299_pos_ar_heatmap_ordered_s15}

\suppzoov{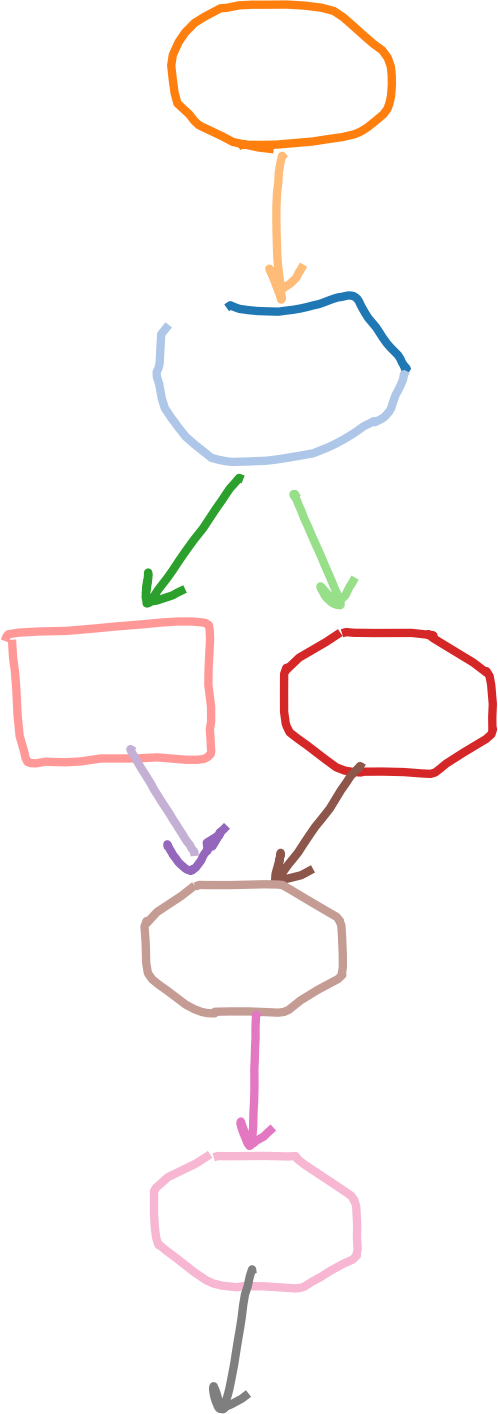}
\suppzoov{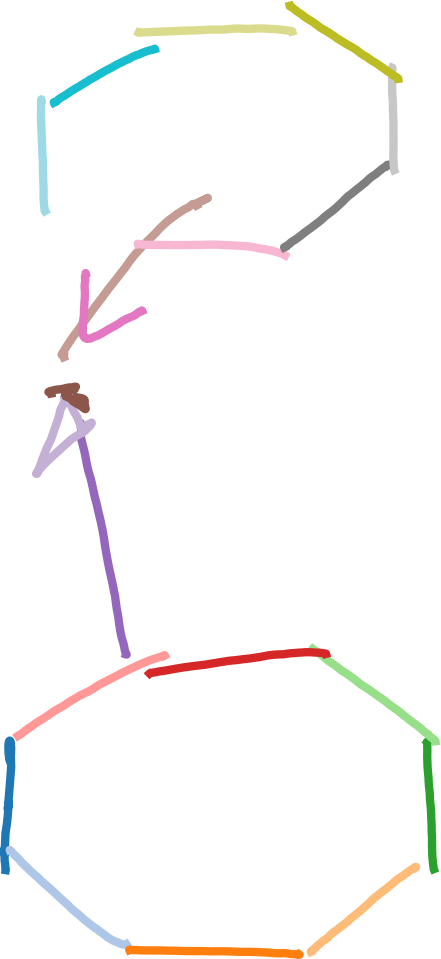}
\suppzoov{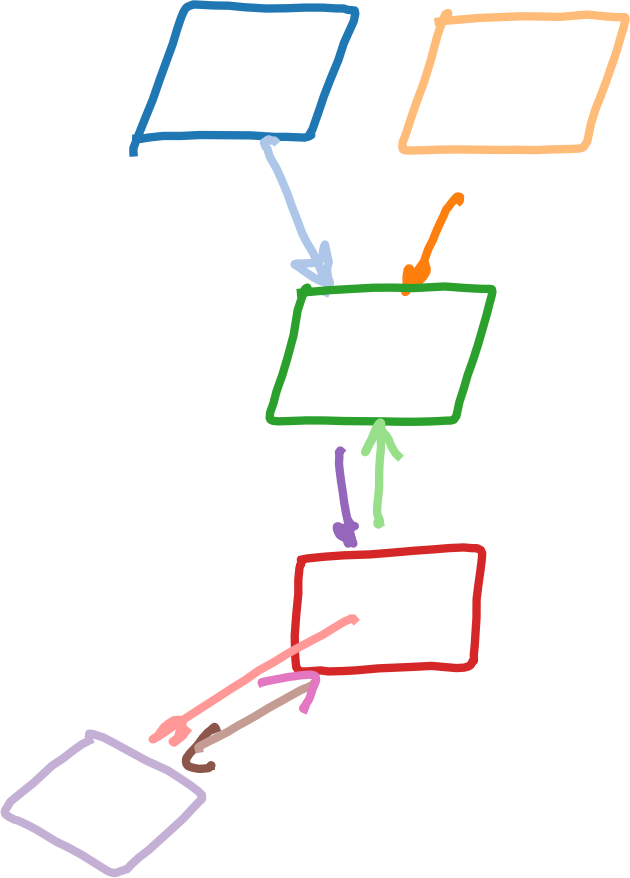}
\suppzoov{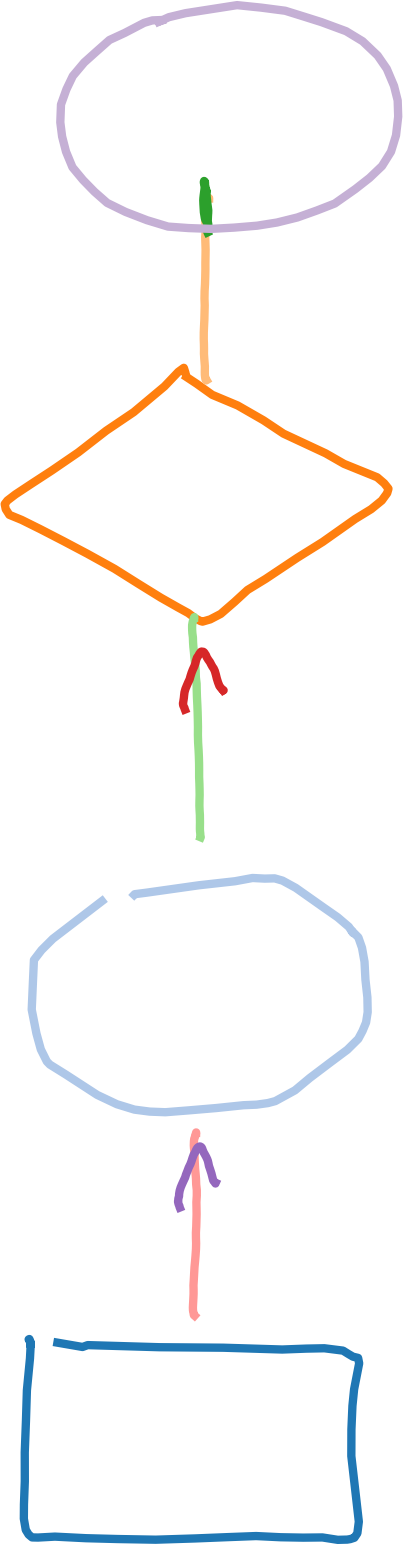} \\[2ex]
\suppzoov{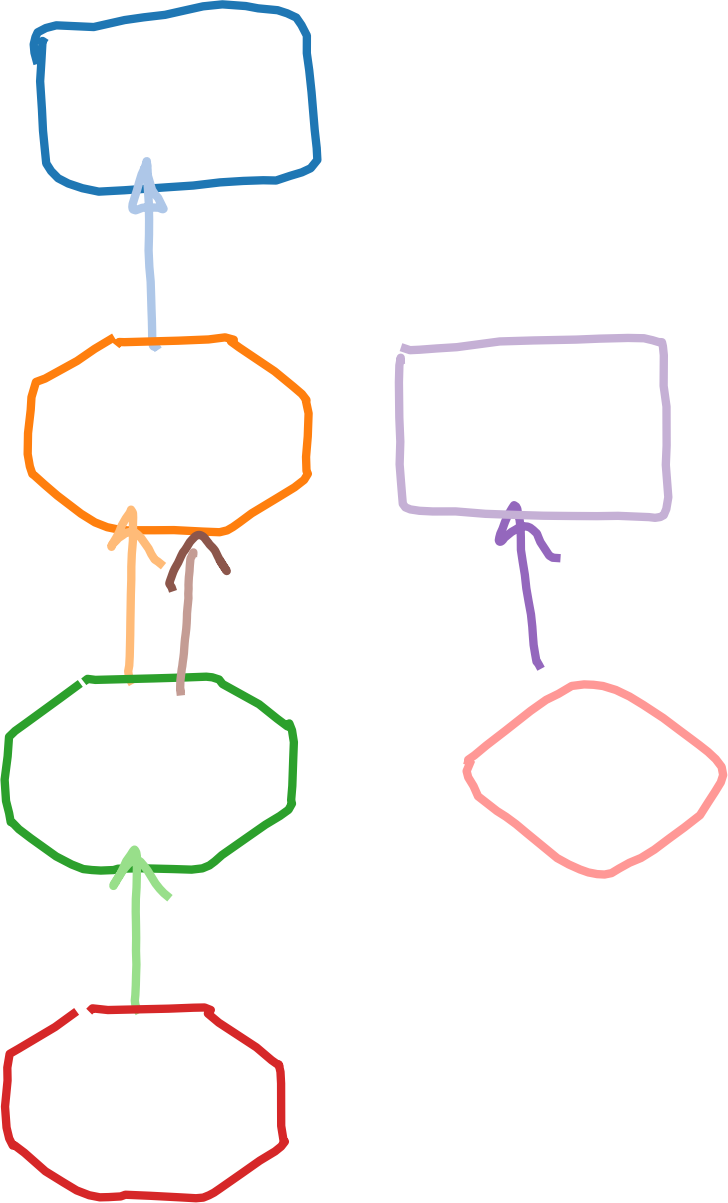}
\suppzoov{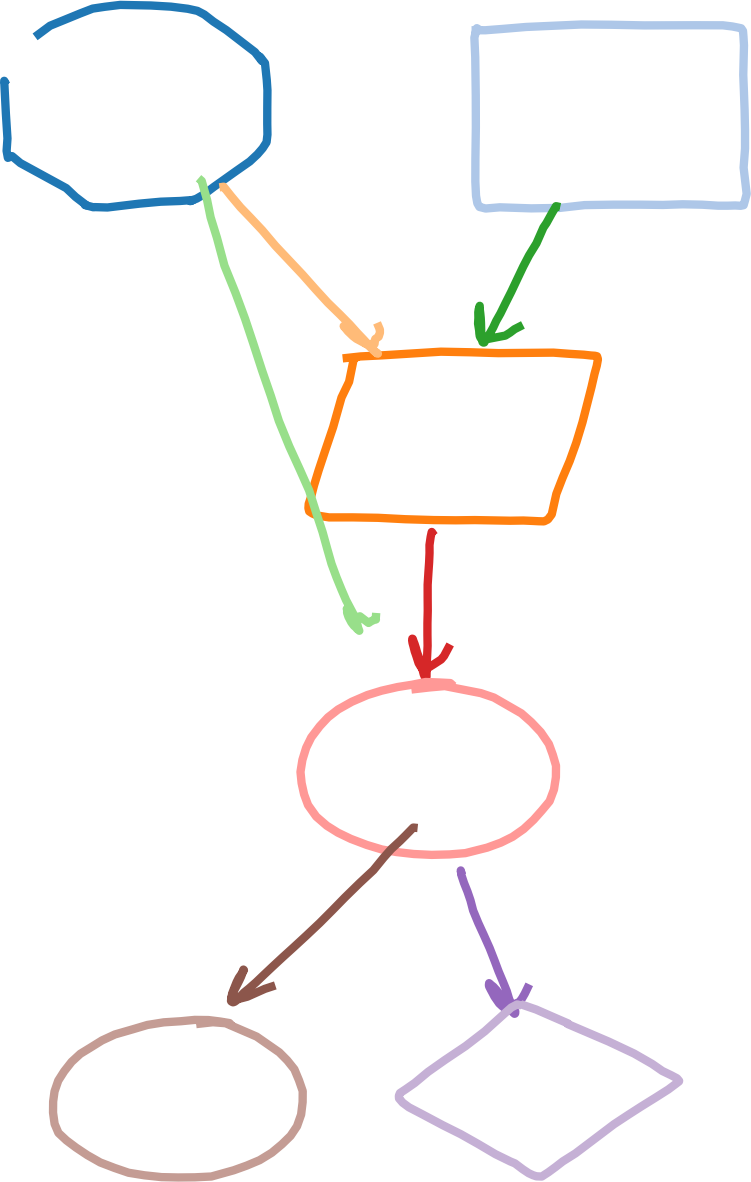}
\suppzoov{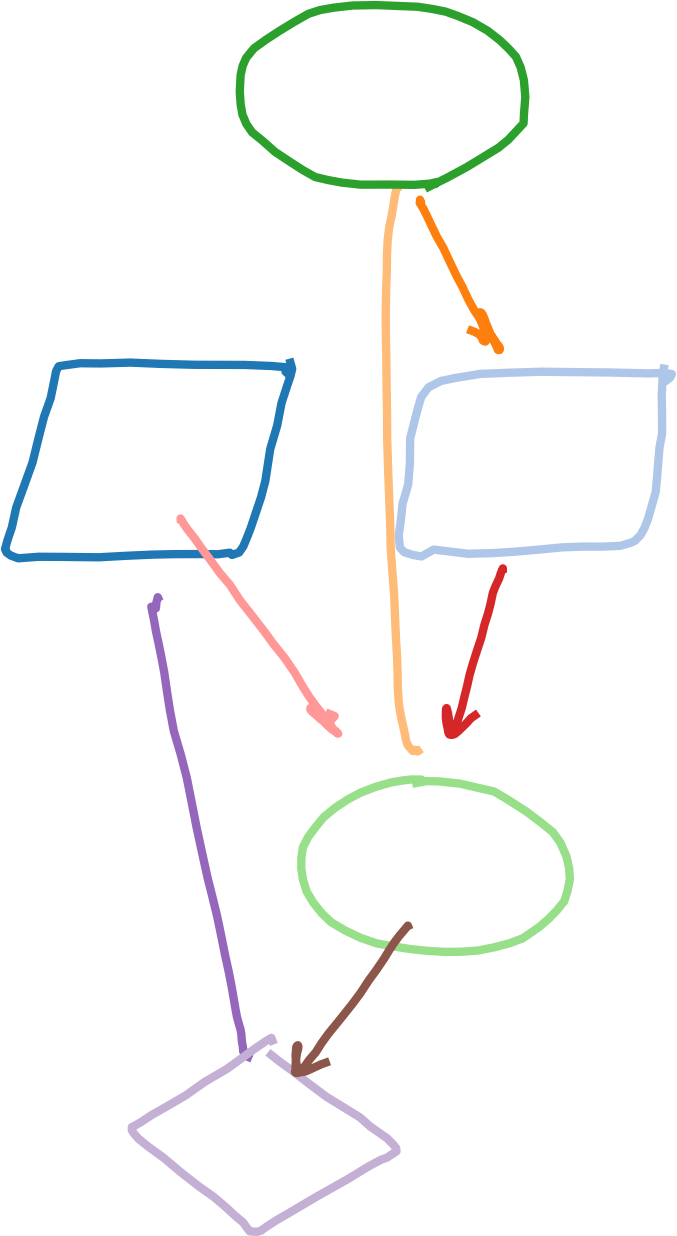}
\suppzoov{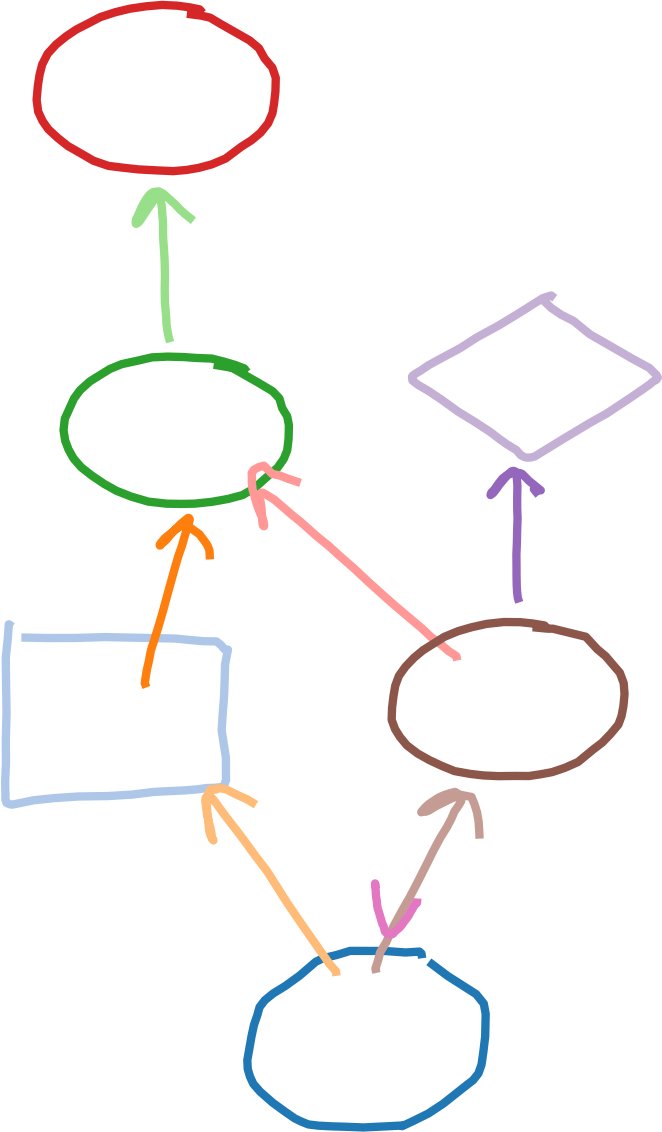}\\[2ex]

\strut\\
\caption{Given the \textcolor{firststroke}{\textbf{first}} and \textcolor{secondstroke}{\textbf{second}} strokes, random predictions of our model. 
}
\label{fig:suppzoo}
\end{figure}

In \Fig{suppzoo}, we present more samples predicted by our model.

